\pdfoutput=1

\documentclass[11pt]{article}

\usepackage[nohyperref, final]{acl}

\usepackage{times}
\usepackage{latexsym}
\usepackage{booktabs}
\usepackage{amssymb}
\usepackage{amsmath}
\usepackage{bbding}
\usepackage{subfig}
\usepackage{graphicx}
\usepackage{multirow}
\usepackage{pifont}
\usepackage[ruled,linesnumbered]{algorithm2e}

\definecolor{citec}{HTML}{882810}
\definecolor{refc}{HTML}{4658cf}
\definecolor{urlc}{HTML}{39a85c}
\usepackage[colorlinks,
            linkcolor=refc,
            anchorcolor=blue,
            urlcolor=urlc,
            citecolor=citec]{hyperref}

\usepackage[T1]{fontenc}

\usepackage[utf8]{inputenc}

\usepackage{microtype}

\usepackage{inconsolata}

\newcommand{\E}{\mathbb{E}_{x\in\mathcal{D}}}

\title{Understanding Token Probability Encoding in Output Embeddings}

\author{Hakaze Cho${}^{1,\text{\ding{73}}}$, Yoshihiro Sakai${}^{1}$, Kenshiro Tanaka${}^{1}$, Mariko Kato${}^{1}$, Naoya Inoue${}^{1,2}$\\
${}^{1}$Japan Advanced Institute of Science and Technology, ${}^{2}$RIKEN\\
${}^{\text{\ding{73}}}$\hyperref[sec:state]{Primary Contributor}, Correspondence to: \texttt{yfzhao@jaist.ac.jp}}

\begin{document}
\maketitle
\begin{abstract}
In this paper, we investigate the output token probability information in the output embedding of language models. We find an approximate common log-linear encoding of output token probabilities within the output embedding vectors and empirically demonstrate that it is accurate and sparse. As a causality examination, we steer the encoding in output embedding to modify the output probability distribution accurately. Moreover, the sparsity we find in output probability encoding suggests that a large number of dimensions in the output embedding do not contribute to causal language modeling. Therefore, we attempt to delete the output-unrelated dimensions and find more than 30\% of the dimensions can be deleted without significant movement in output distribution and sequence generation. Additionally, in the pre-training dynamics of language models, we find that the output embeddings capture the corpus token frequency information in early steps, even before an obvious convergence of parameters starts.
\end{abstract}

\section{Introduction}

Modern \textbf{L}anguage \textbf{M}odels (LMs) have two kinds of token embeddings: one is the \textbf{input embedding} $E^{(i)}$ located at the earliest layer of LMs, for mapping the input token index into distributed inner representation, the other is the \textbf{output embedding} $E^{(o)}$ in the \textbf{L}anguage \textbf{M}odeling \textbf{head} (LM head), for mapping the hidden state to the predicted probability distribution of the next token in the causal language modeling task.

Since the output embeddings were often \textit{tied} with the input embeddings~\cite{chung2020rethinking, press2017using}, i.e. input embeddings are directly used as the output embedding, the behaviors and features of independent output embeddings are rarely investigated. Along with the scaling of LMs, such embedding tying, which is proven to be harmful to model performance~\cite{chung2020rethinking}, is gradually being discarded in modern LMs such as GPT-J~\cite{gpt-j} and LLaMa 2~\cite{touvron2023llama}. This raises attention to the independent output embeddings, and explaining its underlying mechanism can be beneficial to understanding and improving LMs.

\begin{figure}[t]
    \centering
    \includegraphics[width=0.9\linewidth]{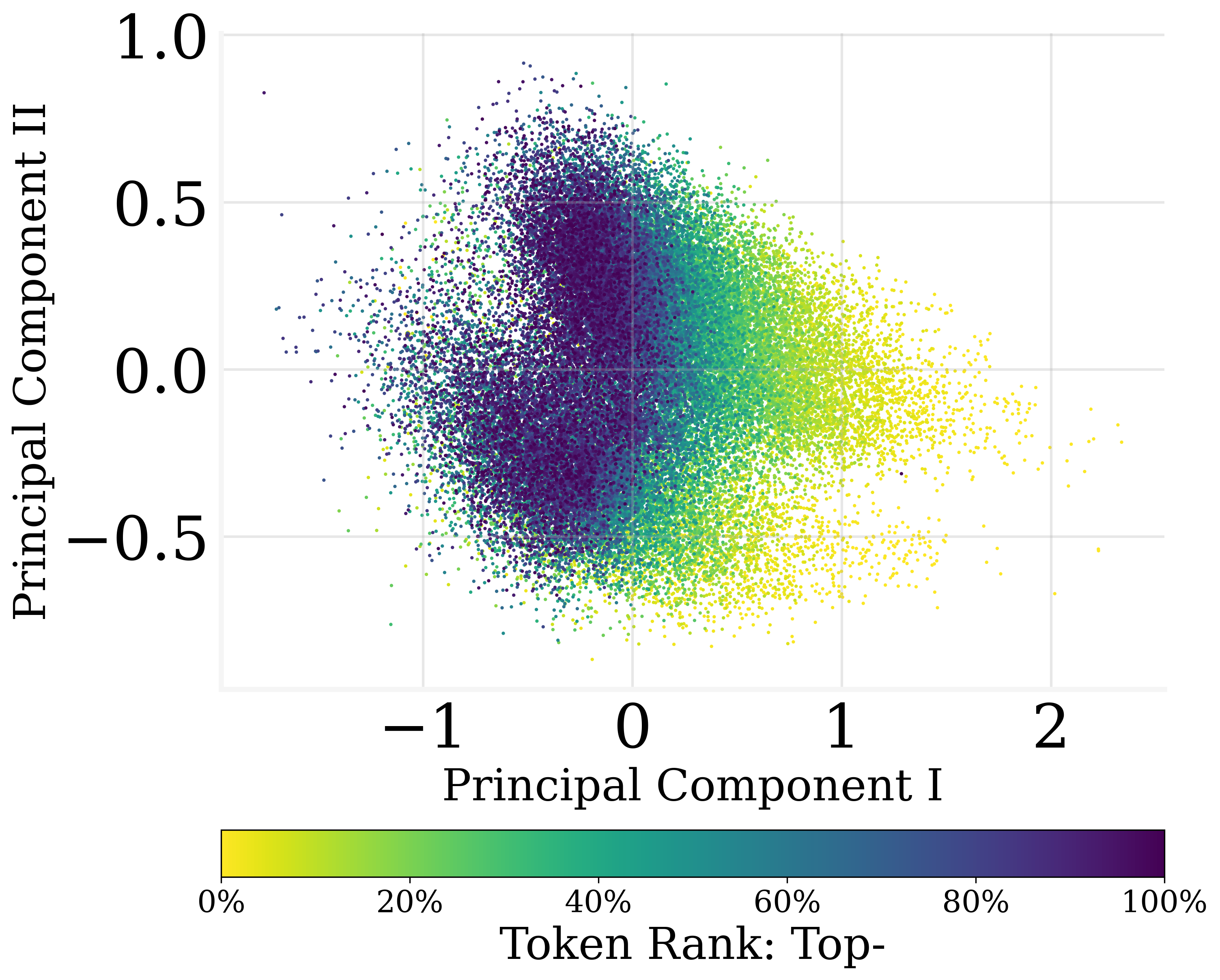}
    \caption{The PCA result of the output embedding vectors of GPT2. Colors refer to the ranking percentile of the averaged output token probability. }
    \label{fig:PCA2d-fre}
\end{figure}

The most obvious and expected role of the LM heads, where the output embedding is located, is to map the last hidden states into token probabilities. Therefore, following~\citet{kobayashi2023transformer}, who found an encoding of the averaged output token probability distribution in the bias term of the output LM head, this paper also focuses on the averaged output probabilities\footnote{We may omit the ``averaged'' in the following.} as an overall representation of LM outputs. We observe there is a linear-like correlation between the output token probabilities and the output embedding as shown in Fig.~\ref{fig:PCA2d-fre}. In \S\ref{sec:feqe}, our mathematical derivation indicates that $\mathrm{softmax}$-based LM head naturally encodes the output probabilities log-linearly in a common direction of the output embeddings, as long as the output dimension is sufficiently large, and the output values are concentrated. 

To prove our derivation empirically, we conduct \textbf{M}ultiple \textbf{L}inear \textbf{R}egression (MLR) on output probabilities against output embeddings, where a strong log-linear correlation is observed, as shown in Fig.~\ref{fig:fit}. Additionally, we find that: \textbf{(1)} Almost all directions highly correlated with output probability are the top principal components of the embedding matrix; \textbf{(2)} Only a few dimensions of output embedding are highly correlated to output probability. 

Based on such findings, to further examine the causality between the encoding and the output probabilities, in \S\ref{sec:editing}, we try to steer the output probability by the output embedding. We use a linear and sparse vectorized algorithm, modifying a small portion of dimensions along the detected encoding direction in the output embeddings to steer the probabilities. Our experiments find that: even on embedding-tied models, our steering method has respectable precision and a large available range, stably scaling the probability of tokens up to 20x (both scaling up or down), with little disturbing on the normal inference process of LMs. Moreover, with the encoding direction estimated from few-shot examples, the steering remains precise. Such results suggest that the log-linear correlation has good stability and generalization.

Moreover, the sparsity of the probability encoding demonstrates that most of the dimensions of the output embedding have minor effects on output. Therefore, we try removing these dimensions to prune the parameters in output embedding without obvious harm to the causal language modeling in \S\ref{sec:RDWE}. Our experiments show that more than 30\% of the output embedding dimensions can be removed without significant impairment.

Additionally, we use such log-linear encoding to investigate how and when the word frequency information of the training corpus is encoded into the output embedding during the training process. In \S\ref{sec:TrainD}, we find that the frequency encoding occurs at the very early steps in the training dynamics of LMs, even earlier than an obvious convergence trend is observed.

\vspace{0.5em}

\noindent\textbf{Our contributions can be summarized as:}

\begin{itemize}
    \item We find a log-linear correlation as an encoding of the output probability in a particular common and sparse direction of the output embeddings.
    \item As a causality examination of the findings, we steer the averaged output probabilities along the detected encoding direction, and remove dimensions with weak correlation to output probabilities without harm to the LMs.
    \item Based on the findings, we find that the LMs catch the token frequency in the training corpus at very early training steps.
\end{itemize}

\begin{table*}[t]
    \centering
    \caption{The goodnesses (Adj.$R^2$) of MLR of $-\log\alpha_{w,\mathcal{D}, \theta}$ against the output($E^{(o)}_w$) / input($E^{(i)}_w$) embedding. \textbf{Random Adj. $R^2$}: The Adj.$R^2$ of normalized random vector against $E^{(o)}_w$. \textbf{Parameter \#}: the number of LM parameters. \textbf{Embedding tied}: whether the output embedding is tied with the input embedding.}
    \label{tab:r2}
    \resizebox{\linewidth}{!}{
    \begin{tabular}{ccccccccc}
        \toprule
        \multirow{2}{*}{\textbf{Model}} & \multicolumn{4}{c}{\textbf{Decoder-only}} & \multicolumn{2}{c}{\textbf{Encoder-only}} & \multicolumn{2}{c}{\textbf{Encoder-decoder}} \\ \cmidrule(lr){2-5} \cmidrule(lr){6-7} \cmidrule(lr){8-9}
        & \textbf{GPT2} & \textbf{GPT2-XL} & \textbf{Pythia} & \textbf{GPT-J} & \textbf{BERT-base} & \textbf{BERT-large} & \textbf{BART-base} & \textbf{BART-large} \\ \midrule
        Parameter \# & 137M & 1.6B & 2.8B & 6B & 110M & 340M & 139M & 406M \\
        Embedding tied & \checkmark & \checkmark & \ding{55} & \ding{55} & \ding{55} & \ding{55} & \ding{55} & \ding{55} \\
        Random Adj.$R^2$ & $0.001$ & $0.000$ & $0.001$ & $0.000$ & $-$ & $-$ & $-$ & $-$ \\  \midrule
        Adj.$R^2$ on $E^{(o)}_w$ & $0.892$ & $0.893$ & $0.856$ & $0.882$ & $0.833$ & $0.826$ & $0.998$ & $0.952$ \\
        Adj.$R^2$ on $E^{(i)}_w$ & $(0.892)$ & $(0.893)$ & $0.814$ & $0.658$ & $-$ & $-$ & $-$ & $-$ \\
        \bottomrule
    \end{tabular}
    }
    \label{tab:my_label}
\end{table*}

\begin{figure*}[t]
\centering
\subfloat[GPT2, 137M, Principal]{
		\includegraphics[width=0.24\linewidth]{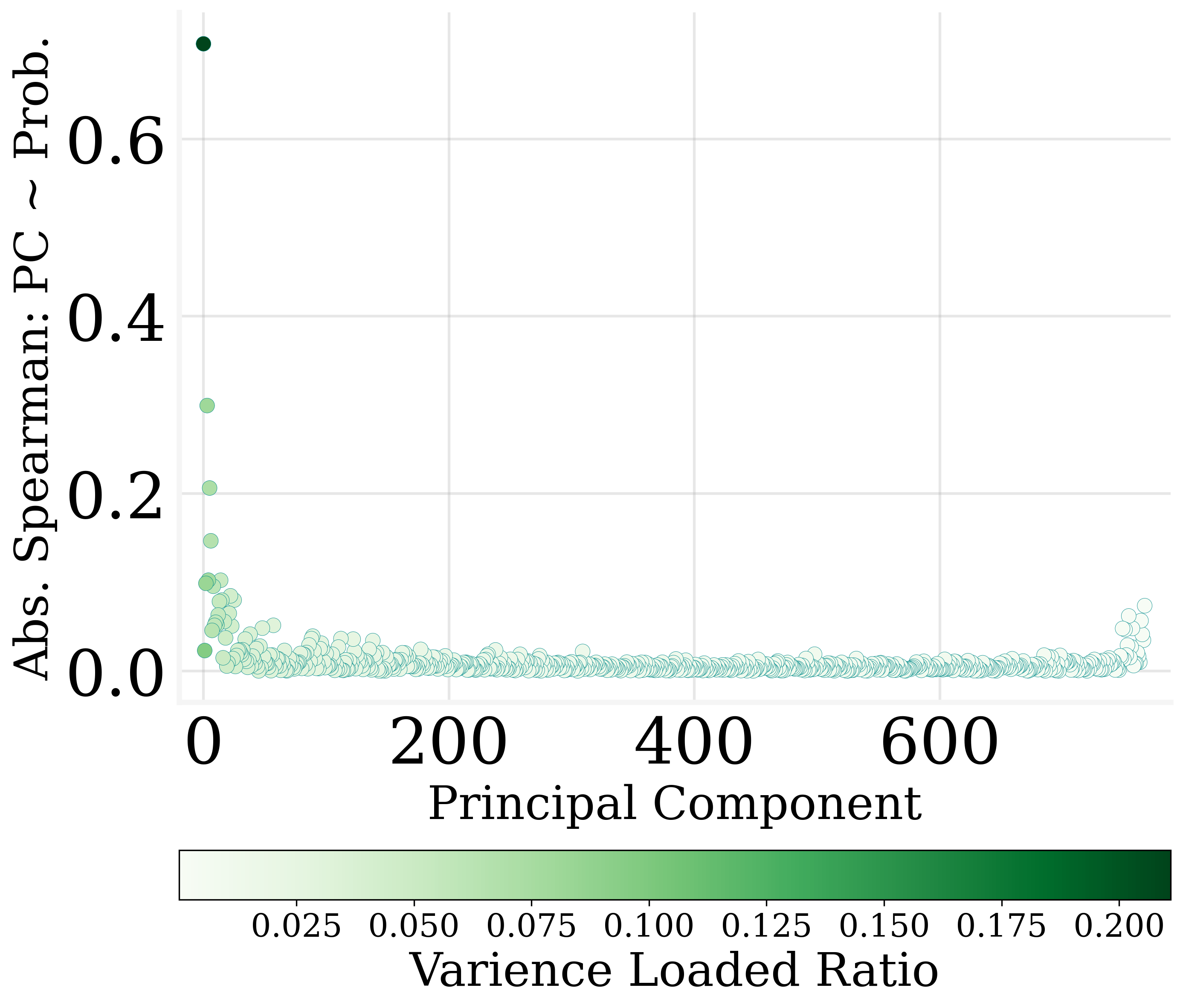}}
\subfloat[GPT2-XL, 1.6B, Principal]{
		\includegraphics[width=0.24\linewidth]{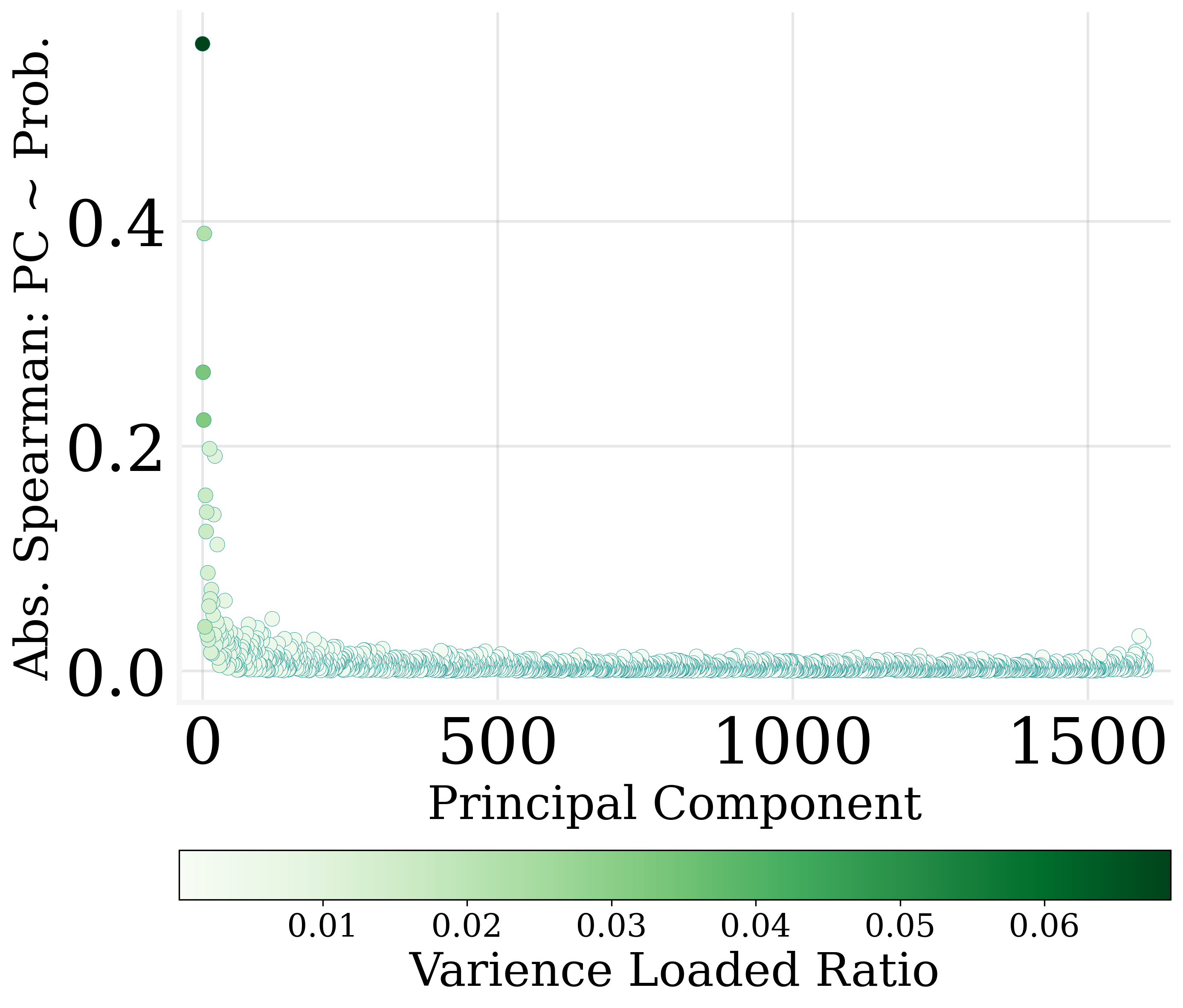}}
\subfloat[Pythia, 2.8B, Principal]{
		\includegraphics[width=0.24\linewidth]{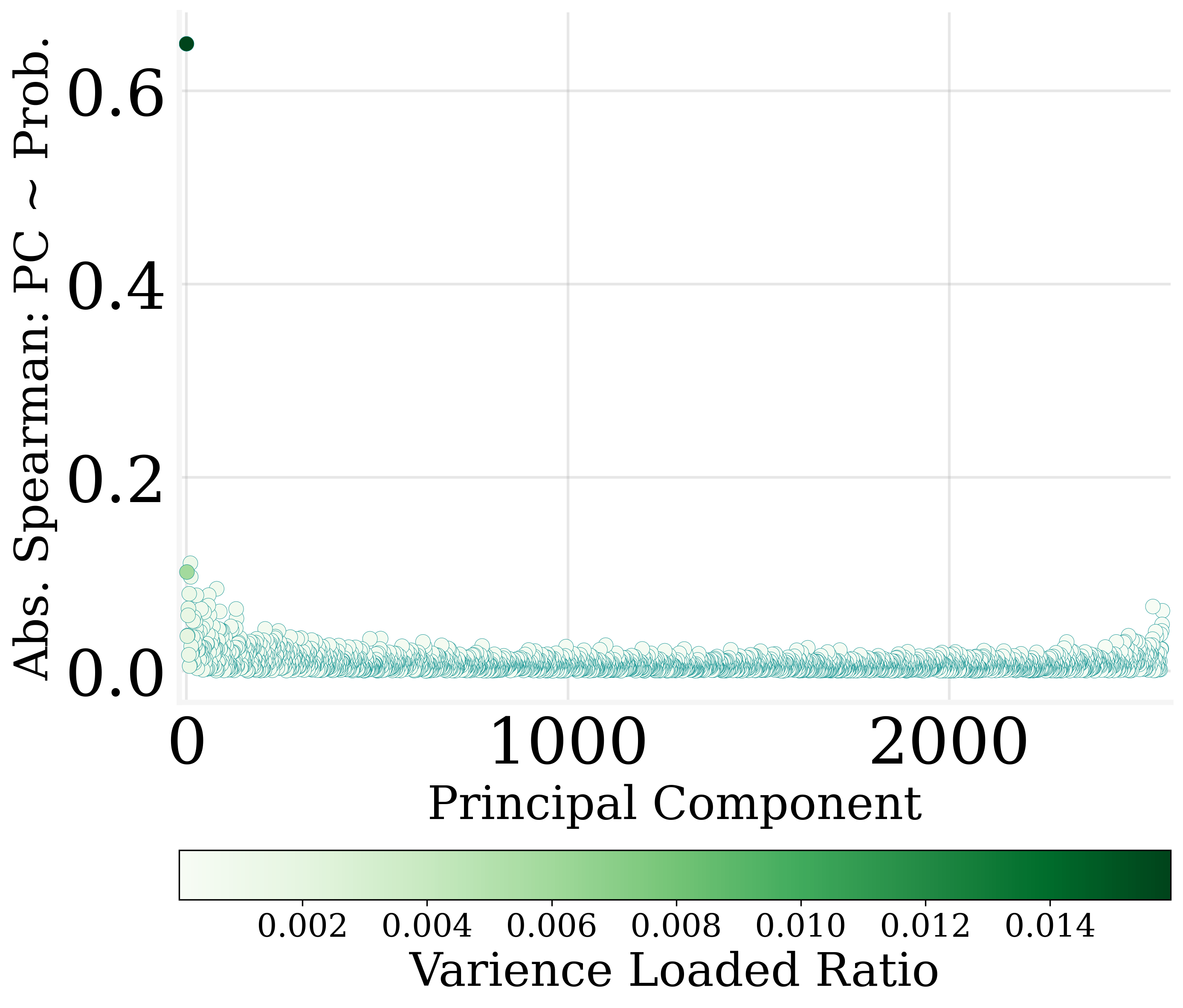}}
\subfloat[GPT-J, 6B, Principal]{
		\includegraphics[width=0.24\linewidth]{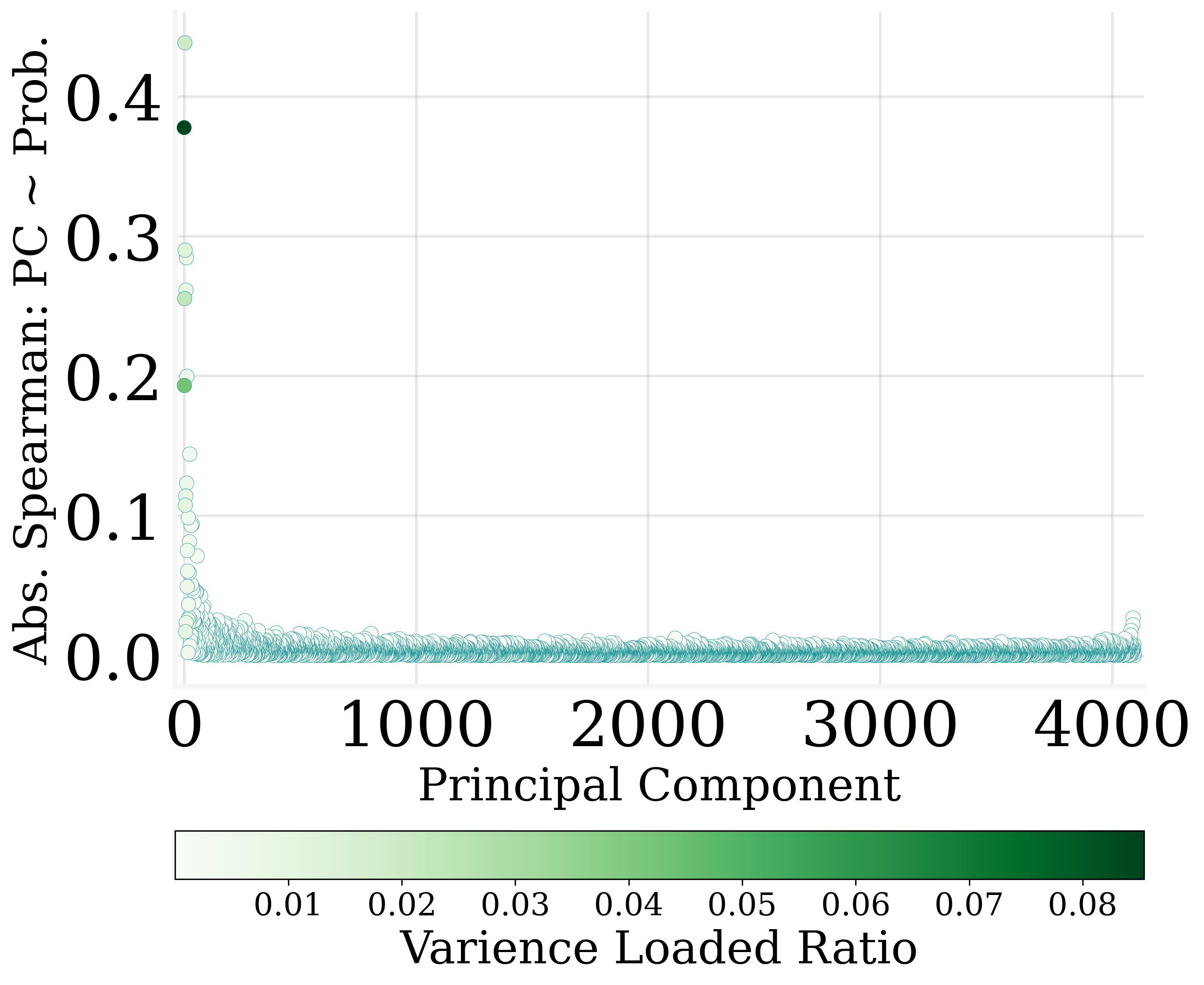}}
\newline
\subfloat[GPT2, 137M, Original]{
		\includegraphics[width=0.24\linewidth]{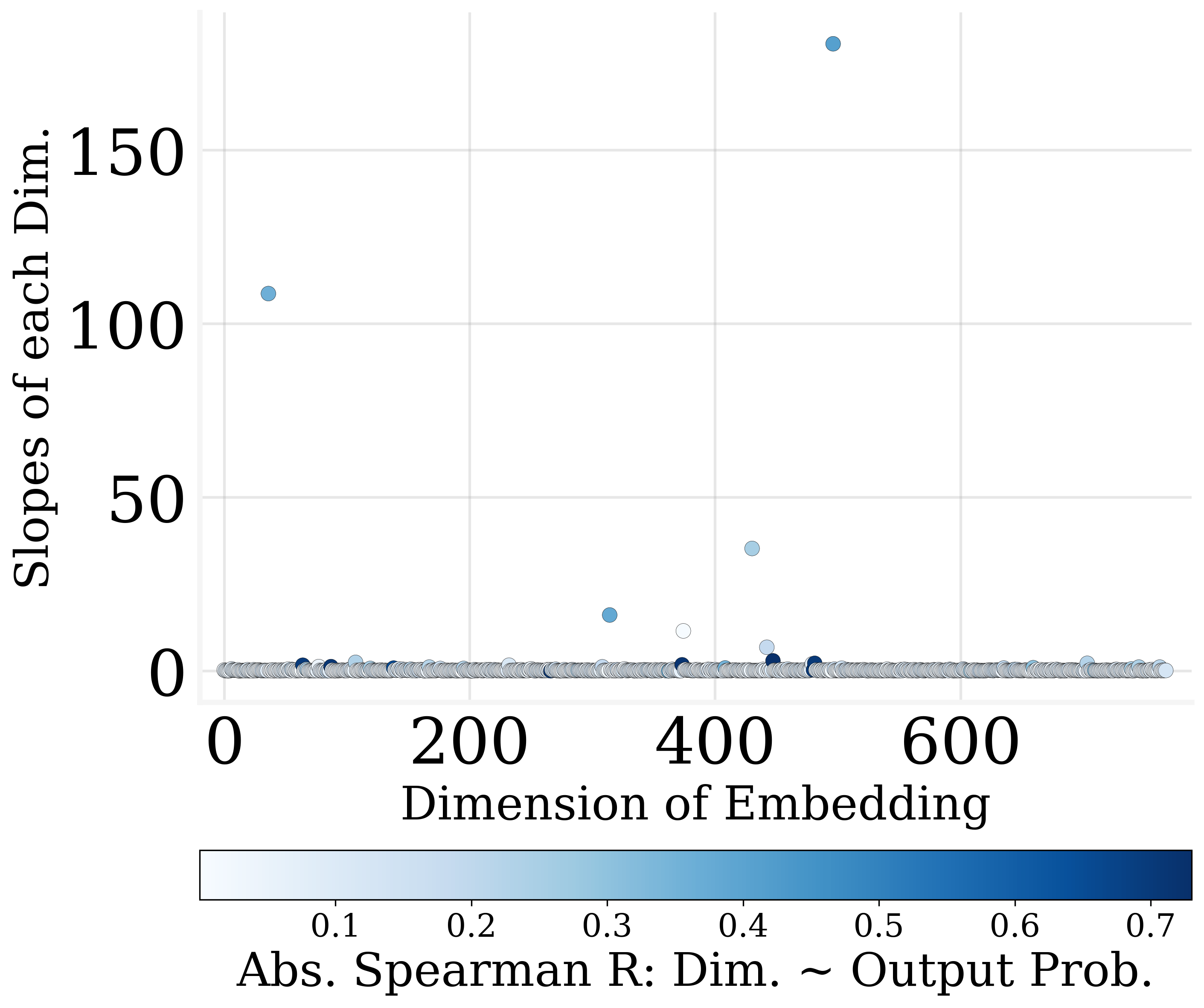}}
\subfloat[GPT2-XL, 1.6B, Original]{
		\includegraphics[width=0.24\linewidth]{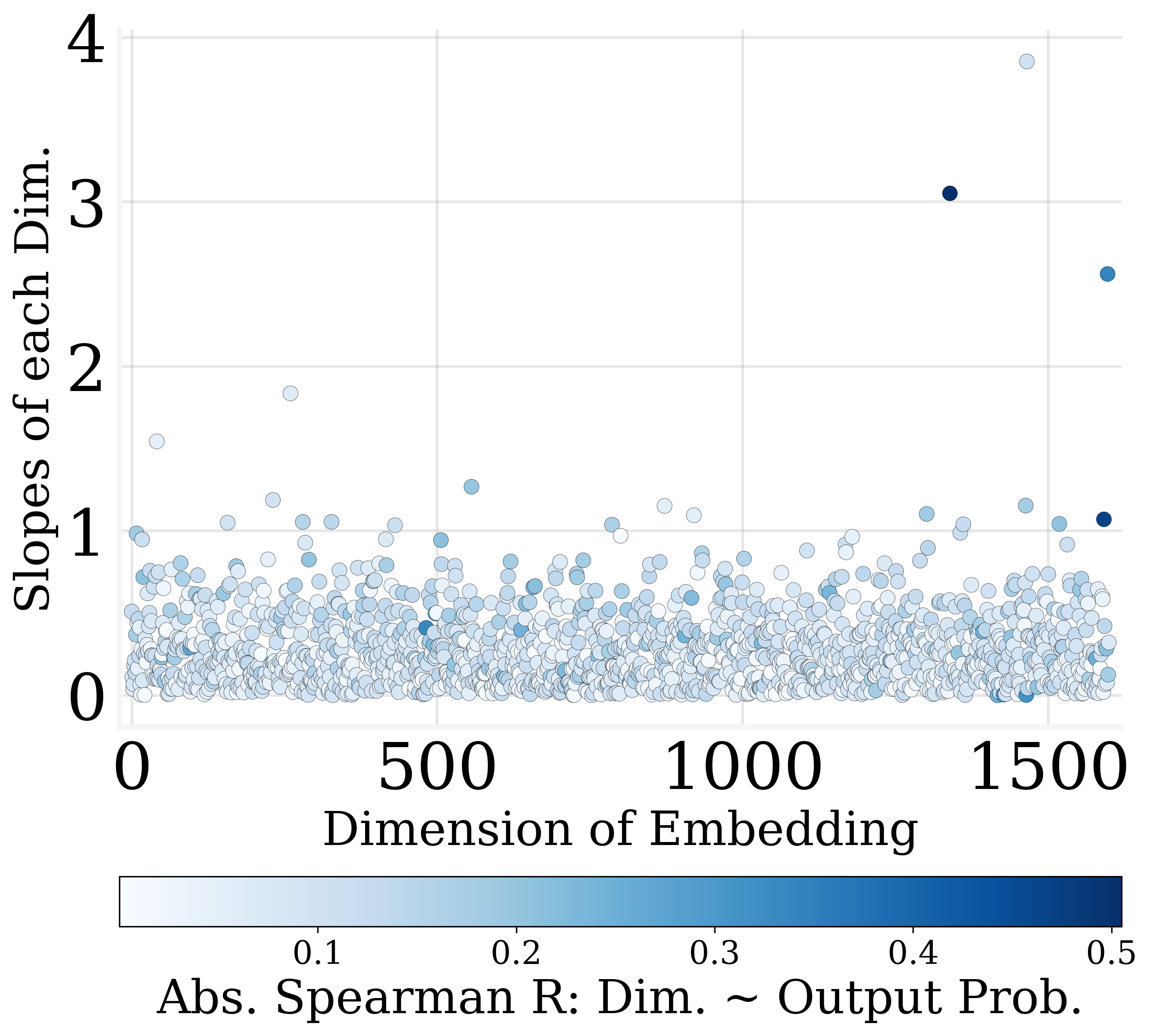}}
\subfloat[Pythia, 2.8B, Original]{
		\includegraphics[width=0.24\linewidth]{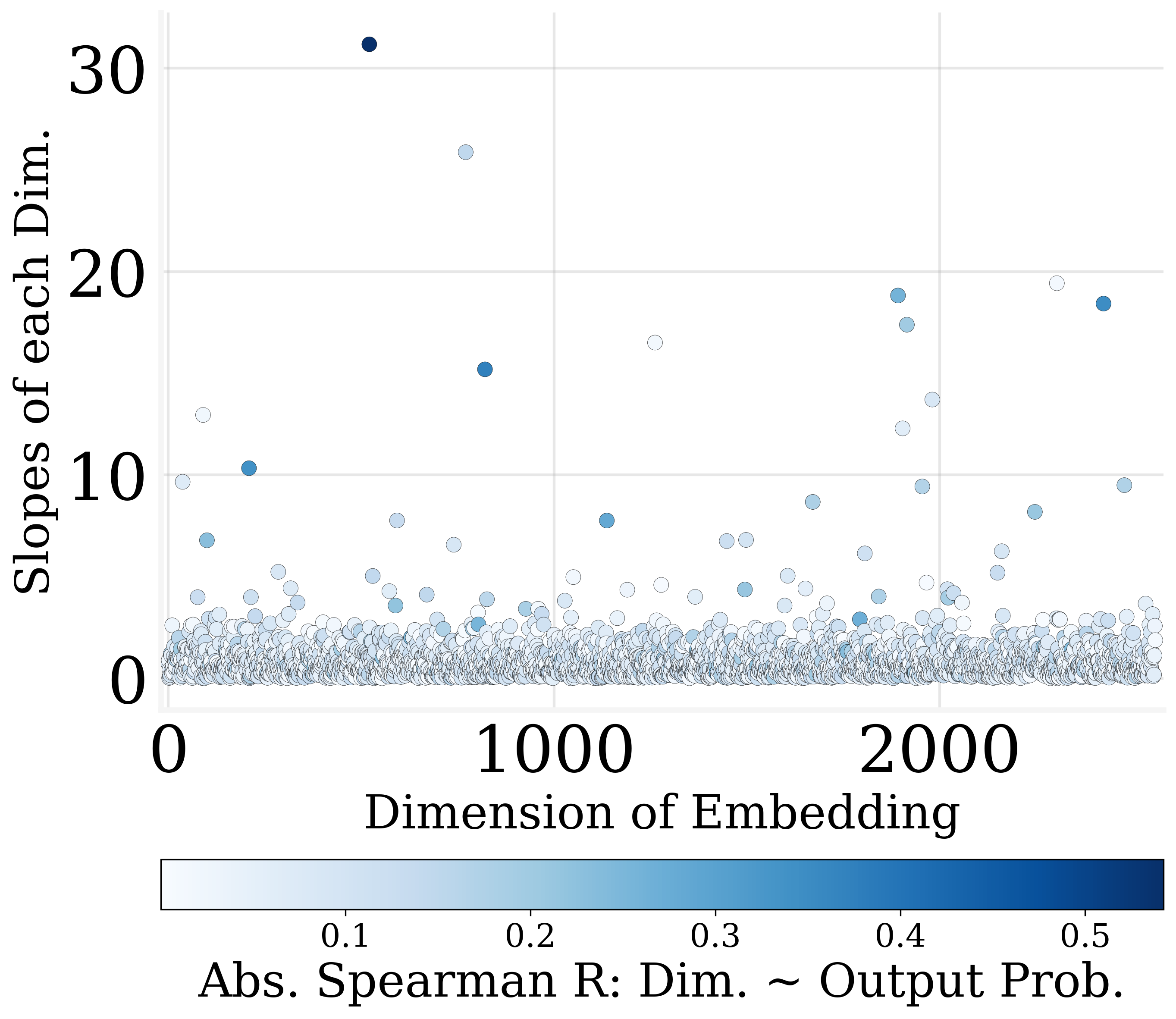}}
\subfloat[GPT-J, 6B, Original]{
		\includegraphics[width=0.24\linewidth]{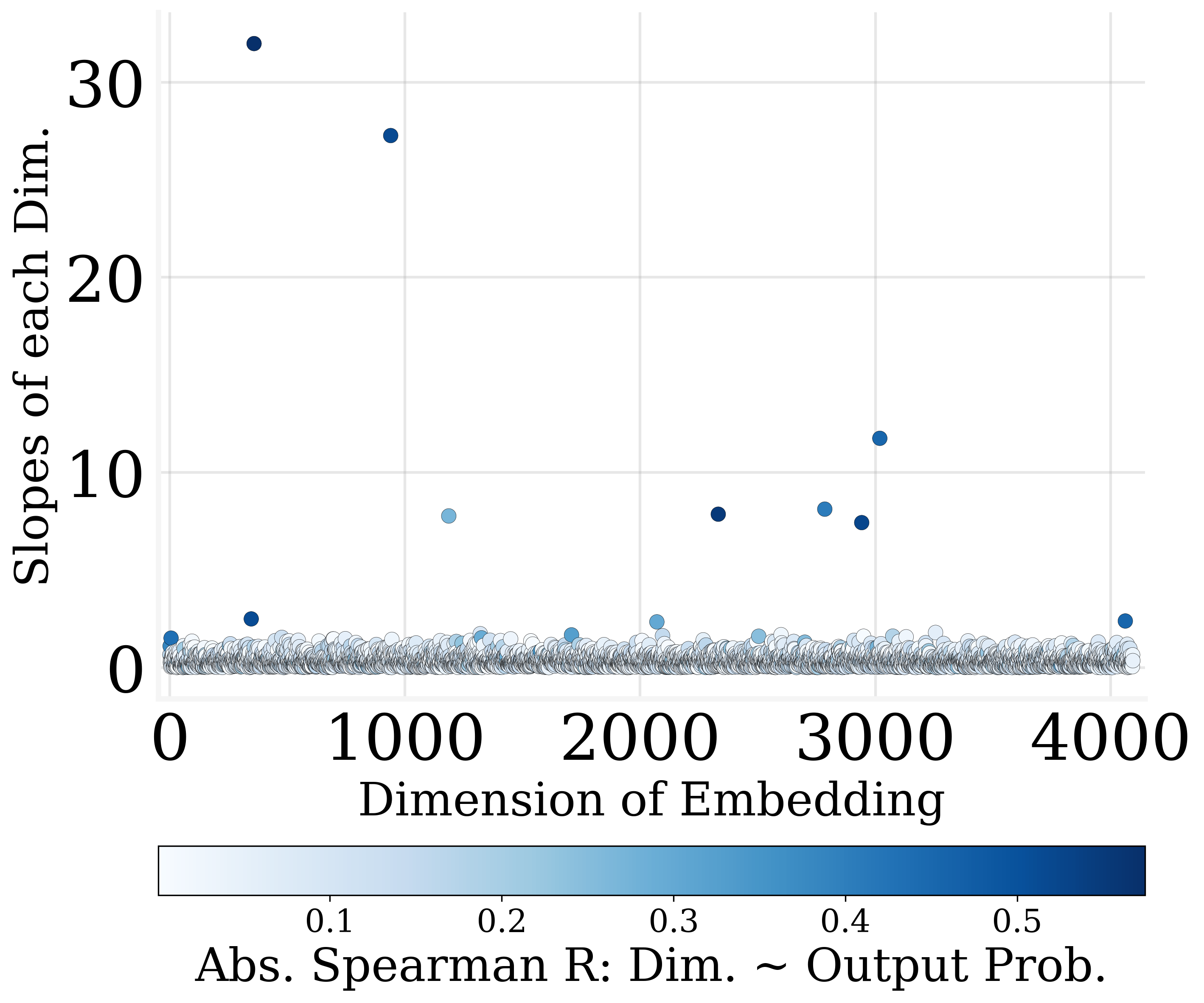}}
\caption{\textbf{Only a few directions/dimensions of output embedding are strongly correlated to the output probabilities.} \textbf{(a-d)}: \textbf{horizontal axis}: the principle components of output embedding, \textbf{vertical axis}: absolute Spearman $r$ between the principle and the output probability distribution, \textbf{color bar}: the variance ratio loaded in the principal component; \textbf{(e-h)}: \textbf{horizontal axis}: original dimensions, \textbf{vertical axis}: absolute MLR slopes between this dimension and the output probability distribution, \textbf{color bar}: the absolute Spearman correlations on the dimension.}
\label{fig:fewPCA}
\end{figure*}

\section{Token Probability Encoding in Output Embedding}
\label{sec:feqe}

In this section, we preliminarily reveal how the output probabilities are encoded in the output embedding by mathematical derivation, then conduct numerical experiments to confirm it empirically.

\begin{algorithm*}[t]
	\caption{Output token probability steering.\\ $\mathbf{\alpha}_{\mathcal{D}, \theta}\in\mathbb{R}^{\vert\mathbb{V}\vert}$: the output token probability distribution among the vocabulary; $|\mathcal{D}|$: the length of the probability detecting dataset; $\mathcal{S}$: the element-wise correlation significance ($p$-value) of MLR; \\ Element-wise calculations: $(\cdot)\times (\cdot)$: multiplication; $(\cdot)^{p}$: $p$-th power; $|\cdot|$: absolute value.}
	\label{alg:algorithm1}
	\KwIn{Language model $\mathrm{P}_\theta(x)$ with output embedding $E^{(o)}$; Probability detecting dataset $\mathcal{D}$; \newline Token index to be steered $w$; Expected scale of the steered token's probability $r$; \newline Steering amount allocation parameter $b$}
	\KwOut{Updated LM $P_{\theta'}(x)$}  
	\BlankLine

    $\mathbf{\alpha}_{\mathcal{D}, \theta} \leftarrow |\mathcal{D}|^{-1} \sum_{x\in\mathcal{D}} \mathrm{P}_\theta(x)$ \tcc*{\textcolor[HTML]{437579}{Calculate the averaged probability distribution}} 

    $A_\mathcal{D}, \mathcal{S} \leftarrow \mathrm{MLR}\left(- \log \mathbf{\alpha}_{\mathcal{D}, \theta}, E^{(o)}\right)$ \tcc*{\textcolor[HTML]{437579}{Conduct MLR, get the slope and significance}}

    $\Omega\leftarrow |A_\mathcal{D}|^b\times \mathcal{S} \times \left\Vert |A_\mathcal{D}|^b\times \mathcal{S} \right\Vert_1^{-1}$ \tcc*{\textcolor[HTML]{437579}{Allocate the steering weight to elements}}

    $E_w^{'(o)}\leftarrow E_w^{(o)} - \log(r) \Omega\times A_\mathcal{D}^{-1}$ \tcc*{\textcolor[HTML]{437579}{Apply the steering to output embedding}}

    \Return Updated LM $\mathrm{P}_{\theta'}(x)$ with new output embedding $E_w^{'(o)}$

\end{algorithm*} 

\subsection{Mathematical Log-linear Form}

Considering an LM parameterized by $\theta$ with vocabulary $\mathbb{V}$. Denoting the last hidden state w.r.t. input sequence $x$ as $h_x$, we can describe the output probability of token $w$ with an output embedding $E_w^{(o)}$ as:
\begin{equation*}
    \mathrm{P}_\theta(w|x) = \frac{e^{E_w^{(o)}\cdot h_x}}{\sum_{i\in\mathbb{V}}e^{E_i^{(o)}\cdot h_x}},
\end{equation*}
\noindent we have:
\begin{equation*}
    -\log \left[\mathrm{P}_\theta(w|x)\right] = - E_w^{(o)}\cdot h_x + \log\left[\sum_{i\in\mathbb{V}}e^{E_i^{(o)}\cdot h_x}\right].
\end{equation*}
When we calculate the averaged output token probability $\alpha_{w,\mathcal{D}, \theta}=\E[\mathrm{P}_\theta(w|x)]$ of token $w$ on a dataset $\mathcal{D}$ (\textit{detecting dataset}):
\begin{equation}
\begin{split}
    & -\log \alpha_{w,\mathcal{D}, \theta} \approx \E[-\log [\mathrm{P}_\theta(w|x)]] \\
    = & - \E[E_w^{(o)}\cdot h_x] + \E[\log(\sum_{i\in\mathbb{V}}e^{E_i^{(o)}\cdot h_x})] \\
    = & -E_w^{(o)}\cdot\E[h_x^{(o)}] + \E[\log(\sum_{i\in\mathbb{V}}e^{E_i^{(o)}\cdot h_x})].  
\label{eq.first}
\end{split}
\end{equation}
We make a local linear approximation in the approximately equal sign, while we confirm it is precise when the output logits are concentrated in Appendix~\ref{sec:appendix.Error}. Notice that if the LM head is biased, the bias term can be re-constructed equally by fixing one dimension of $h_x$ to $1$, w.l.o.g. 

We denote $A_{\mathcal{D}} = -\E[h_x]$, and $B_{\mathcal{D}}=\E\left[\log\left(\sum_{i\in\mathbb{V}}e^{E_i^{(o)}\cdot h_x}\right)\right]$. Here we make another approximation that the $B_{\mathcal{D}}$ is independent to $E_w^{(o)}$, which is precise when the logits of the output token $w$ are small, and the vocabulary size $|\mathbb{V}|$ is large. Then, we get an approximated log-linear form between the $\alpha_{w,\mathcal{D}, \theta}$ and the $E_w^{(o)}$:
\begin{equation}
    -\log \alpha_{w,\mathcal{D}, \theta} \approx A_{\mathcal{D}}\cdot E_w^{(o)} + B_{\mathcal{D}}.
    \label{eq:linear_output}
\end{equation}
As a special case, we consider a fixed-to-one dimension in $h_x$ (also in $A_\mathcal{D}$) for a biased LM head, where the bias re-constructed in $E_w^{(o)}$ becomes a linear factor of $-\log \alpha_{w,\mathcal{D}, \theta}$ with slope $1$. 

With such a derivation, we find that the phenomenon shown in Fig.~\ref{fig:PCA2d-fre} is the nature of $\mathrm{softmax}$-based output head if it has plenty of output dimensions, and small and concentrated logits to make the approximations numerically accurate. LMs have a very wide output space and undergo regularized training, which makes LMs meet the requirements to have an accurate log-linear correlation between output probabilities and output embeddings. That is, the output probabilities are encoded within a common direction of output embedding.

\subsection{Empirical Confirmation}

\begin{figure}[t]
\centering
\subfloat[GPT2, 137M]{
		\includegraphics[width=0.49\linewidth]{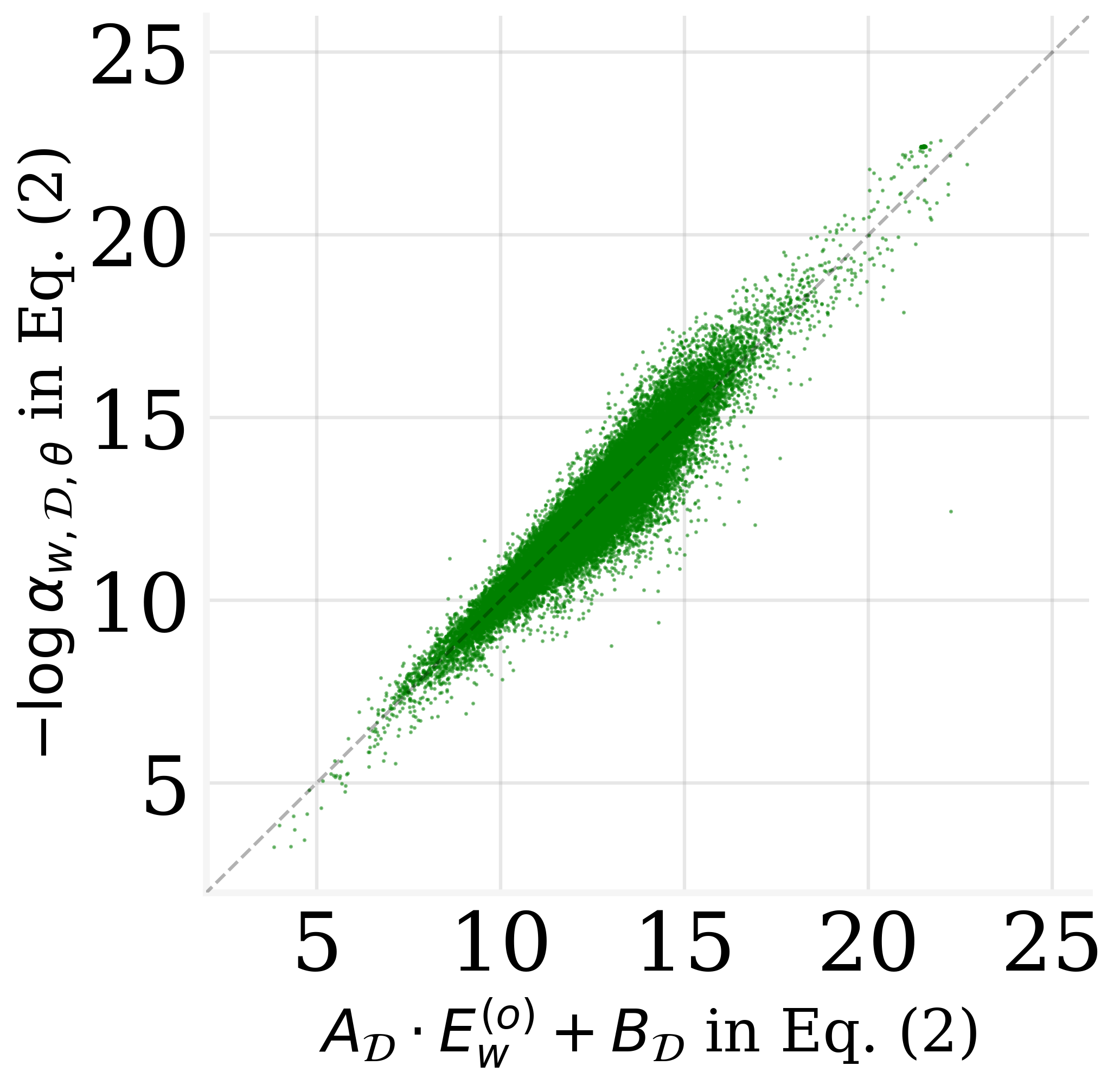}}
\subfloat[GPT-J, 6B]{
		\includegraphics[width=0.49\linewidth]{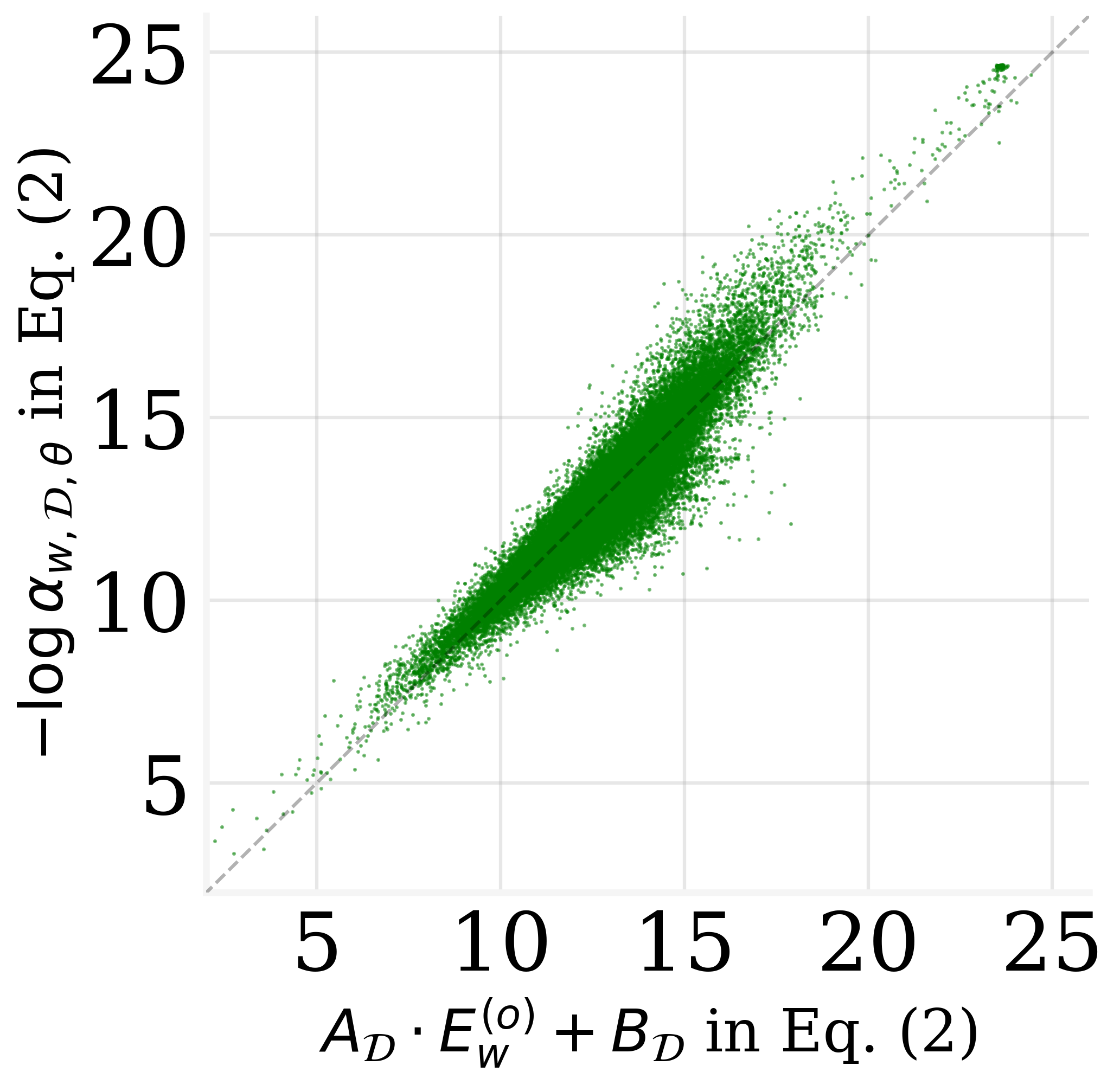}}
\caption{The MLR results on GPT2 and GPT-J.}
\label{fig:fit}
\end{figure}

We conduct experiments to prove our derivation in Eq.~\ref{eq:linear_output} empirically accurate. First, following \citet{kobayashi2023transformer}, we define the detect dataset $\mathcal{D}$ as an 8192-sample of shuffled \textsc{WikiDpr} dataset~\cite{karpukhin-etal-2020-dense} to calculate the averaged output probabilities. In detail, we input the sampled data points into the model, and average the output probability distribution among every time step (aligned with input token) of every input sequence as the token probability distribution of the dataset (see Appendix~\ref{sec:appendix.ATF}). Then, we conduct MLR to fit the $A_\mathcal{D}$ and the $B_\mathcal{D}$. 

We conduct such experiment on GPT2, GPT2-XL~\cite{radford2019language}, Pythia 2.8B~\cite{biderman2023pythia}, and GPT-J~\cite{gpt-j}. The fitting results are shown in Fig.~\ref{fig:fit}, where good fittings are observed. We also list the adjusted $R^2$ i.e. the goodness of fitting in Table \ref{tab:r2}. Surprisingly, both input and output embeddings have strong correlations with the token probabilities. Additionally, as a reference, we provide some rough results for the encoder-only (BERT-base and BERT-large~\cite{kenton2019bert}) and encoder-decoder (BART-base and BART-large~\cite{lewis2019bart}) models, in addition to the decoder-only model\footnote{Note: This paper focuses exclusively on an in-depth discussion of the decoder-only model.}.

More interestingly, as shown in Fig. \ref{fig:fewPCA}, we find that only the top principal components and a few dimensions in the output embedding are highly correlated to the output token probabilities. That is, such probability encoding is \textbf{sparse}. In detail, we calculate the correlation coefficient (absolute Spearman $r$ / MLR slope) between the output embedding's sorted principle components / original dimensions against the output probabilities, and we find that the overwhelming majority of both kinds of dimensions have poor correlations with the output probabilities.

\section{Token Probability Steering on Output Embeddings}
\label{sec:editing}


Our findings in \S\ref{sec:feqe} suggest that the output probabilities are encoded in a common direction in the output embeddings. So, as a causality examination, we steer the output embeddings along this detected direction and confirm that \textbf{(1)} such steering can control the output probabilities accurately on a large scale, implying a robust log-linear token probability encoding. \textbf{(2)} steering based on a few-shot detect dataset $\mathcal{D}$ can still be effective, confirming that such encoding is significant and widely present.

\subsection{Algorithm}

\begin{table*}[t]
\centering
\caption{Main results on the evaluation of Algorithm~\ref{alg:algorithm1}. \textbf{Unedited}: the baseline without any steering. \textbf{Random}: the baseline with a random $A_\mathcal{D}$. \textbf{Shuffled}: the baseline with a shuffled $A_\mathcal{D}$ from the calculated value. $b$: the softness parameter in Algorithm~\ref{alg:algorithm1}. Fine-grained results are reported in the Appendix \ref{sec:appendix.full}.}
\label{table:edit_res}
\resizebox{0.85\textwidth}{!}{
\begin{tabular}{@{}cccccccc@{}}
\toprule
& \multirow{2}{*}{\textbf{Emb. Tied}} & \multirow{2}{*}{\textbf{$b$}} & \textbf{Reliability} & \multicolumn{2}{c}{\textbf{Generalization}} & \multicolumn{2}{c}{\textbf{Specificity}} \\ \cmidrule(lr){4-4} \cmidrule(lr){5-6} \cmidrule(lr){7-8}
& & & $e_\mathrm{local}\downarrow$ & $e_\mathrm{id}\downarrow$ & $e_\mathrm{ood}\downarrow$ & $d_\mathrm{KL}^r\times 10^{-7}\downarrow$ & \textsc{MAUVE}$\uparrow$ \\ \midrule
\multicolumn{8}{c}{- \textit{Baseline Values} -} \\
\textbf{unedited} & $-$ & $-$ & ${1.10}_{1.06}$ & ${1.10}_{1.06}$ & ${1.10}_{1.06}$ & ${0.00}_{0.00}$ & ${1.00}_{0.00}$ \\ 
\textbf{137M, random} & \checkmark & 2 & ${1.33}_{1.28}$ & ${1.33}_{1.28}$ & ${1.29}_{1.24}$ & ${2.31}_{1.02}$ & ${0.96}_{0.01}$\\
\textbf{137M, shuffled} & \checkmark & 2 & ${1.18}_{1.07}$ & ${1.18}_{1.08}$ & ${1.18}_{1.08}$ & ${1.64}_{0.48}$ & ${0.96}_{0.01}$\\
\midrule
\multicolumn{8}{c}{- \textit{Experimental Values} -} \\
\textbf{137M} & \checkmark & 2 & ${0.19}_{0.30}$ & ${0.20}_{0.32}$ & ${0.15}_{0.28}$ & ${9.51}_{5.87}$ & ${0.96}_{0.01}$ \\
\textbf{1.6B} & \checkmark & 2 & ${0.64}_{0.67}$ & ${0.61}_{0.65}$ & ${0.65}_{0.63}$ & ${1.51}_{3.66}$ & ${0.91}_{0.17}$\\
\textbf{6B} & \ding{55} & 5 & ${0.31}_{0.43}$ & ${0.25}_{0.39}$ & ${0.10}_{0.12}$ & ${3.64}_{14.65}$ & $-$\footnotemark{} \\
 \bottomrule
\end{tabular}
}
\end{table*}

Based on the fact that the output probabilities are log-linearly encoded in a direction on the output embedding sparsely, we propose an output probability steering algorithm as shown in Algorithm~\ref{alg:algorithm1}. \textbf{Step 1:} We first estimate an averaged output probability distribution $\mathbf{\alpha}_{\mathcal{D}, \theta}$ on a detect dataset $\mathcal{D}$ and conduct MLR to calculate the encoding direction $A_\mathcal{D}$ and the correlation significance ($p$-value used) $\mathcal{S}$ of each element of the common encoding. \textbf{Step 2:} Then we assign a steering weight $\Omega$ to every dimension in the embedding vector based on the significance of its correlation to the output probabilities, as shown in line 3 of Algorithm~\ref{alg:algorithm1}. We allocate more steering amount to stronger correlations to obtain smaller and more accurate steering\footnotetext{It is difficult to conduct MAUVE experiments of GPT-J on such a repeating scale due to computational costs.}\footnote{Parameter $b$ is introduced to control the \textit{softness}, or \textit{sparsity} of such allocation, while the algorithm is stable on the parameter as shown in Appendix \ref{sec:appendix.Alloc}. Basically, we suggest that a large $b$ is suitable for a large model.}. \textbf{Step 3:} Given the token index to be steered and the expected steering scale, we calculate the detailed steering amount on each dimension and update it as shown in line 4 of Algorithm~\ref{alg:algorithm1}. 

As data costs, our steering only needs a detect set $\mathcal{D}$ and a feed-forward process to calculate the $A_\mathcal{D}$ and $\mathcal{S}$, and we are about to prove that it is stable for the size of $\mathcal{D}$ and consistently precise.

\subsection{Experiment Settings}
\label{sec:Experimentsettings}

We use the same detect dataset $\mathcal{D}$ as in \S\ref{sec:feqe}, and a set of scales of \{1, 1.1, 1.2, 1.5, 2, 5, 10, 20\} for both scaling up and down. We randomly select 10 tokens to be steered and conduct experiments on GPT2, GPT2-XL, and GPT-J.

\paragraph{Metrics.} We use 3 metrics to test the algorithm. 

\begin{itemize}
    \item \textbf{Scale error $e$}: To describe the precision of the probability steering, given the expected steering scale $r$ and the actual measured steered scale $\hat{r}$ on a test dataset, the scale error is calculated as $e = \vert\log(r) - \log(\hat{r}) \vert$.
    \item \textbf{KL divergence on the retained token $d_\mathrm{KL}^r$:} To investigate the side effect on the retained tokens, we calculate KL divergence between the probability distribution before and after steering with the steered token excluded.
    \item \textbf{\textsc{MAUVE}:} To investigate the side effect on text generation, we generate a set of sentences from the steered model, then calculate the MAUVE\footnote{Proposed by \citet{pillutla2021mauve}, a measurement of the similarity of two language datasets. The value range is $[0, 1]$, the larger means a greater similarity.} with the generated set from the original model (see Appendix \ref{sec:appendix.MAUVE}).
\end{itemize}

\paragraph{Evaluations.} Applying the aforementioned metrics, following the widely-used aspects of model editing evaluations~\cite{yao2023editing}, we define our evaluations in 3 categories: 

\begin{figure}[t]
\centering
	\includegraphics[width=0.85\linewidth]{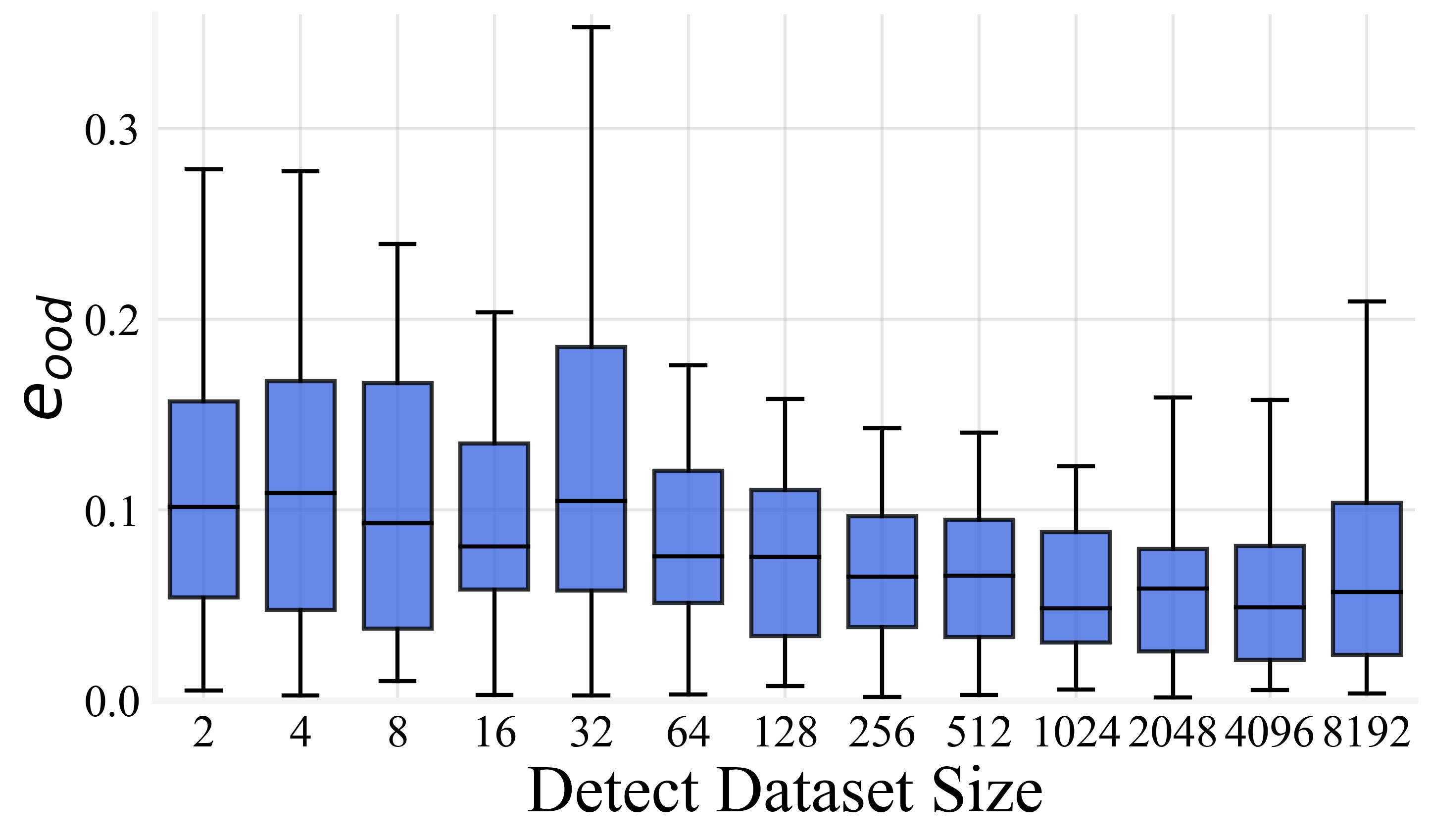}
\caption{The $e_{ood}$ on detect datasets with various numbers of sentence (averaged token per sentence $\approx 134$).}
\label{fig:length}
\end{figure}

\begin{itemize}
    \item \textbf{Reliability:} The local effectiveness of the steering. We use $e_\mathrm{local}$, the scale error with the detect dataset $\mathcal{D}$ as the test dataset.
    \item \textbf{Generalization:} The global effectiveness of the steering. We set multi-level generalization evaluations to ensure robustness: \textbf{(i)} In-domain Scale Error $e_\mathrm{id}$: the scale error with \textit{the other} 8192 samples of \textsc{WikiDpr} (where the updated parameters are estimated) as the test dataset; and \textbf{(ii)} Out-of-domain Scale Error $e_\mathrm{ood}$: the scale error on 2048 samples of \textsc{BookCorpus}~\cite{bookcorpus} as the test dataset. 
    \item \textbf{Specificity:} The harmless to the retained part. We use two metrics for this evaluation: \textbf{(i)} the averaged $d_\mathrm{KL}^r$ on the three data samples (detect set, in-domain, out-of-domain), and \textbf{(ii)} the aforementioned MAUVE.
\end{itemize}

\subsection{Results}

The main statistical evaluation results are shown in Table~\ref{table:edit_res}. Compared to the baseline values, the results from Algorithm~\ref{alg:algorithm1} are more accurate, generalizable, and harmless, even on the out-of-domain data. Especially, despite the embedding tying in GPT2 and GPT2-XL, such a steering algorithm is still accurate and harmless, which demonstrates that the log-linear probability encoding is \textit{orthogonal} to the possible semantical encoding in the output and also input embedding.

\begin{figure}[t]
\centering
\subfloat[GPT2, 137M]{
		\includegraphics[width=0.48\linewidth]{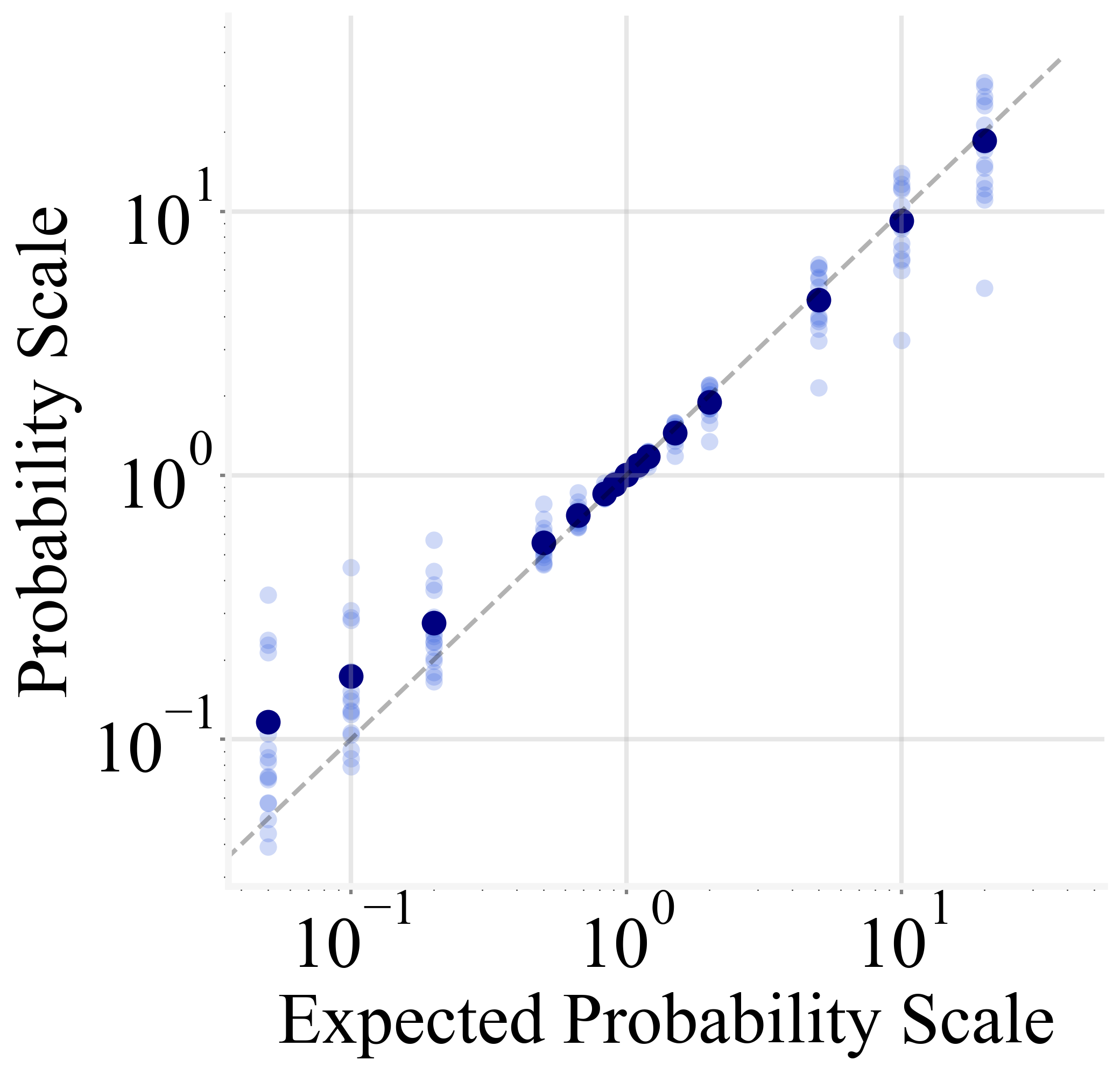}}
\subfloat[GPT-J, 6B]{
		\includegraphics[width=0.48\linewidth]{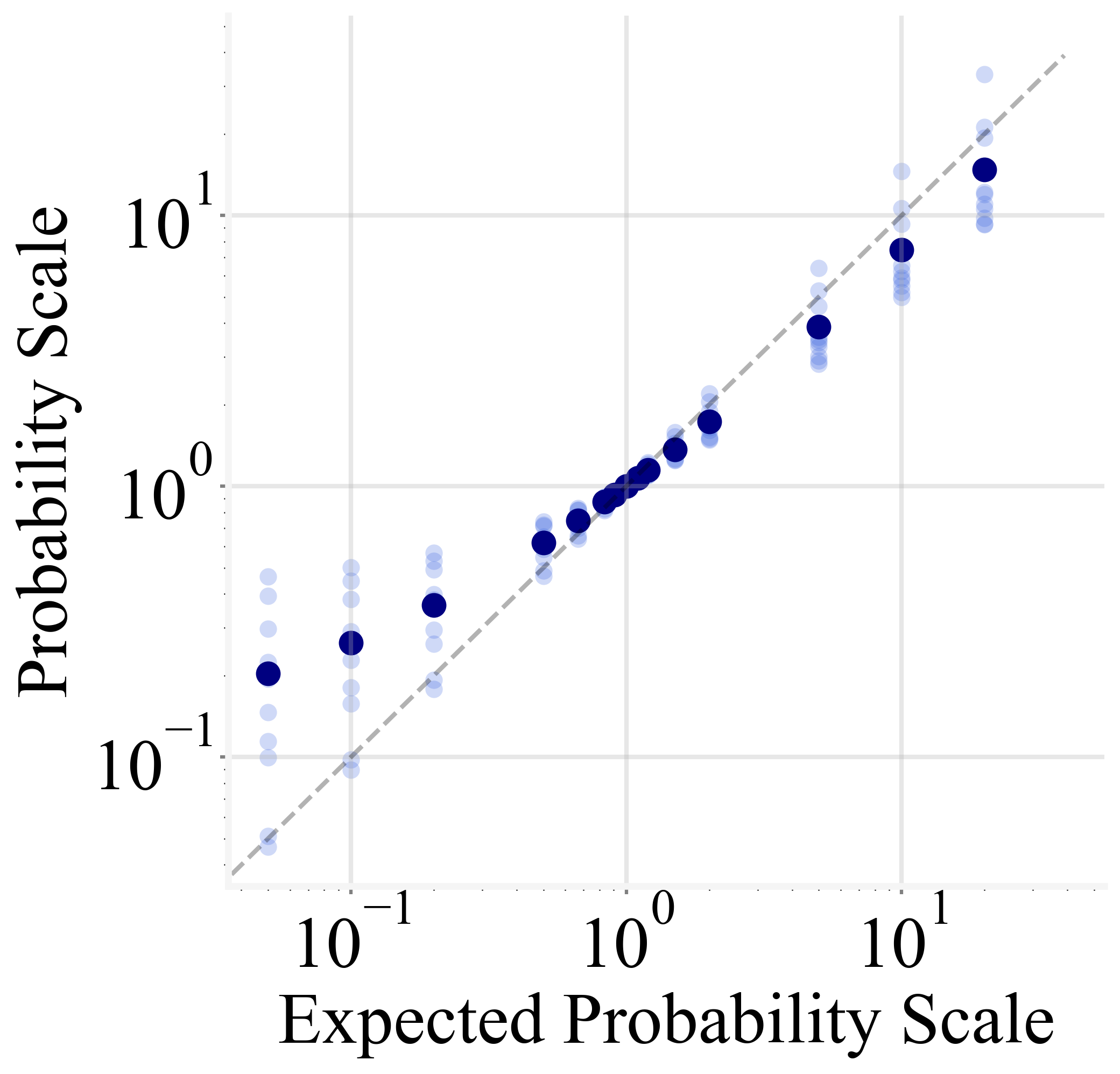}}
\caption{The expected probability scales against the actually steered scales measured in the steered LMs.}
\label{fig:editxy}
\end{figure}

\begin{figure*}[t]
\centering
\subfloat[GPT2, 137M, dim=768, tied]{
		\includegraphics[width=0.32\linewidth]{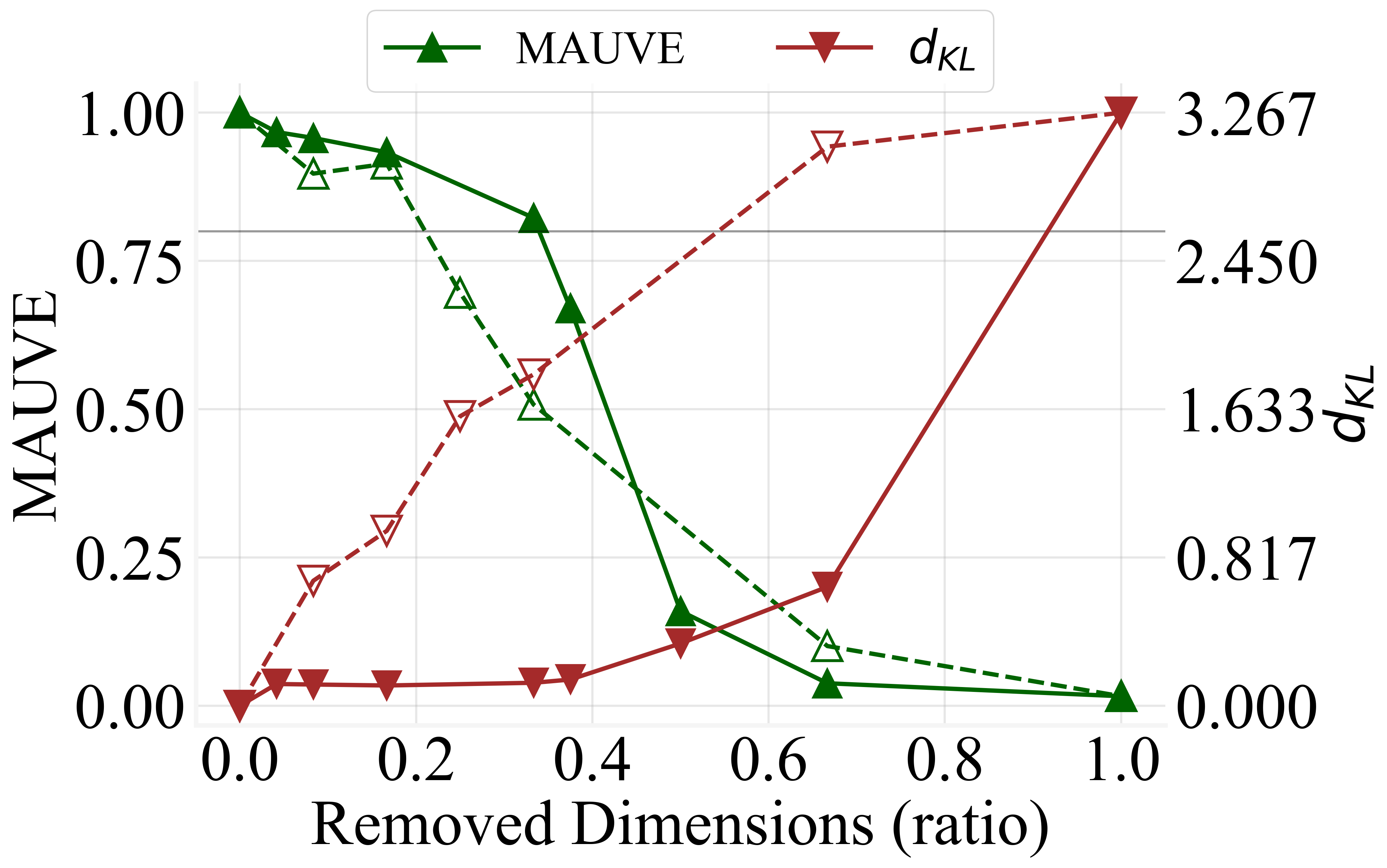}}
\subfloat[Pythia, 2.8B, dim=2560]{
		\includegraphics[width=0.32\linewidth]{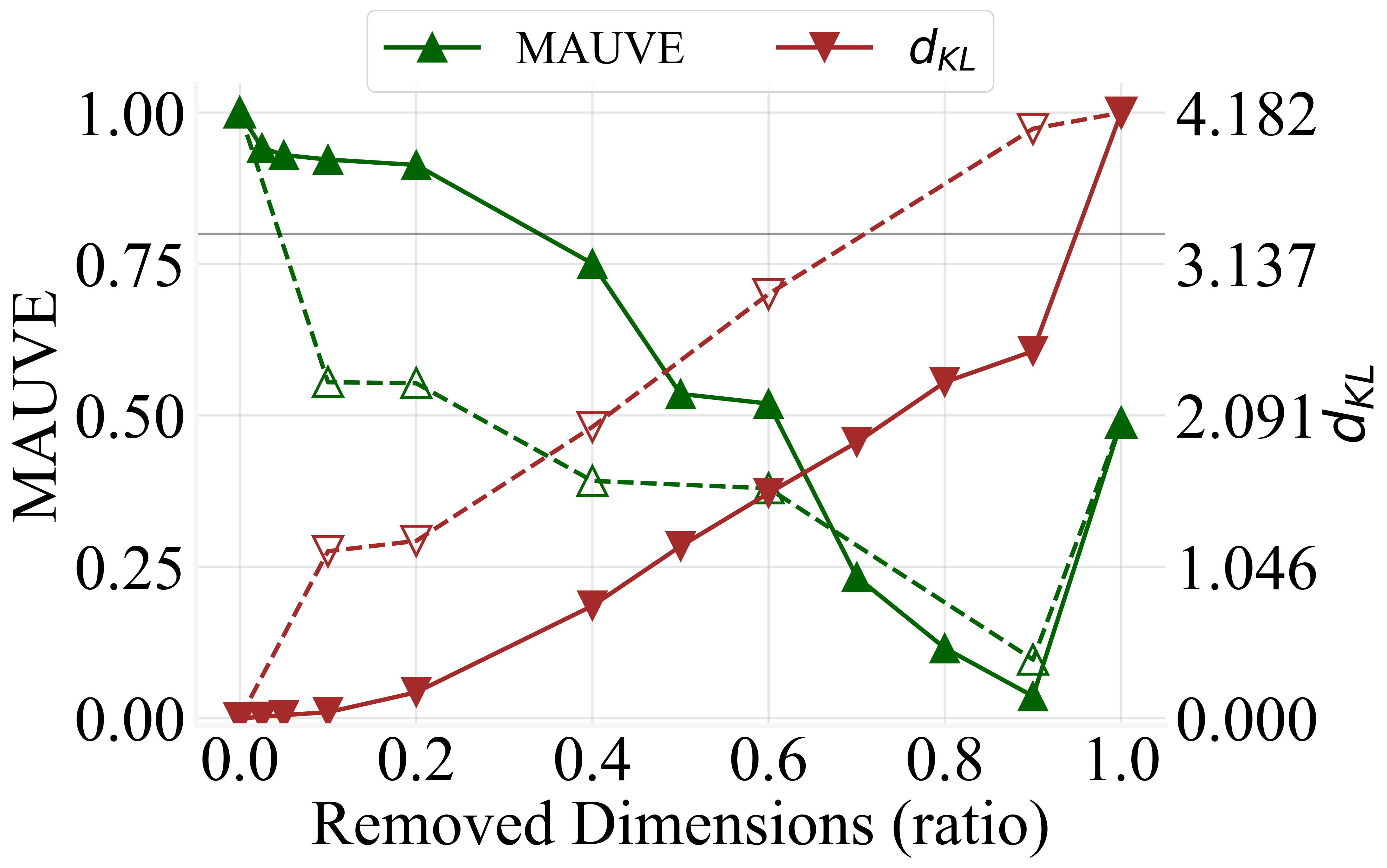}}
\subfloat[GPT-J, 6B, dim=4096]{
		\includegraphics[width=0.32\linewidth]{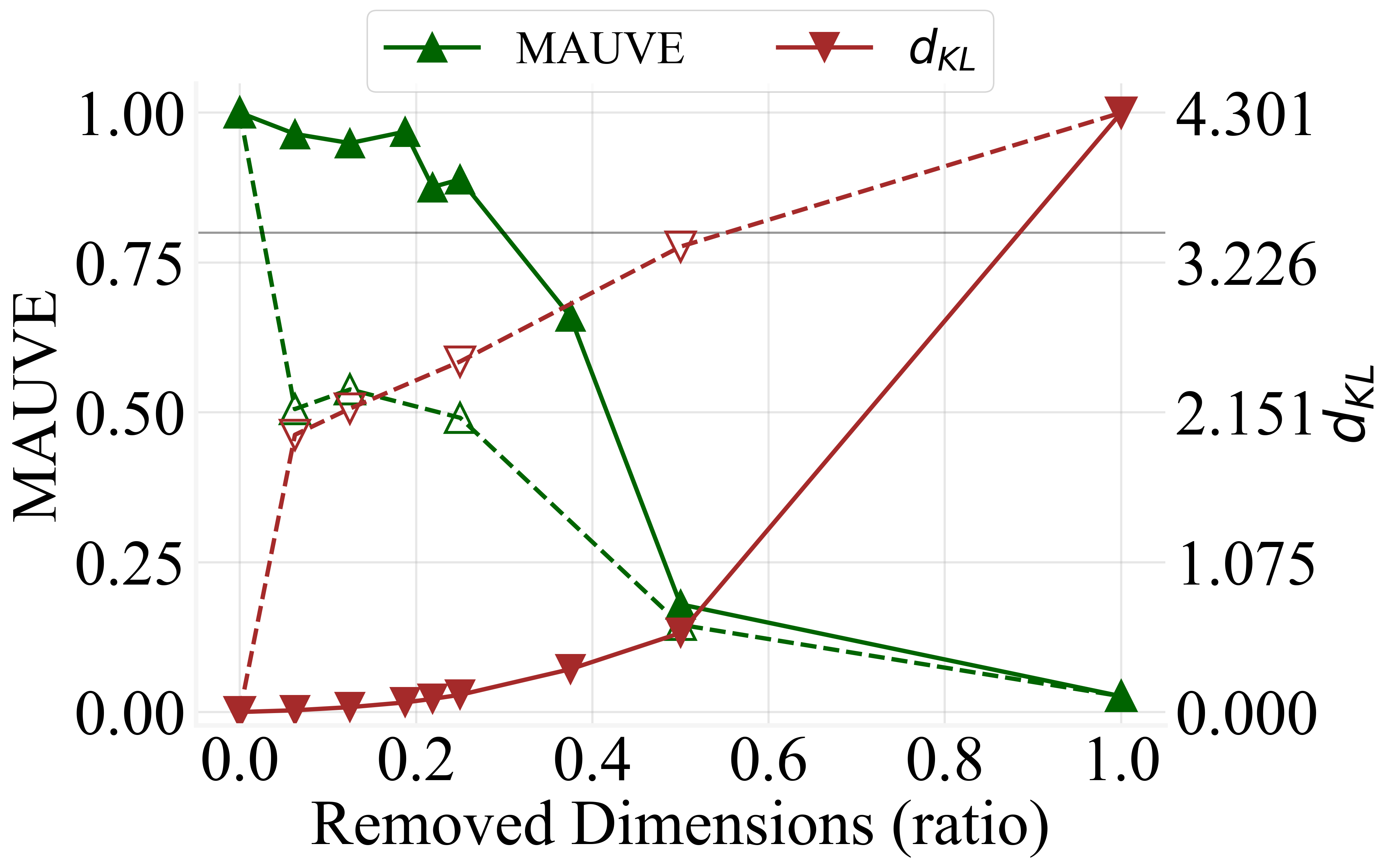}}
\caption{The \textcolor[HTML]{006400}{MAUVE$(\uparrow)$} and \textcolor[HTML]{a52a2a}{KL divergence$(\downarrow)$ of output token probability distribution $d_\mathrm{KL}$ against original distributions} w.r.t. the dimension removing ratio on the output embedding. \textbf{Solid curves}: results of removing dimensions from the least important ones to the most important ones; \textbf{Dashed curves}: adversarial controlling experiment, removing dimensions reversely.}
\label{fig:del}
\end{figure*}

\begin{figure*}[t]
\centering
\subfloat{
		\includegraphics[width=0.45\linewidth]{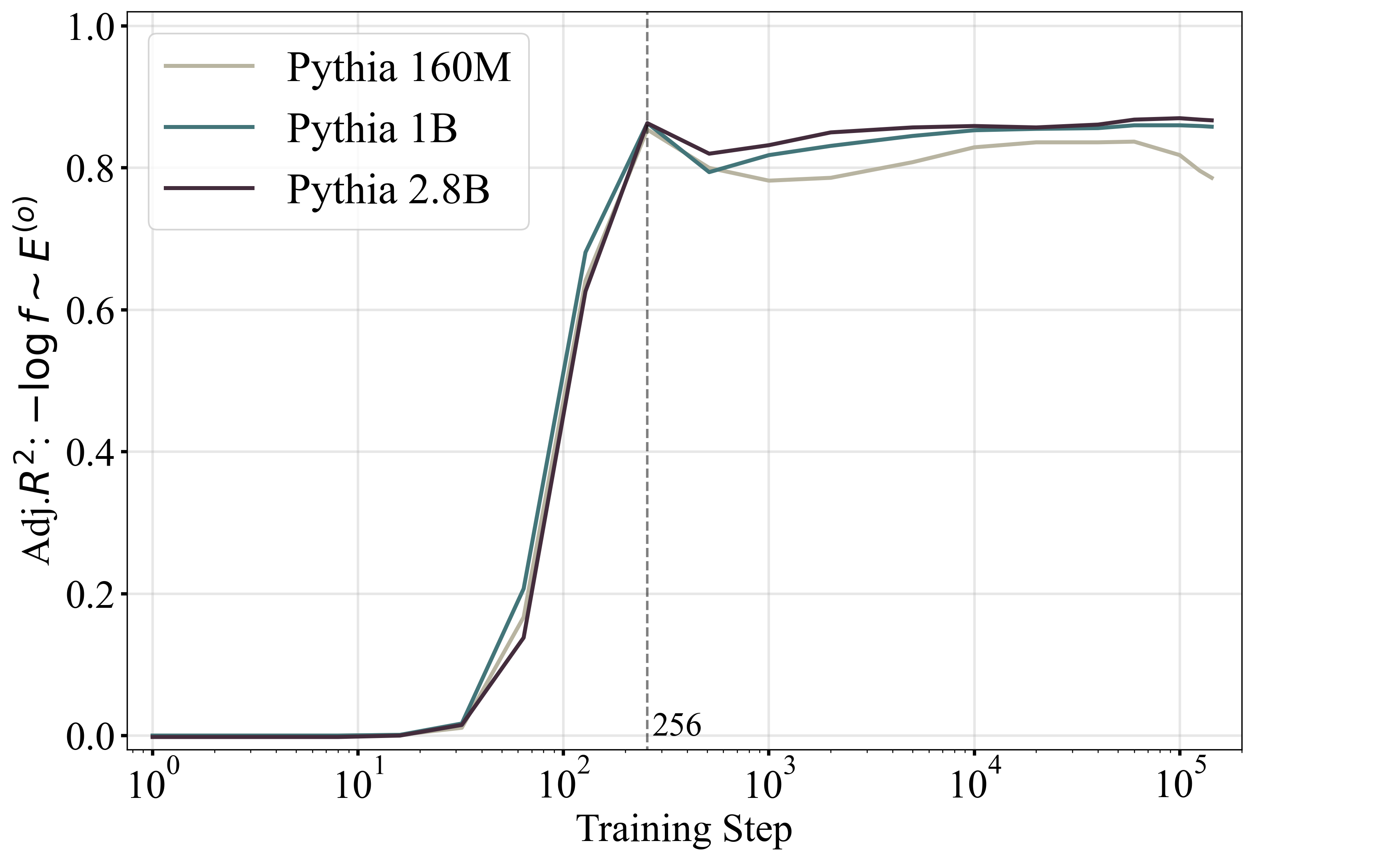}}    
\subfloat{
		\includegraphics[width=0.45\linewidth]{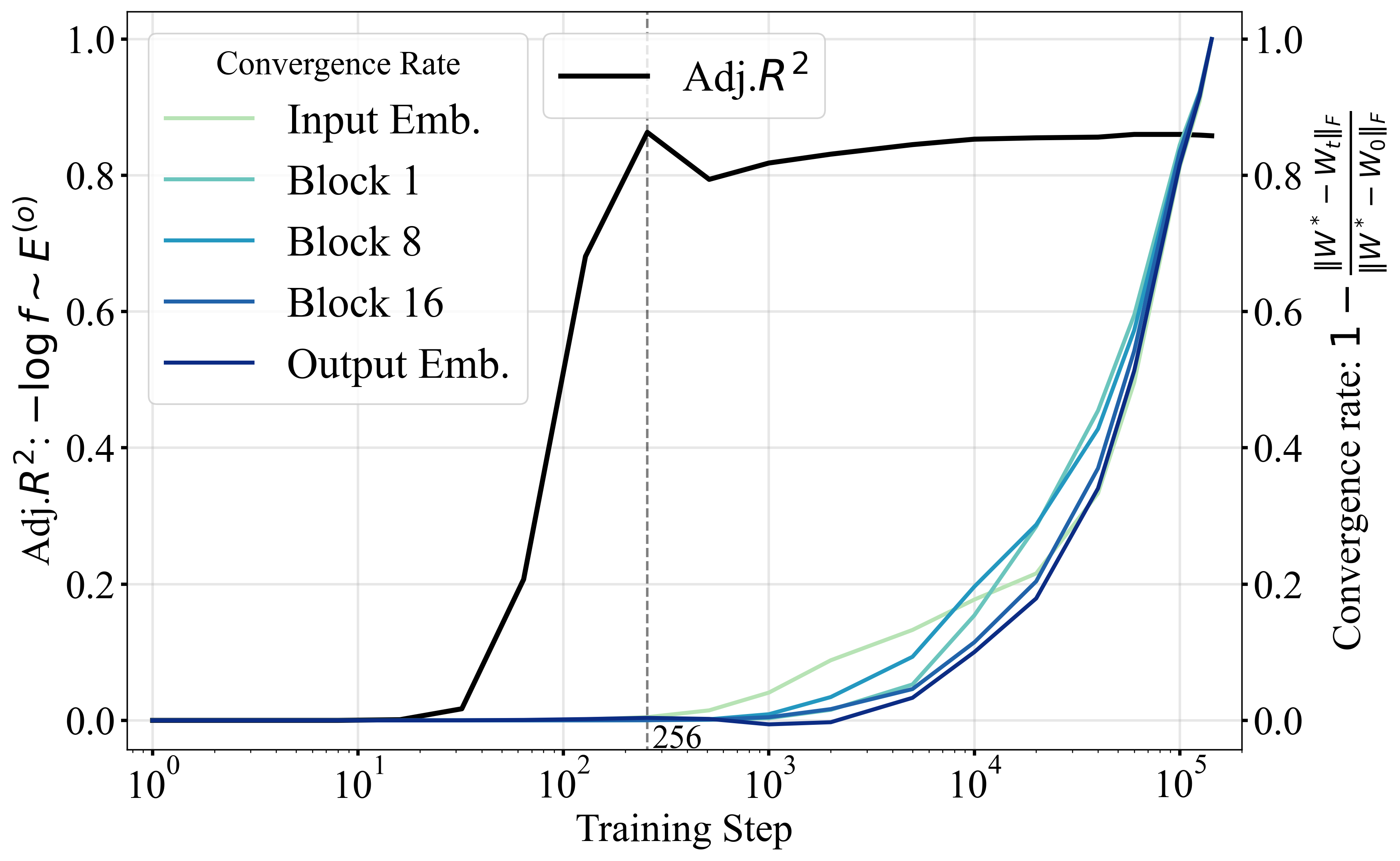}}
\caption{The training dynamics of Pythia. \textbf{Left}: the MLR goodness (adjusted $R^2$) of the negative logarithm of corpus frequency against the output embedding. \textbf{Right}: on the Pythia-1B, the MLR goodness in the left figure, and the convergence rate of various representative blocks. Notice the horizontal axis is logarithmically scaled.}
\label{fig:TrainD}
\end{figure*}

\paragraph{Wide-scale stable: Large-scaled probability steering is supported by a global log-linear pattern.} The correlation of actually steered scales of token probability against the expected scales is shown in Fig.~\ref{fig:editxy}. The steering remains accurate even on a large scale of up to 20x. This indicates that the algorithm and the log-linear encoding are wide-ranged, not only locally effective, i.e. the log-linear encoding is a widely stable common essence. 

\paragraph{Few-shot generalizable: Encoding remains distinct even by an $A_\mathcal{D}$ estimated by few-shot corpus.} Instead of using the 8192 examples for the detecting dataset $\mathcal{D}$, we try various numbers of examples to test the generalization of the encoding on GPT2. As shown in Fig.~\ref{fig:length} and~\ref{fig:appendix.S5}, we find that even 2 examples can produce a distinct averaged probability distribution for precise probability steering. This result strengthens the significance and generalization of our findings, that is, effective encodings can be detected in a small set of samples.

\vspace{0.5em}

These steering results demonstrate a clear causality between the log-linear encoding and the token probabilities, further than a simple statistical covariance relationship, and also demonstrate a wide-scale stability and generalization of our algorithm, while reflecting the same attributes of token probability encoding in output embedding. We can confirm that such log-linear encoding of token probabilities is an inherent attribute of output embedding, which \textbf{(1)} has wide-scale linearity, allowing a large-scaling probability steering, and \textbf{(2)} is common among tokens, so only a small number of samples are needed to find an accurate encoding. By our steering experiments, the properties of probability encoding are strengthened.

\section{Removing Dimensions with Weak Probability Encoding}
\label{sec:RDWE}

Moreover, based on the sparsity of probability encoding as shown in Fig.~\ref{fig:fewPCA}, we can infer that a large number of dimensions in the output embeddings are less effective for next token decoding. So we try to reduce the less related dimensions towards a lightweight output head.

\subsection{Method \& Experiment Settings}

First, we assign the aforementioned absolute MLR slopes as shown in (e-h) of Fig.~\ref{fig:fewPCA} as the importance (or, \textit{saliency}) to each dimension of output embedding. Then, we remove the dimensions in ascending order of such weight, i.e. we remove the identified less important dimension early. In detail, as a prototype setting in the laboratory, we only zero out the dimensions in the embedding matrix, while the dimensions of the attention mapping matrix, the feed-forward layer, and the dimensions of the last hidden state corresponding to the removed dimensions can also be removed equally. 

\paragraph{Experiment Settings.} We use the same MLR settings as \S\ref{sec:feqe} on GPT2, Pythia 2.8B, and GPT-J. As metrics for the causal language modeling quality, we test the MAUVE of the pruned model similarly to \S\ref{sec:Experimentsettings}, and test the KL divergence of the averaged probability distribution against the original model on the out-of-domain dataset mentioned before. As an adversarial controlling experiment, we remove dimensions in \textit{descending} order of importance, i.e. we remove the more important dimension early, reverse to the original settings.

\subsection{Result}

The results are shown in Fig.~\ref{fig:del}. With the ascending removing order (the positive experiment, solid curve), at the beginning of the removal, both metrics deteriorate slightly until around 50\% of the dimensions are removed. Regarding $\mathrm{MAUVE}>0.80$ as a threshold, we can confirm that $30\%\sim40\%$ of the dimensions in the output embedding can be removed without significant impairment of the causal language modeling quality of LMs. In contrast, in the adversarial experiment with a descending removing style (dashed curve), both metrics deteriorate sharply at the beginning of removing. Such results suggest that the weights from the MLR are faithful importance metrics.

Noticeably, our method also works on the tied models (Fig.~\ref{fig:del} (a)), i.e.~removing unnecessary dimensions concurrently from the input embedding and the output embedding, the metrics remain at a considerable level. 

Consistent with previous works~\cite{kovaleva2021bert, timkey2021all, gordon2020compressing}, our results identify the anisotropy in the importance among the dimensions of the output embedding. Moreover, we confirm by dimensional ablation that the MLR results of the log-linear correlation are a good metric of importance, i.e. saliency score. This can be a new paradigm of saliency if a direct one-step statistical model (like MLR in our paper) can be estimated between the features and the model output.

\section{Output Embedding Learns Corpus Frequency in Early Training Steps} 
\label{sec:TrainD}

Additionally, the findings in this paper inspire us to utilize such log-linear correlation for detecting the encoding of \textit{corpus token frequency} in the output embedding during the pre-training of LMs, for a closer observation of the pre-training dynamics. Intuitively, since the model can refract overall output probability distribution in the output embedding matrix, it should be forced to produce the same output distribution with the corpus token frequency by the training objective. Therefore, when a log-linear encoding of \textit{token frequency} of the training corpus is observed in the output embedding, we can confirm that the output embedding learns the token frequency information of the corpus.


We use the Pythia suite~\cite{biderman2023pythia}, where sequences of the model intermediate training checkpoints are accessible. We estimate the \textit{token frequencies} of the pre-training corpus \textsc{Pile}~\cite{pile1, pile2} by sampling 14.3B tokens, then conduct MLR on the negative logarithm of the estimated \textbf{token frequencies} w.r.t. the output embeddings on each training checkpoint, with various training steps. The results of MLR fitting goodness against training steps are shown in the left part of Fig.~\ref{fig:TrainD}, where we can confirm an effective encoding since the very early steps of the training process, and larger models have slightly better fitting goodness but almost no difference in the timing of emergence.

Moreover, we want to know whether such a phenomenon is a subsidiary or subsequent effect of the convergence of the parameters. We use the convergence rate following the \citet{chen2022layer} to describe the actual training completion: denoting the \textit{trained} parameter matrix as $\theta^*$, the \textit{initialized} parameter matrix as $\theta_0$, and the parameter matrix \textit{at step $t$} as $\theta_t$, the convergence rate at step $t$ can be written as $(1 - \Vert \theta^*-\theta_t \Vert_F/\Vert \theta^*-\theta_0 \Vert_F)$. We visualize the convergence rate of the input and output embeddings and the query-key-value mapping matrix of the multi-head attention blocks on Pythia-1B, as shown in the right part of Fig.~\ref{fig:TrainD}. We find that the log-linear correlation occurs even earlier than an obvious convergence trend is observed by the convergence rate, and the appearance of such correlation happens to be the starting point of the convergence process of the model. We infer that the output embedding should learn the coarse-grained output pattern earlier than the semantics details.

Additionally, we initially find that each layer of the transformer appears to have a uniform convergence curve, instead of an obvious deeper-slower pattern found by~\citet{chen2022layer}. Especially, the input embedding and output embedding have almost overlapping training curves. We speculate that this is the effect of such full-residual connection networks, which makes the layering of the network inconspicuous during the gradient descent.

\section{Conclusion and Discussion}

\paragraph{Conclusion.} In this paper, based on the observation of a linear-like correlation between the output token probability and the output token embeddings shown in Fig.~\ref{fig:PCA2d-fre}, we derive an approximate log-linear correlation from the nature of $\mathrm{softmax}$-based output head with a large output space and concentrated output value, i.e. the output token probabilities are encoded in a common direction of output embeddings. To test the causality of such correlation, we steer the token probabilities along such encoding direction with high accuracy, stability, and generalizability. Then, based on the sparsity of the encoding, we can distinguish the contributions of dimensions for the output probability of the model, which allows us to remove the determined non-contributing dimensions, and no critical deterioration is found. Finally, based on the findings, we find that the LMs catch the token frequency in training data at very early steps in the training process log-linearly, even earlier than an obvious convergence trend is observed. This paper reveals the inner mechanism of LM heads on the causal language modeling task and helps understand the global principles and training dynamics of LMs.

\paragraph{Comparing to previous works.} Previous work about analyzing LM heads was conducted by~\citet{kobayashi2023transformer}, where a correlation between the output token probability and bias term in the LM head was found. They declared that the bias term is a projection to extract the probability from the output embedding, but almost no more discussion about the embedding matrix, which is the major component of the LM head. Making up for their work is one of the original motivations for our work. More discussions about related works can be found in \S\ref{sec:appendix.RW}.

\paragraph{Demonstrations towards application.} Our output probabilities steering algorithm on the LM head reveals the possibility of model steering on only the output probability rather than in the hidden states or the lower layers~\cite{dai2022knowledge, meng2022locating}, which is more concise, and easy to explain. Some global toxicity generation~\cite{gehman-etal-2020-realtoxicityprompts}, or biases in some application scenario~\cite{fei2023mitigating} can be suppressed by such a method, but as we will mention in the Limitations, it is elegant but not engineering-oriented. Moreover, the dimension compression method in this paper can be an easy-to-use and harmless inference-time acceleration. Notice that we have supervision on such dimension compression by the MLR slopes, so it can be more accurate than the previous random or unsupervised pruning~\cite{gordon2020compressing}.

\paragraph{Towards a new saliency score of output embeddings.} Saliency score~\cite{saliency_survey1, saliency_survey2} is to weight a component (feature or parameter)
in a deep learning model by its contribution. In this paper, we find that the log-linear token probability encoding works like a saliency score towards the output embedding, and build a log-MLR model to assign saliency scores to the parameters. Such a posterior statistical method can be a new paradigm of model-based saliency~\cite{dabkowski2017real}, if one-step correlations can be detected between the output and components of the neural network, a closed-form saliency model can be proposed instead of a universal statistical model.


\section{Related Works}
\label{sec:appendix.RW}

As mentioned before,~\citet{kobayashi2023transformer} mainly found a correlation between the output token probability and the bias term in the LM head and tried to remove this bias towards more diversified text generation. However, they didn't analyze the output embedding matrix, which has the most parameters in the LM head, and this paper completes their research.

\paragraph{Geometry of Input Embedding.} As a similar research object with the output embedding, it was found that the word embeddings in LMs, as well as the hidden states, are anisotropy~\cite{mu2018all, ethayarajh2019contextual, gao2018representation}, i.e., these vectors share a common radial bias. Such anisotropies hurt the expressiveness of word embeddings, and the word frequency in the corpus may be an inducement~\cite{mu2018all, valentini2023investigating}. Also, some efforts tried to remove the harmfulness of anisotropies and towards isotropy word embeddings~\cite{mu2018all, gong2018frage}. These works are based on input embedding, while our paper is on output embedding. Although we can confirm that the input and output embeddings act similarly, they are still completely different components of untied LMs. So the existing conclusions on input embeddings cannot overwrite our work.

\paragraph{Embedding Tying in LMs.} LMs in previous generations often have a shared output embedding from the input embedding, such as BERT~\cite{kenton2019bert}, RoBERTa~\cite{liu2019roberta}, GPT2~\cite{radford2019language}, etc. That is, the LM head maps the last hidden state to the token probability by dpt-producing the input embedding. Such a paradigm is recommended by~\citet{press2017using} for fewer parameters. And also refuted by~\citet{chung2020rethinking} for a harmfulness to expressiveness. Such a paradigm is being deprecated currently, but the model behavior with and without embedding tying is still interesting to analyze.

\paragraph{(Language) Model Editing and Model Pruning.} Recent Large LMs are expensive to fine-tune or retrain, so there are many model editing methods to control the output of LMs~\cite{yao2023editing}. Current LM parameter editing methods are mainly oriented to entity relationship editing, where they locate some parameters with correlations to the entities, and interference is applied on such parameters~\cite{dai2022knowledge, meng2022locating}. Moreover, vectorized methods are also proposed with the arithmetic of parameter vectors with editing information~\cite{ilharco2022editing, ortiz2024task}. As a specific and extreme scenario of model editing, research on \textit{model pruning}, similar to our dimension removing is also proposed in current years~\cite{zhu2023survey, frantar2023sparsegpt, kovaleva2021bert, timkey2021all, gordon2020compressing}. These pruning are usually unsupervised, where our dimension removing can be a new practice in the supervised pruning paradigm.

\paragraph{Training Dynamics (of LMs).} Investigating what is happening in the training process of language models and other deep learning models is an attractive research topic. Many works about training trajectory~\cite{kalra2024phase, jastrzkebski2018relation, lewkowycz2020large}, early period training behaviors~\cite{frankle2019early, kalra2024phase, achille2018critical}, loss landscape~\cite{neyshabur2020being, keskar2017large, li2018visualizing}, and ``knowledge'' earned in different stages of training~\cite{tirumala2022memorization, liu2021probing} have been done. Differences in training speed among various network layers (the deeper-slower pattern) have been discovered by~\citet{chen2022layer}, while in this paper, following their method, we don't find a similar pattern.

\section{Limitations}

Although we declare that the probability steering algorithm proposed in \S\ref{sec:editing} is only an experimental method for investigation, we acknowledge that it is elegant but not practical. It can never be faster, more accurate, and more harmless than a filter on the output head \cite{guo2017calibration}. Future works can be focused on a \textit{local} or \textit{directional} probability editing method, limiting the detecting dataset $\mathcal{D}$, and only editing the probability on specific input prefix to resolve a controlled-generation task.

The dimension-reducing method in \S\ref{sec:RDWE} may lead LMs to be unavailable on other tasks depending on the last hidden state, such as sentence summarizing vectors encoding, etc. However, we can always keep the original checkpoint to restore these additional abilities of LM heads easily. 

Furthermore, despite our efforts, we cannot confirm the source of the sparsity of the probability encoding mentioned in Fig.~\ref{fig:fewPCA}. Future works can be focused on the detailed training dynamics to trace such a sparsity.

The findings in this paper seriously depend on the properties of the last hidden state of LMs. Although the layer normalization~\cite{ba2016layer, vaswani2017attention} in current Transformer-based LMs provides some intuitive assurance for the stability and consistency of the last hidden state, further discussion is still needed to confirm the homogeneity or heterogeneity of the models' intrinsic properties to explain the differences between different models in the token probability encoding phenomena investigated in the paper (e.g. the reason of our method perform better on GPT-J than GPT2-XL in Table~\ref{table:edit_res}, or, the reason of the sparsity of GPT2-XL is weaker than all the models we investigated in Fig.~\ref{fig:fewPCA}), to establish a connection with the essential properties of LMs. Also, we should examine the distribution of the last hidden state so that the output probability to find how the accuracy of the averaged output probabilities can reflect the individual output probability.

\section*{Acknowledgements}

This work is supported by the Nakajima Foundation.

\noindent The authors would like to thank Mr. Yuxuan Wang at Beijing Institute of Technology for his proofreading and constructive criticism.

\newpage
\bibliography{custom}

\appendix

\begin{table*}[t]
\centering
\caption{Supplementary results of Table~\ref{table:edit_res} with various $b$ on GPT2.}
\label{tab:appendix.1}
\begin{tabular}{ccccccc}
\toprule
\multicolumn{2}{c}{\multirow{2}{*}{ }} & \textbf{Reliability} & \multicolumn{2}{c}{\textbf{Generalization}} & \multicolumn{2}{c}{\textbf{Specificity}} \\ \cmidrule(lr){3-3} \cmidrule(lr){4-5} \cmidrule(l){6-7}
\multicolumn{2}{l}{} & $e_{local}\downarrow$ & $e_{id}\downarrow$ & $e_{ood}\downarrow$ & $d_{KL}^r\times 10^{-7}\downarrow$ & \textsc{MAUVE}$\uparrow$  \\  \midrule
\multicolumn{7}{c}{- \textit{Baseline Values} -} \\
\multicolumn{2}{c}{\textbf{unedited}} & ${1.09}_{1.06}$ & ${1.09}_{1.06}$ & ${1.09}_{1.06}$ & ${0.00}_{0.00}$ & ${1.00}_{0.00}$ \\
\multicolumn{2}{c}{\textbf{random}} & ${1.33}_{1.28}$ & ${1.33}_{1.28}$ & ${1.29}_{1.24}$ & ${2.31}_{1.02}$ & ${0.96}_{0.01}$\\
\multicolumn{2}{c}{\textbf{137M, shuffled}} & ${1.18}_{1.07}$ & ${1.18}_{1.08}$ & ${1.18}_{1.08}$ & ${1.64}_{0.48}$ & ${0.96}_{0.01}$\\\midrule
\multicolumn{7}{c}{- \textit{Experimental Values} -} \\
\textbf{Average} & ($b=-\infty$) & ${2.79}_{3.21}$ & ${2.86}_{3.24}$ & ${2.61}_{3.12}$ & ${4.00\times10^3}$ & ${0.92}_{0.08}$ \\
\multirow{3}{*}{\textbf{Softmax}} & $b=1$ & ${0.17}_{0.29}$ & $\mathbf{0.17}_{0.31}$ & ${0.18}_{0.33}$ & ${11.97}_{8.22}$ & ${0.95}_{0.02}$ \\
 & $b=2$ & ${0.19}_{0.30}$ & ${0.20}_{0.32}$ & $\mathbf{0.15}_{0.28}$ & $\mathbf{9.51}_{5.87}$ & $\mathbf{0.96}_{0.01}$ \\
 & $b=5$ & $\mathbf{0.17}_{0.28}$ & ${0.18}_{0.31}$ & ${0.16}_{0.28}$ & ${12.25}_{7.91}$ & ${0.95}_{0.01}$  \\
\textbf{Argmax} & ($b=+\infty$) & ${1.21}_{1.68}$ & ${1.27}_{1.70}$ & ${1.30}_{1.84}$ & ${9.29\times10^3}$ & ${0.92}_{0.14}$ \\ \bottomrule
\end{tabular}
\end{table*}

\section{Calculation Details}
\subsection{Calculation of Averaged Token Probability}
\label{sec:appendix.ATF}

Given a dataset $\mathcal{D}=\{x_i\}_{i=1}^{m}$, we input each $x_i$ into LMs in a teacher forcing style. Denote the length of $x_i$ as $l_i$, we can get output token probability distribution on each time step (noted as $j$) $\alpha_{\mathcal{D},\theta,i,j}\in\mathbb{R}^{|\mathbb{V}|}$ of an amount of $l_i$.

We average all the $\alpha_{\mathcal{D},\theta, i,j}$ on every $i$ and $j$, and get the averaged token probability distribution $\alpha_{\mathcal{D}, \theta}$ on dataset $\mathcal{D}$.

\subsection{Error Analysis of Eq.~\ref{eq.first}}
\label{sec:appendix.Error}

In Eq.~\ref{eq.first}, we do a local linear approximation as:
\begin{equation*}
\begin{split}
    -\log \alpha_{w,\mathcal{D}, \theta} &= -\log \E\left[\mathrm{P}_\theta(w|x)\right] \\
    & \approx \E\left[-\log \left[\mathrm{P}_\theta(w|x)\right]\right]. \\
\end{split}
\end{equation*}
That is, given a set of $\{p_i\}_{i=0}^n$, where $\forall i, p_i > 0$ we approximate that $\log \mathbb{E}_{i\in[0,n]}[p_i] \approx \mathbb{E}_{i\in[0,n]}[\log p_i]$. Assume that we have a non-descending sequence of $p$, that is, $p_0\leqslant p_1\leqslant p_2\leqslant \dots \leqslant p_n$, w.l.o.g. We can confirm that $\mathbb{E}_{i\in[0,n]}[p_i]\in [p_0, p_n]$. So we have: $\exists \xi \in (p_0, \mathbb{E}_{i\in[0,n]}[p_i]), s.t.$,
\begin{equation*}
    \log \mathbb{E}_{i\in[0,n]}[p_i] = \frac{1}{\xi}(\mathbb{E}_{i\in[0,n]}[p_i]-p_0) + \log p_0
\end{equation*}
We have:
\begin{equation*}
    \mathbb{E}_{i\in[0,n]}[\log p_i] \geqslant \log p_0,
\end{equation*}
\noindent since $\log'(\cdot) > 0$. That is:
\begin{equation}
\begin{split}
    & [\log \mathbb{E}_{i\in[0,n]}[p_i] - \mathbb{E}_{i\in[0,n]}[\log p_i]]^2 \\
    \leqslant &  \frac{1}{\xi^2}\left(\mathbb{E}_{i\in[0,n]}[p_i]-p_0\right)^2 \\
    \leqslant & \frac{1}{p_0^2}\left(\mathbb{E}_{i\in[0,n]}[p_i]-p_0\right)^2.
\end{split}
\label{eq:error}
\end{equation}

\begin{figure}[t]
\centering
\subfloat[GPT2, 137M]{
		\includegraphics[width=0.49\linewidth]{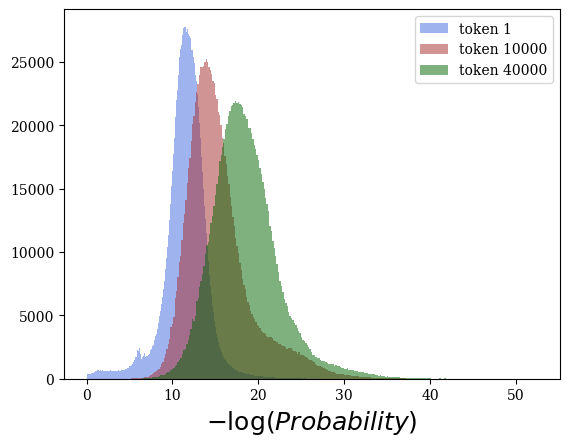}}
\subfloat[GPT2-XL, 1.6B]{
		\includegraphics[width=0.49\linewidth]{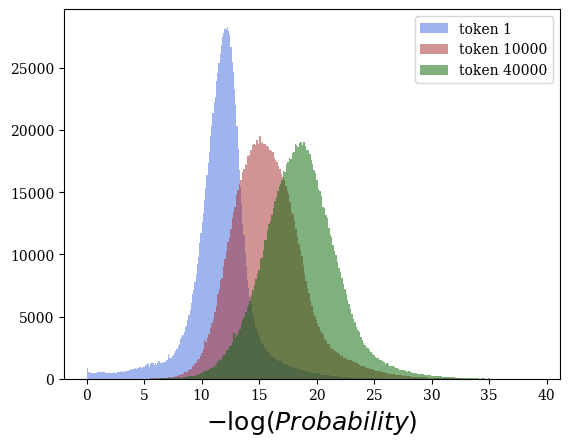}}
\newline
\subfloat[Pythia, 2.8B]{
		\includegraphics[width=0.49\linewidth]{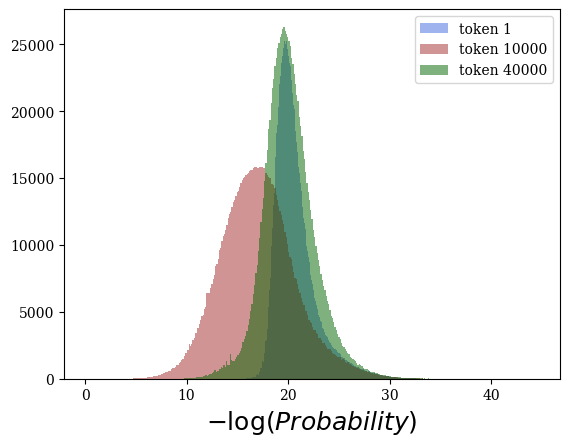}}
\subfloat[GPT-J, 6B]{
		\includegraphics[width=0.49\linewidth]{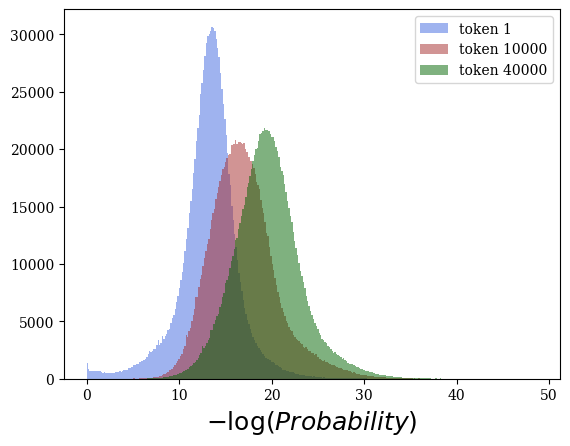}}
\caption{Examples of the probability distribution of one token among input prefixes.}
\label{fig:appendix.ErrorAna}
\end{figure}

We can empirically confirm the concentration of $p_i$, shown as examples in Fig.~\ref{fig:appendix.ErrorAna}, which makes the error shown in Eq.~\ref{eq:error} acceptably small. Additionally, a low-probability token shows a wide probability distribution, which is consistent with our findings in Fig.~\ref{fig:fit}, where a low-probability token is assigned with more inaccurate predictions (manifested as a comet-shaped figure).

\begin{figure}[t]
    \centering
    \includegraphics[width=0.8\linewidth]{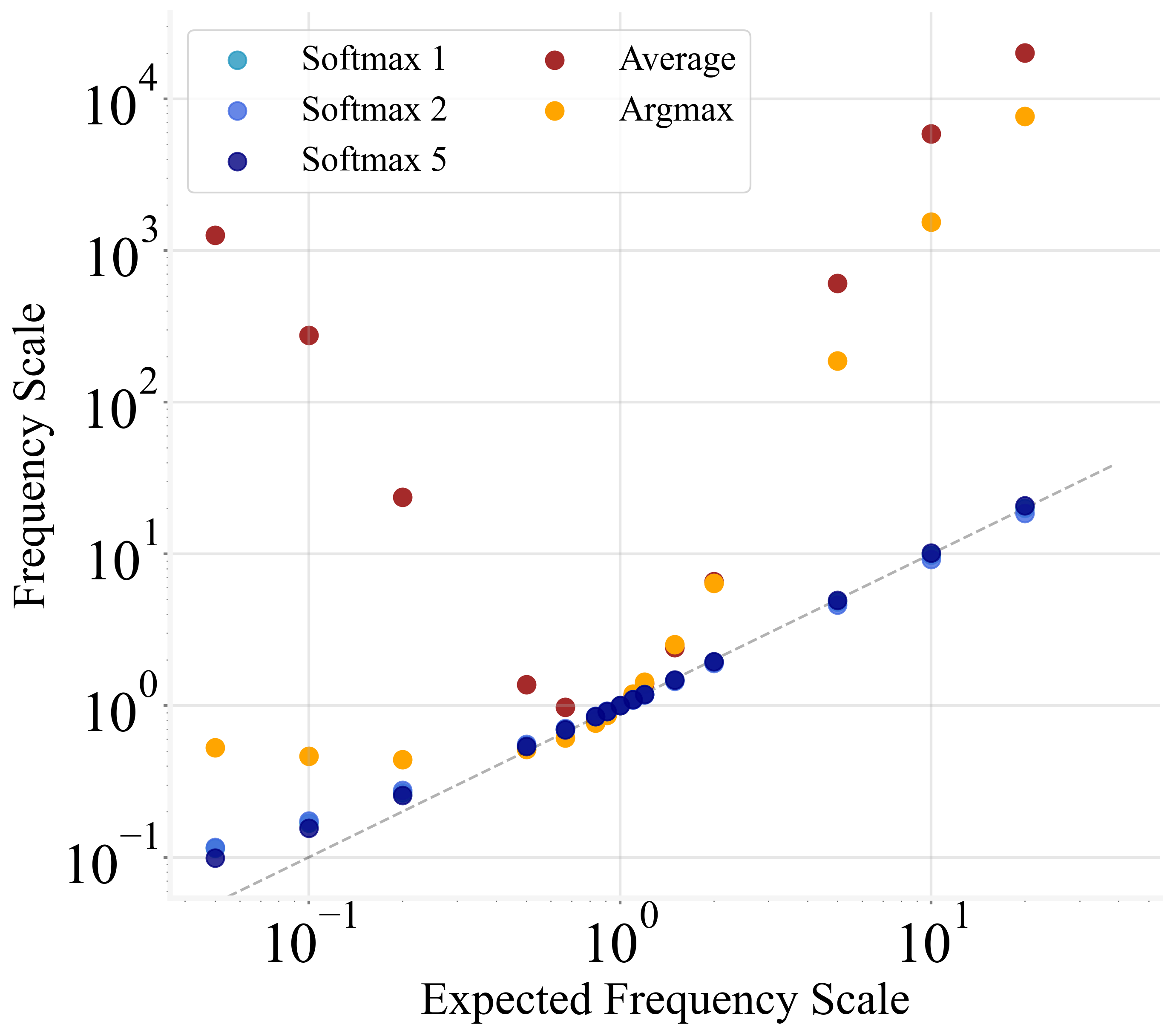}
    \caption{The expected probability scales and the actually steered scales measured in the edited LMs w.r.t. different values of $b$.}
    \label{fig:appendix.X-Y-A}
\end{figure}

\begin{figure}[t]
\centering
\subfloat[GPT2-XL, 1.6B]{
		\includegraphics[width=0.48\linewidth]{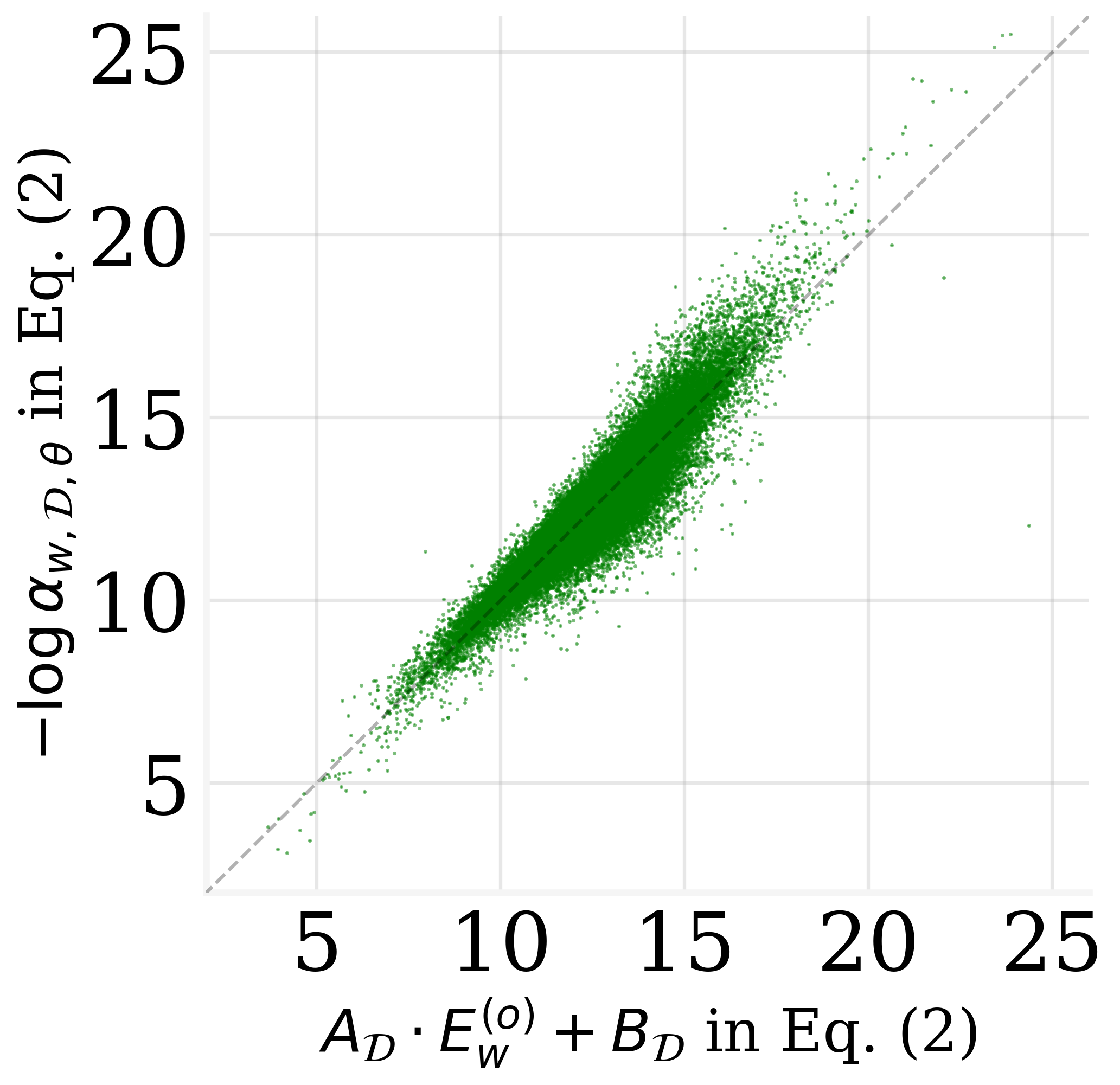}}
\subfloat[Pythia, 2.8B]{
		\includegraphics[width=0.48\linewidth]{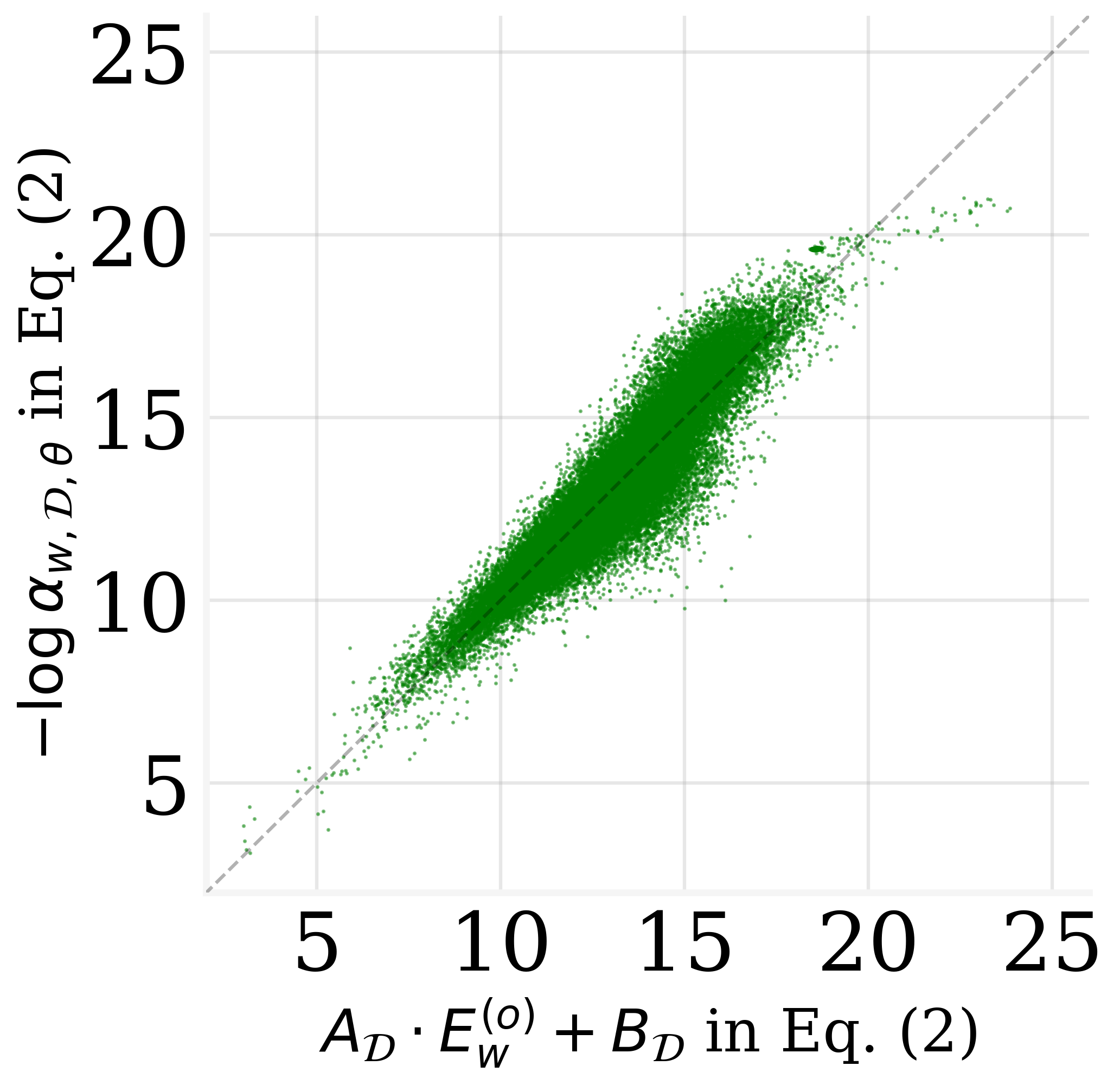}}
\caption{Supplement for Fig.~\ref{fig:fit}. The MLR results on GPT2-XL and Pythia-2.8B.}
\label{fig:appendix.S2}
\end{figure}

\begin{figure}[t]
    \centering
    \includegraphics[width=0.6\linewidth]{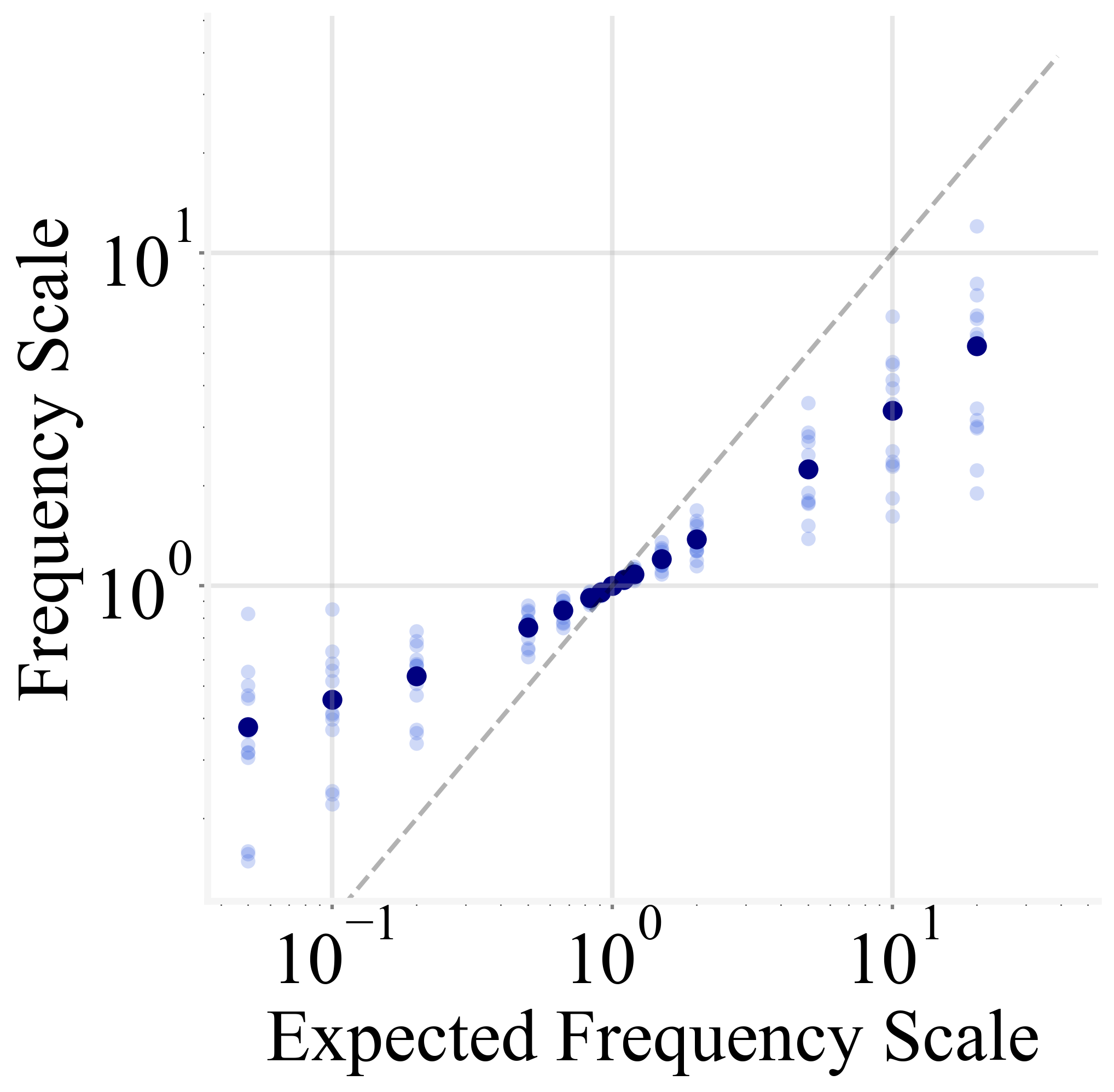}
    \caption{Supplement for Fig.~\ref{fig:editxy}. The expected probability scales and the actually steered scales measured in the steered GPT2-XL.}
    \label{fig:appendix.S4}
\end{figure}

\begin{figure*}[t]
\centering
\subfloat[GPT2-XL, 1.6B]{
		\includegraphics[width=0.32\linewidth]{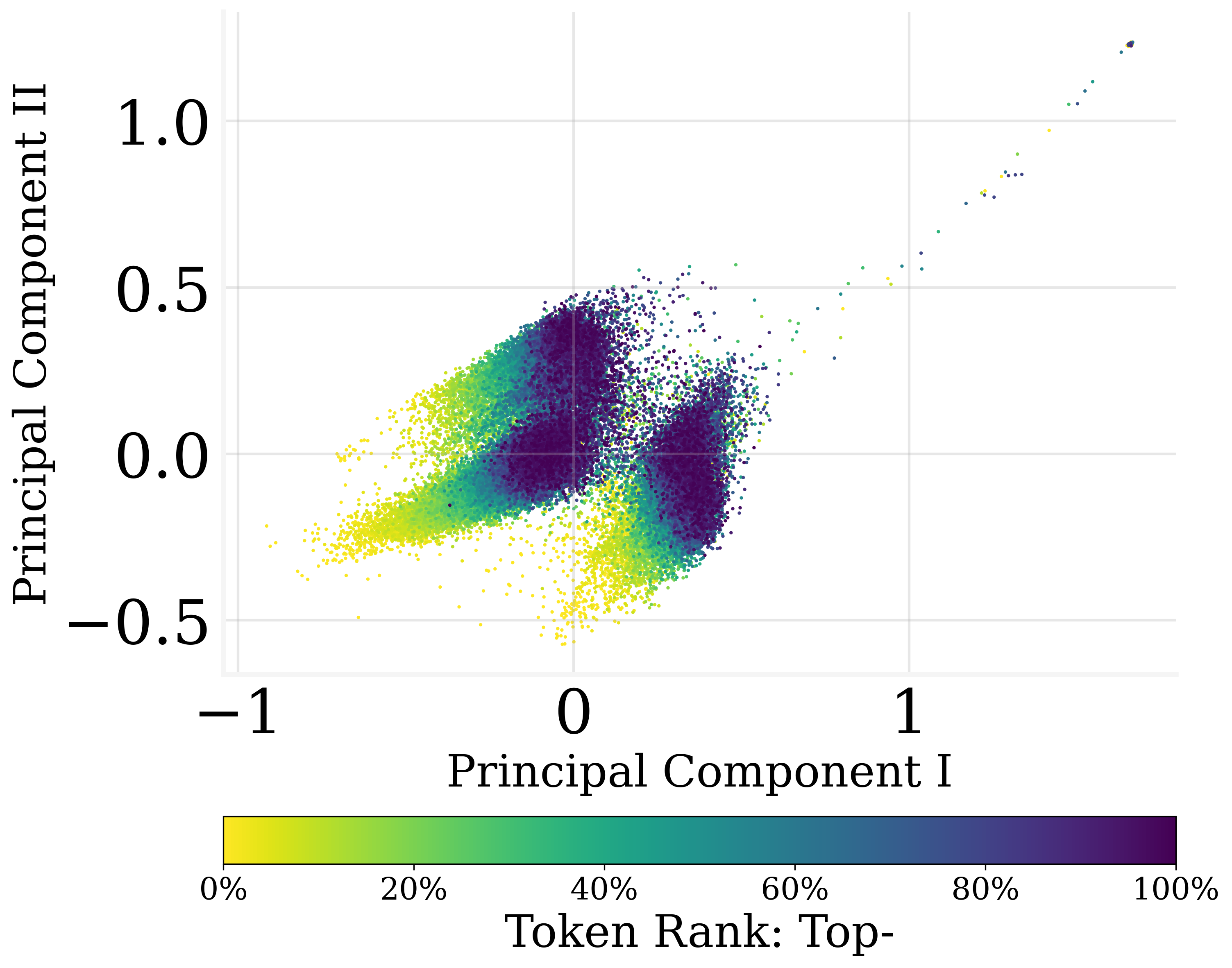}}
\subfloat[Pythia, 2.8B]{
		\includegraphics[width=0.32\linewidth]{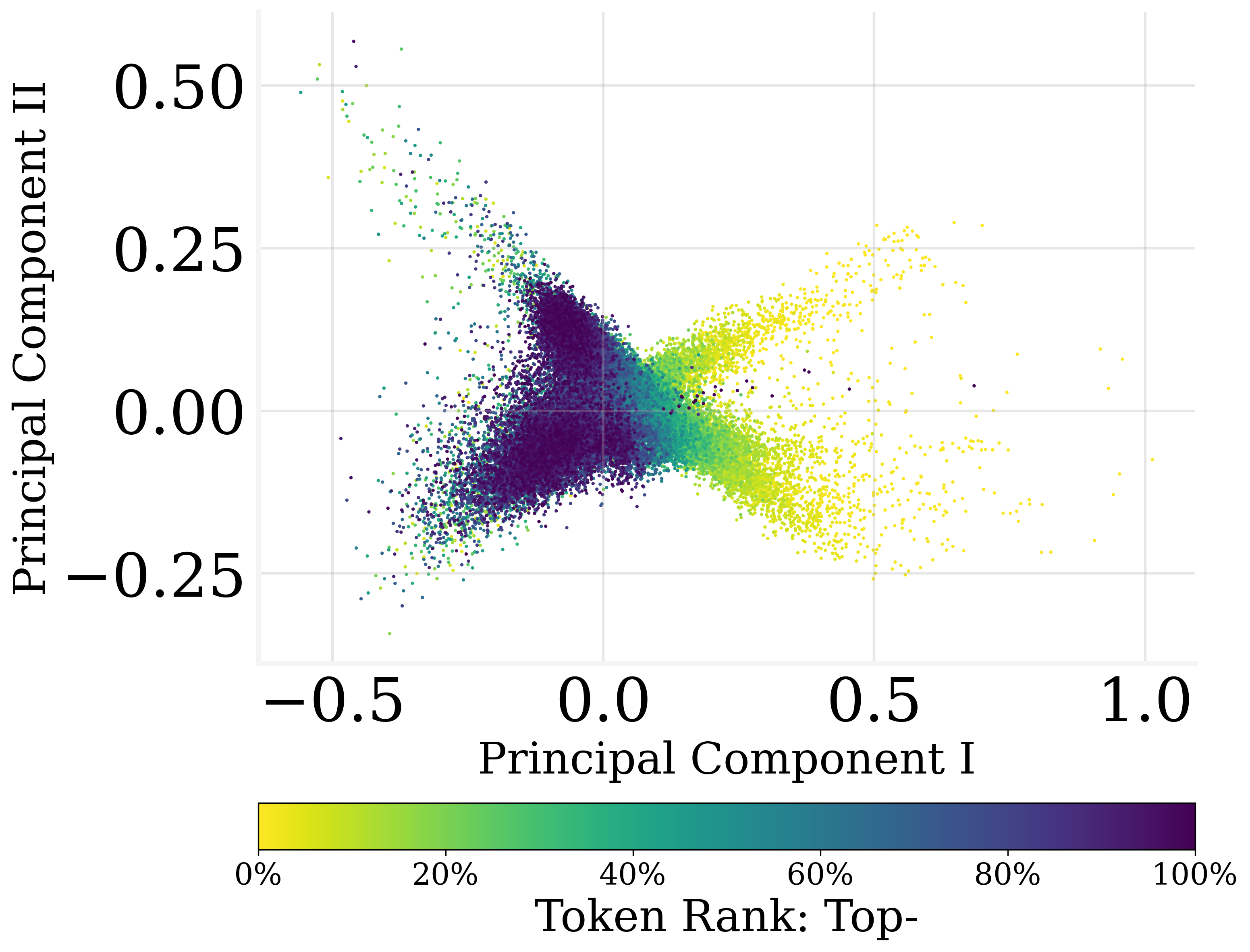}}
\subfloat[GPT-J, 6B]{
		\includegraphics[width=0.32\linewidth]{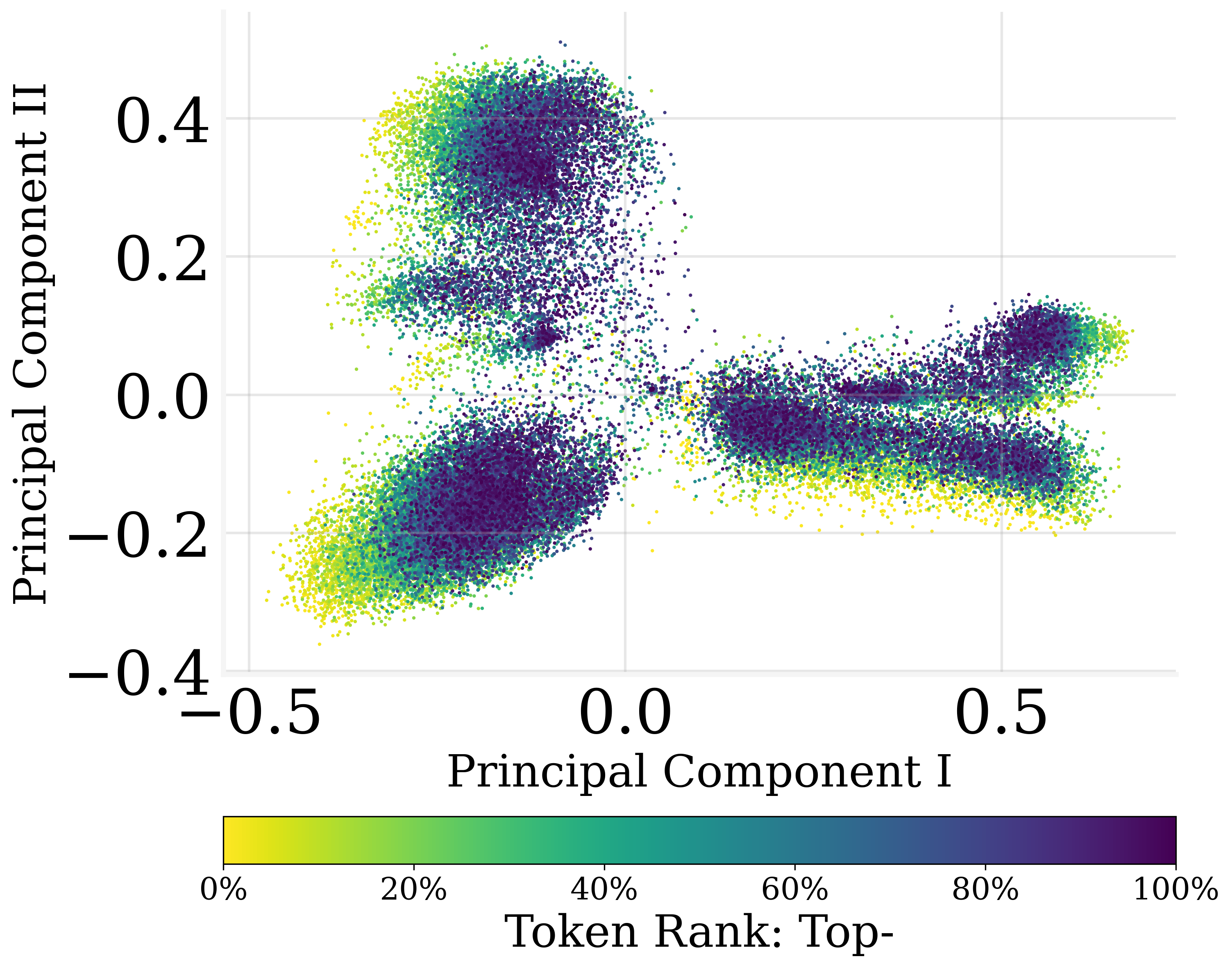}}
\caption{Supplement for Fig.~\ref{fig:PCA2d-fre}. Output embedding visualizations w.r.t. output token probability for GPT2-XL, Pythia, GPT-J.}
\label{fig:appendix.S1}
\end{figure*}

\begin{figure*}[t]
\centering
\subfloat[$e_\mathrm{local}$]{
		\includegraphics[width=0.19\linewidth]{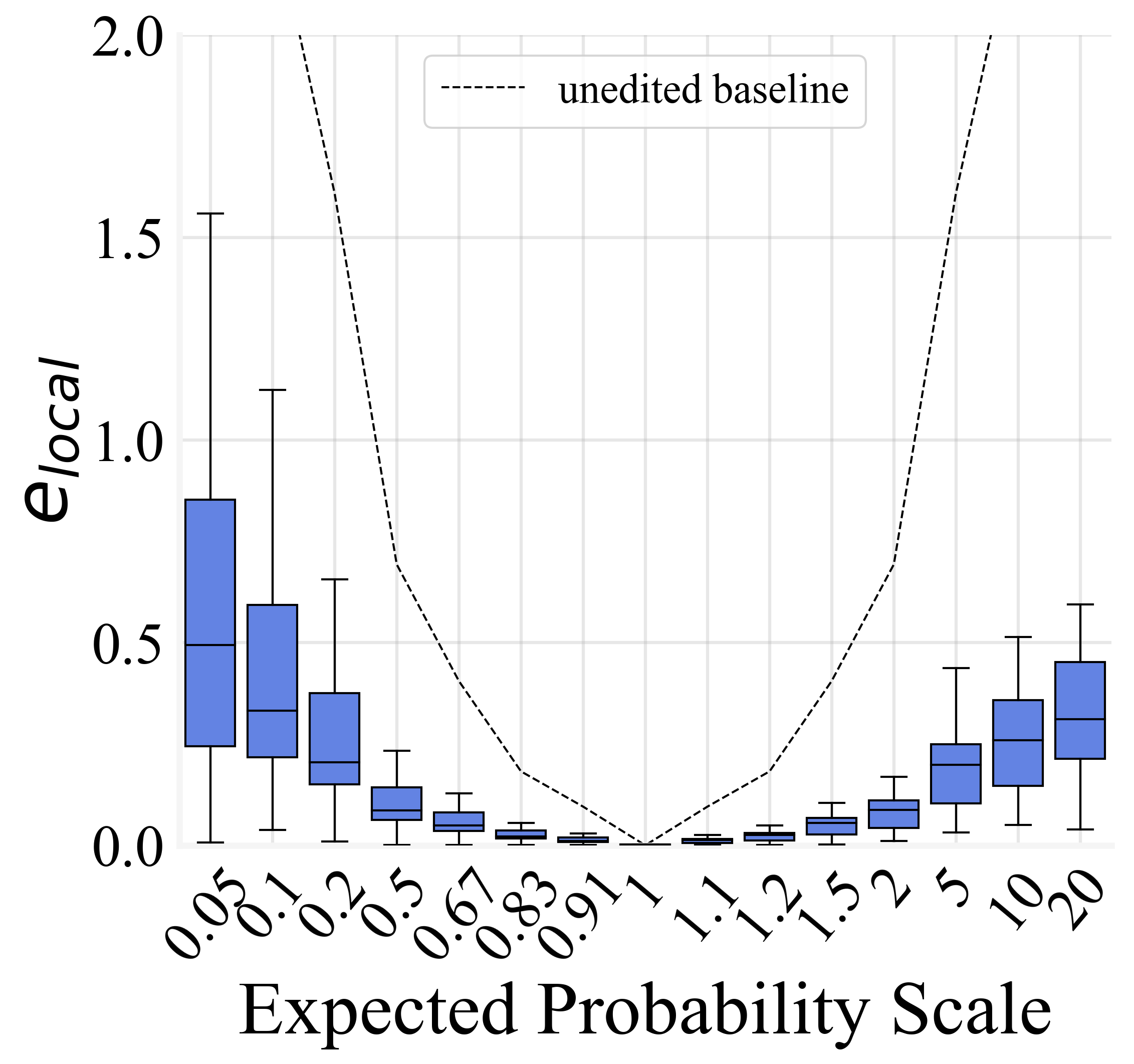}}
\subfloat[$e_\mathrm{id}$]{
		\includegraphics[width=0.19\linewidth]{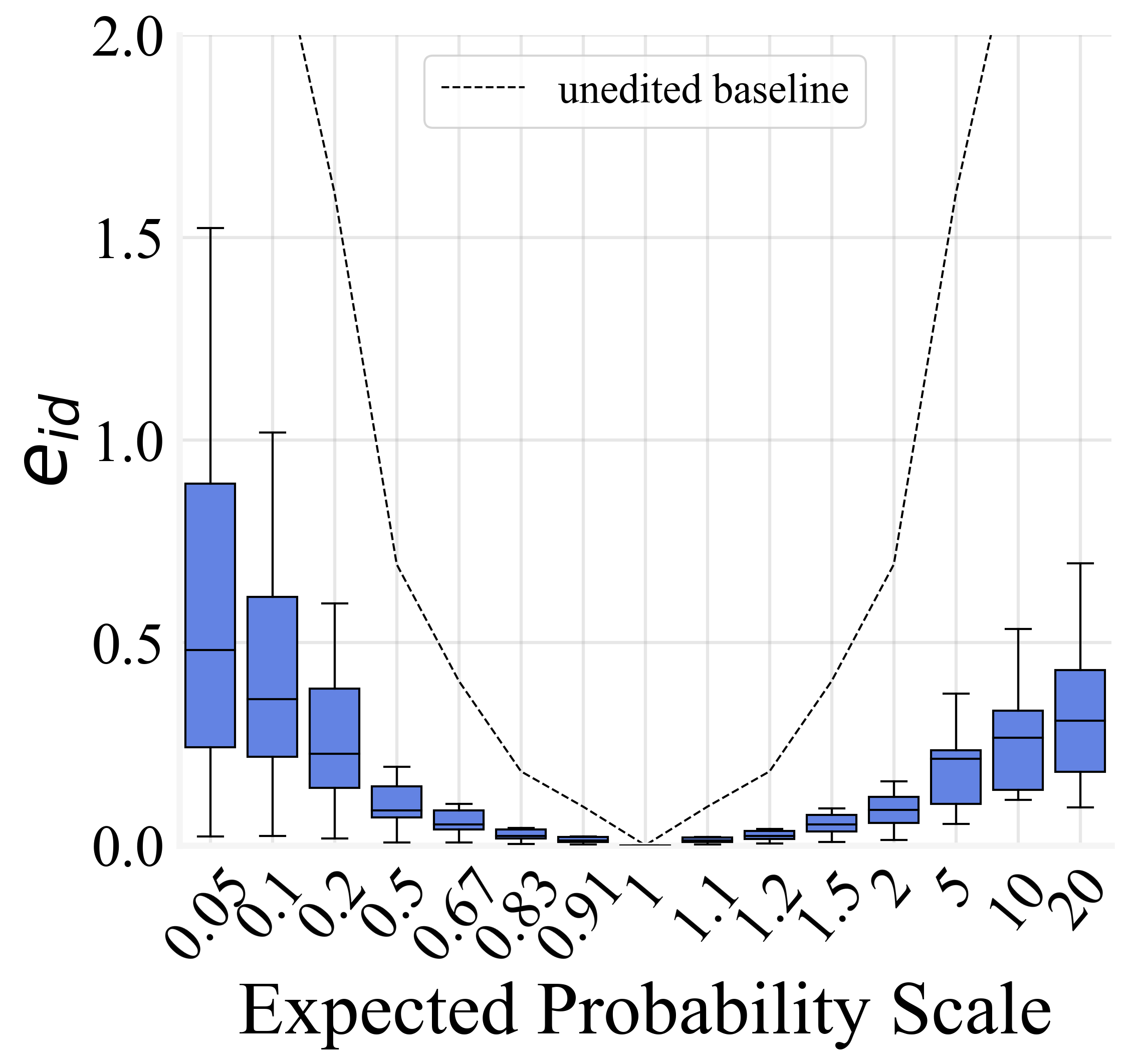}}
\subfloat[$e_\mathrm{ood}$]{
		\includegraphics[width=0.19\linewidth]{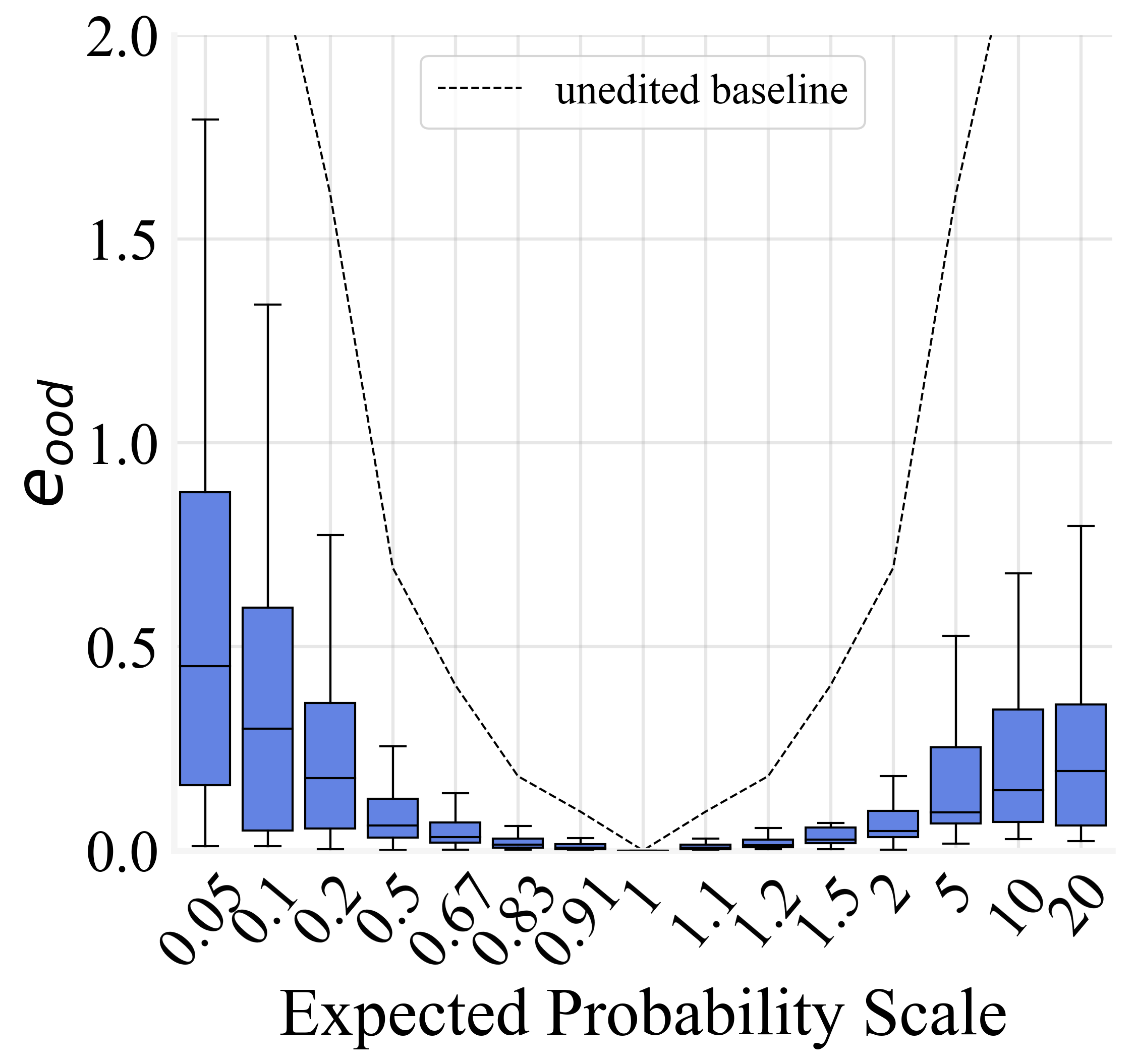}}
\subfloat[$d_\mathrm{KL}^r$]{
		\includegraphics[width=0.19\linewidth]{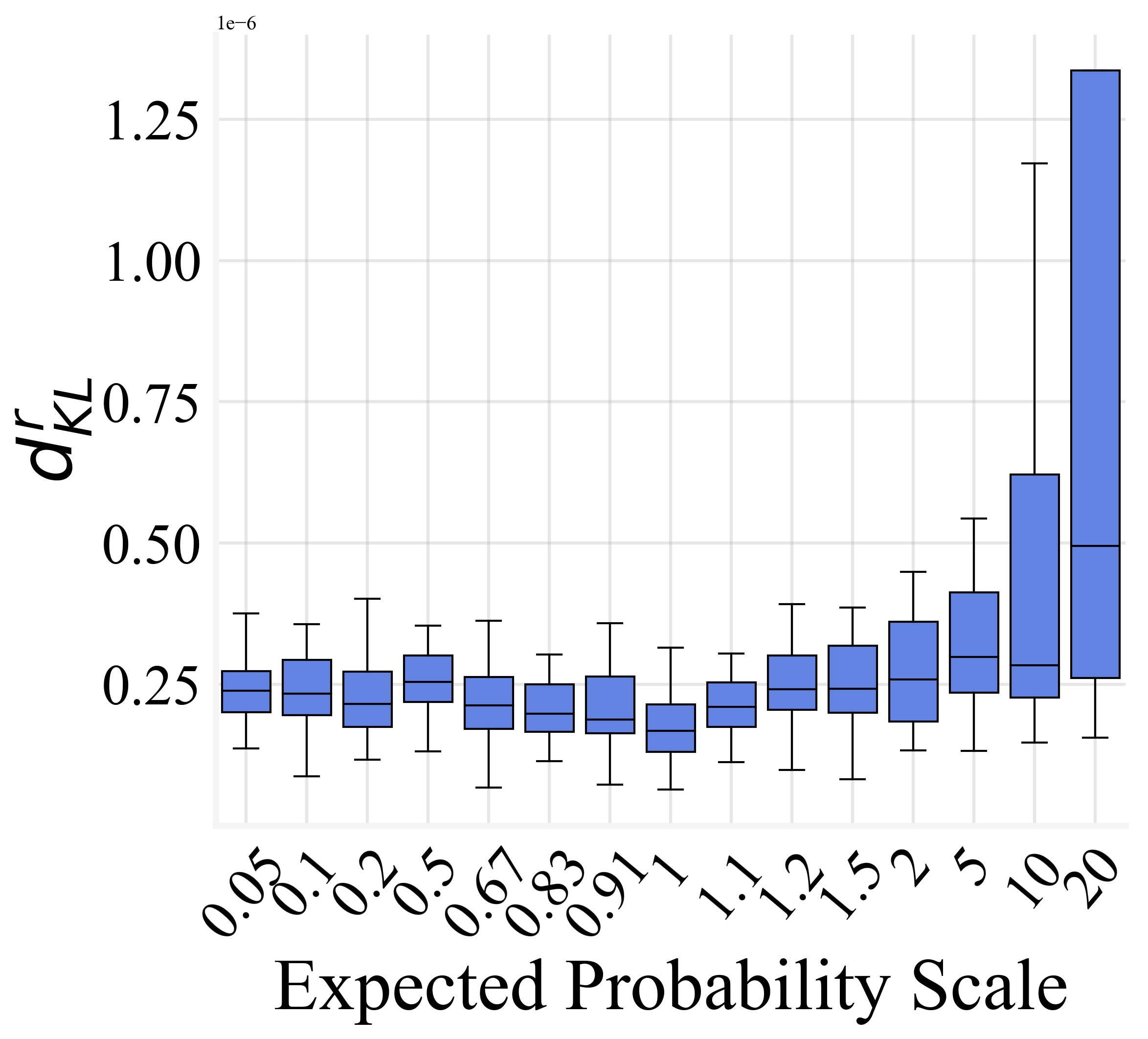}}
\subfloat[$MAUVE$]{
		\includegraphics[width=0.19\linewidth]{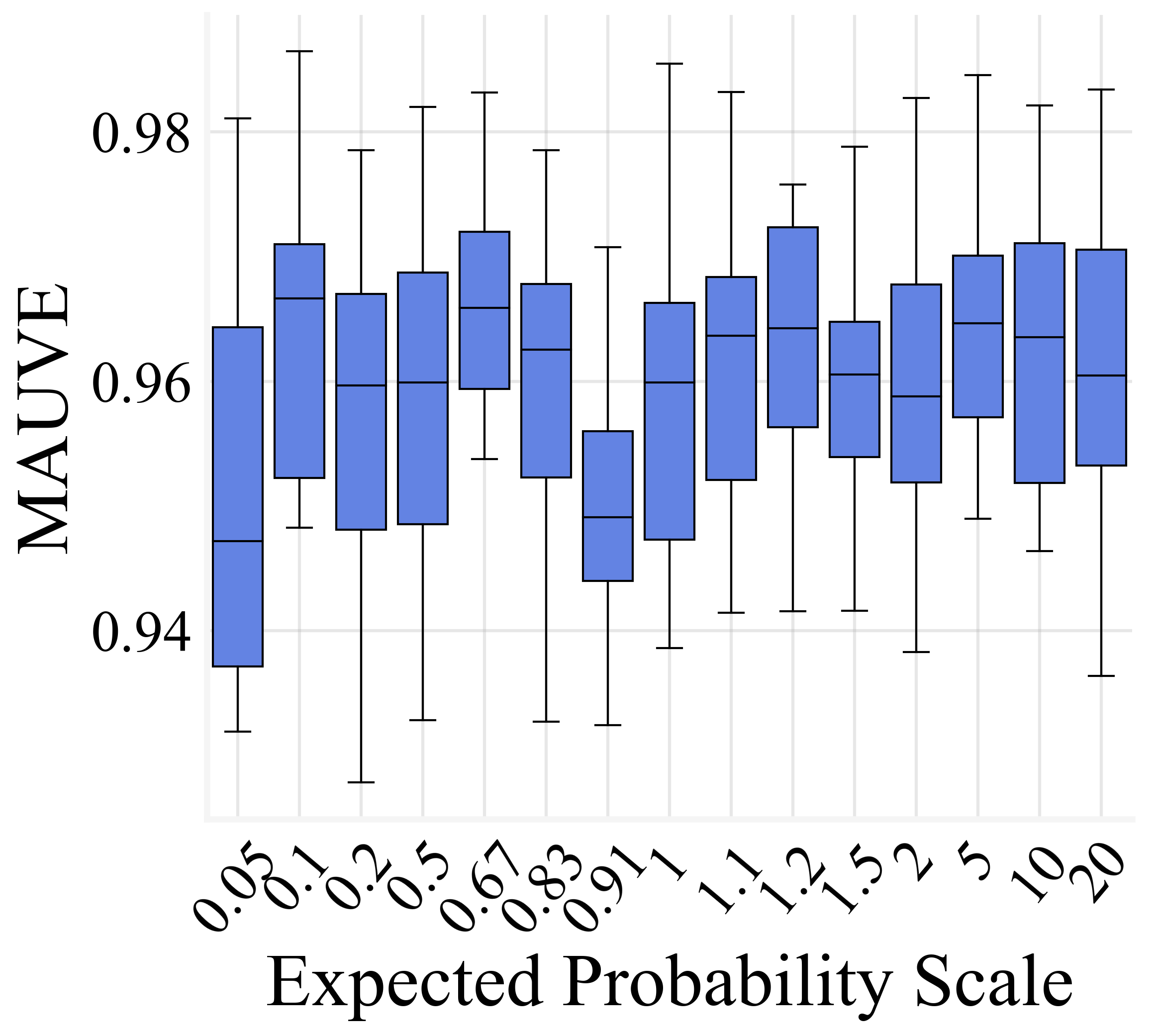}}
\caption{Fine-grained results w.r.t. expected steering scales of GPT2.}
\label{fig:appendix.T1-GPT2}
\end{figure*}

\begin{figure*}[t]
\centering
\subfloat[$e_\mathrm{local}$]{
		\includegraphics[width=0.19\linewidth]{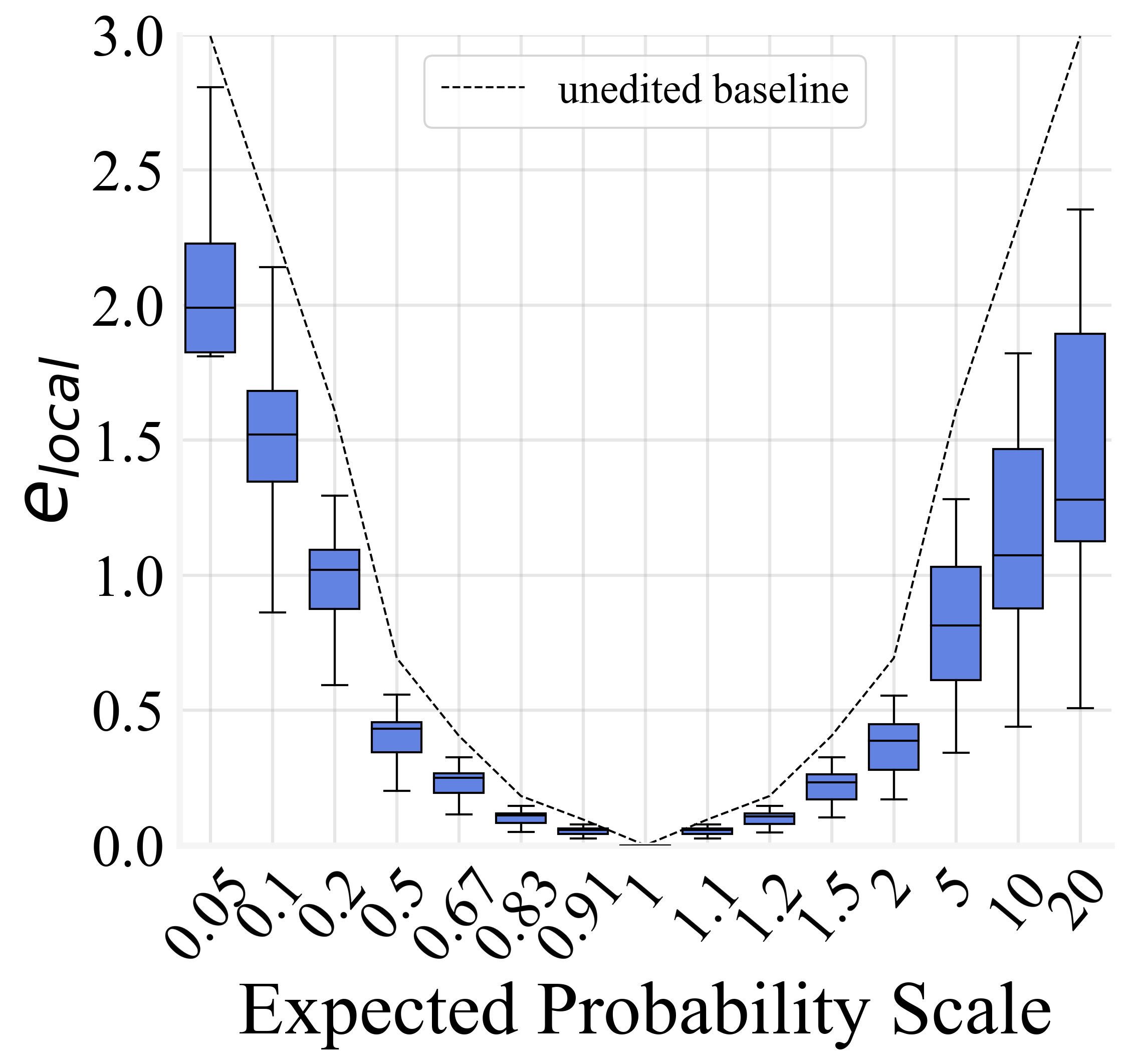}}
\subfloat[$e_\mathrm{id}$]{
		\includegraphics[width=0.19\linewidth]{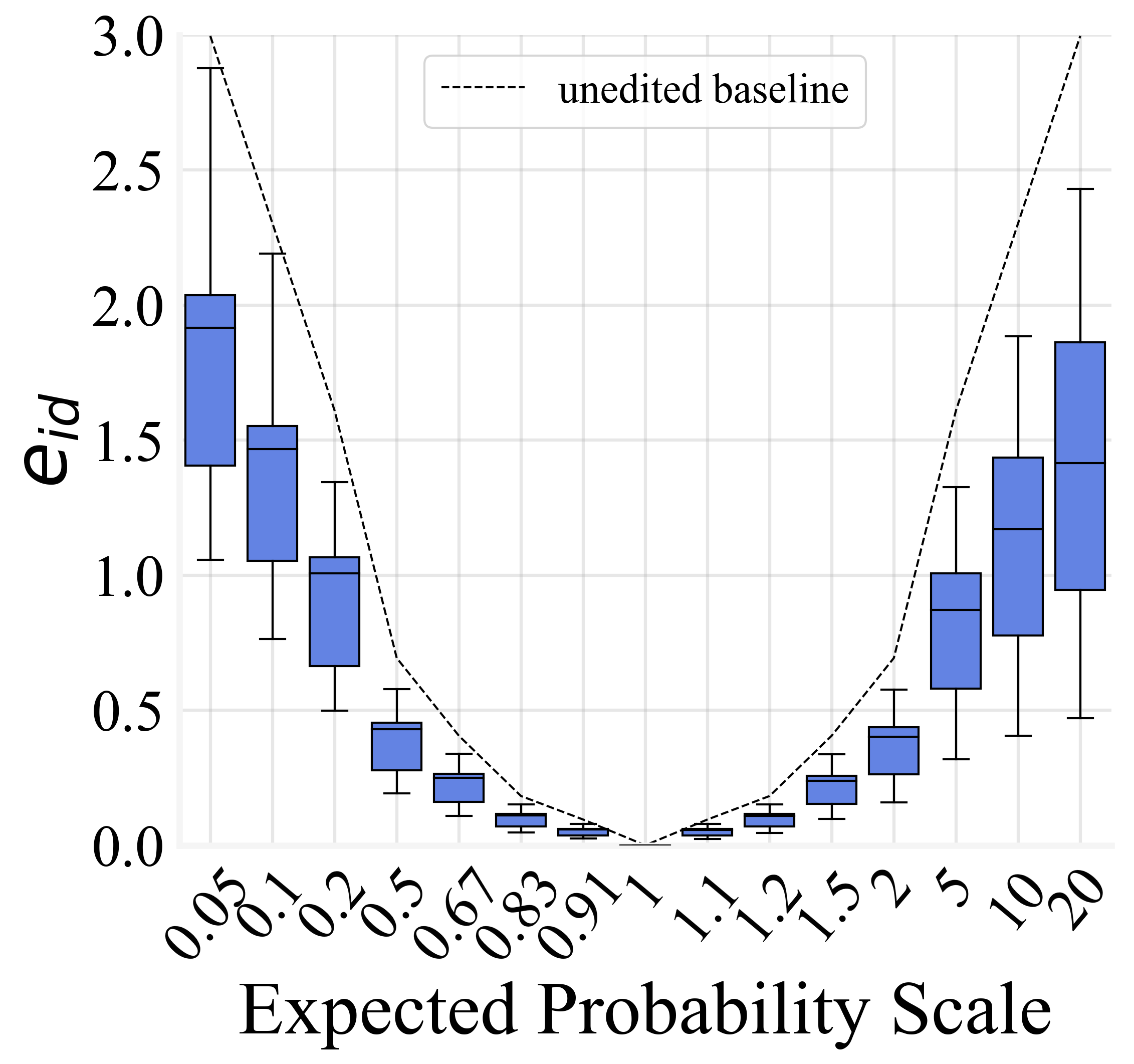}}
\subfloat[$e_\mathrm{ood}$]{
		\includegraphics[width=0.19\linewidth]{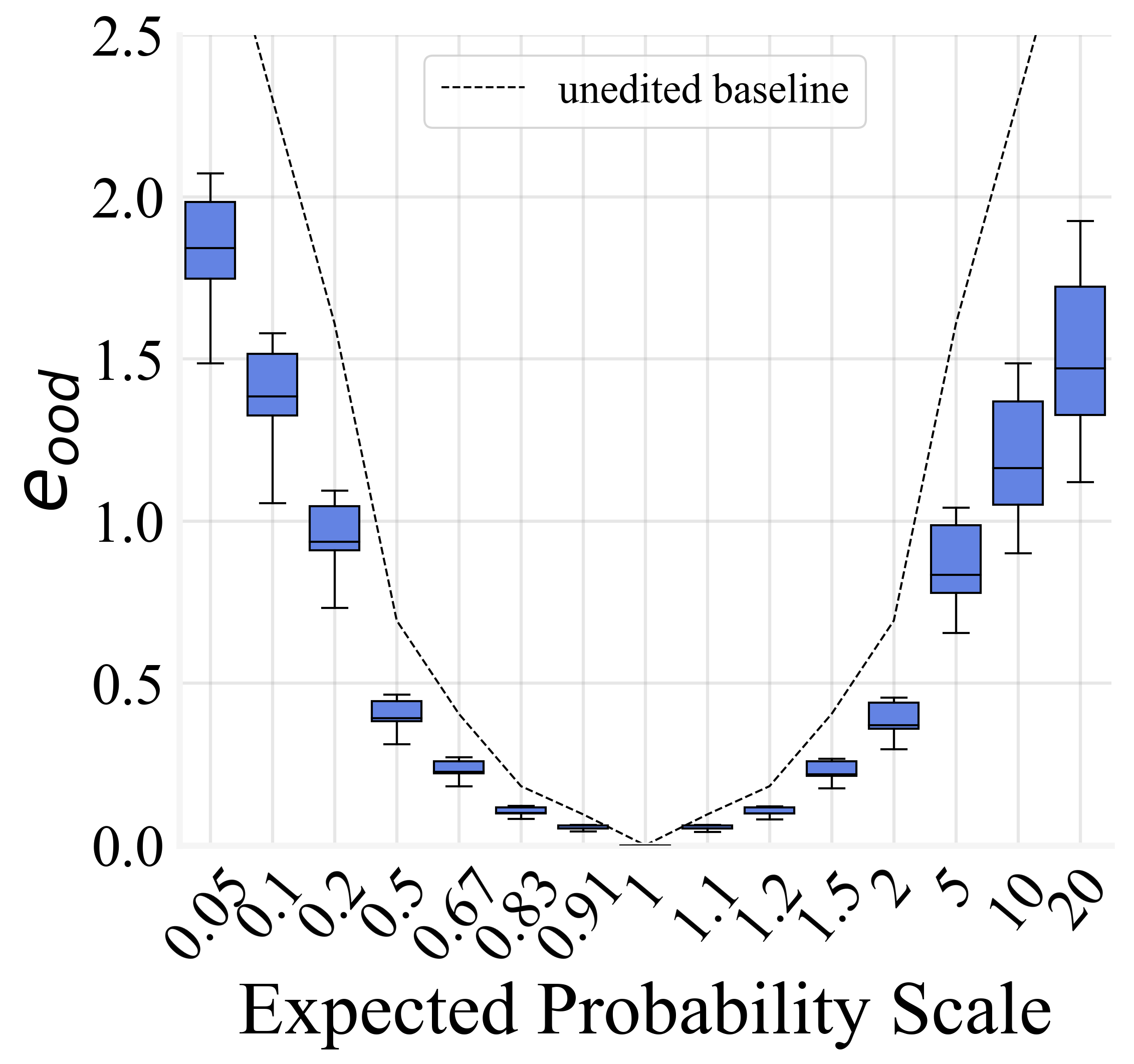}}
\subfloat[$d_\mathrm{KL}^r$]{
		\includegraphics[width=0.19\linewidth]{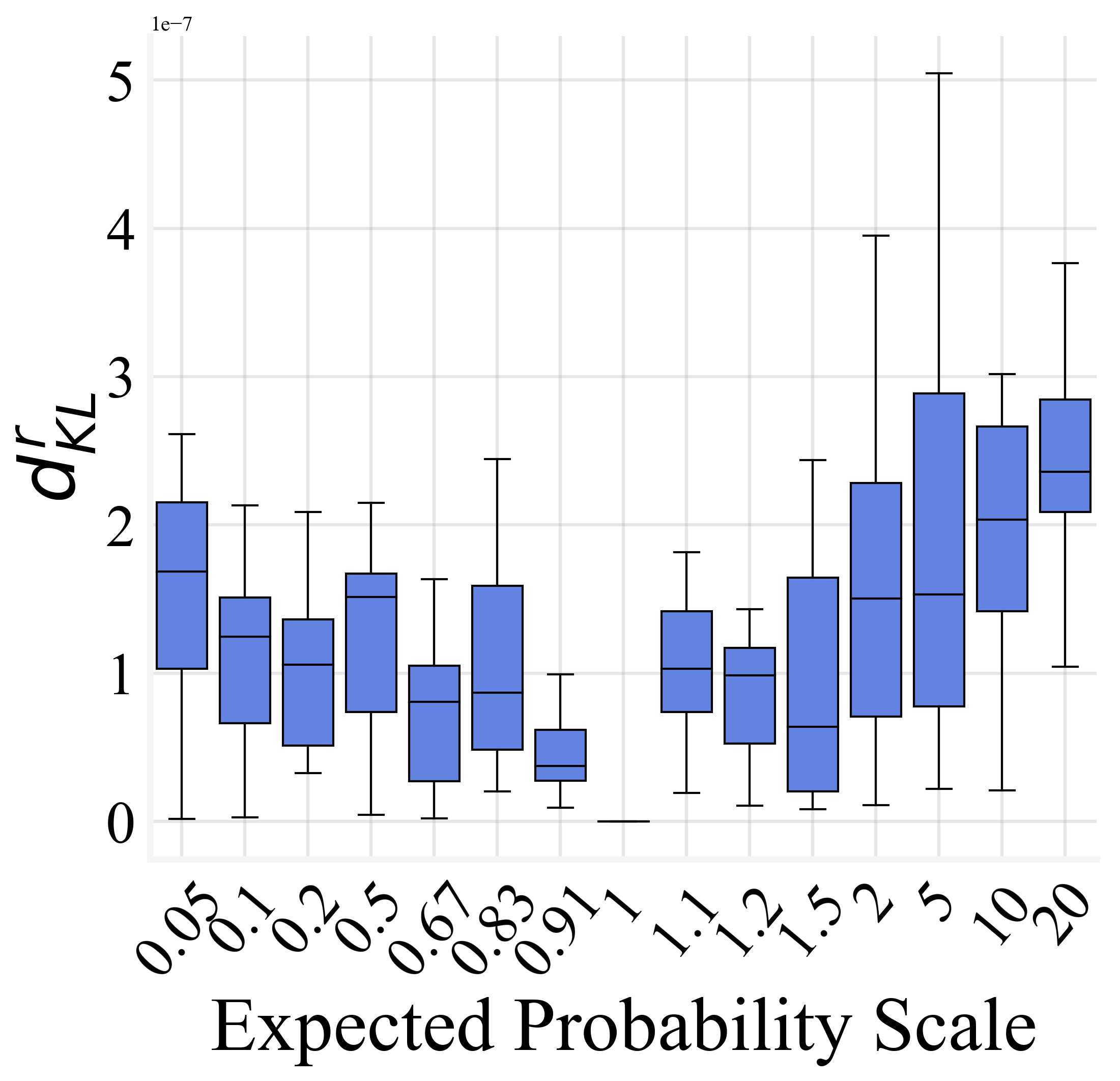}}
\subfloat[$MAUVE$]{
		\includegraphics[width=0.19\linewidth]{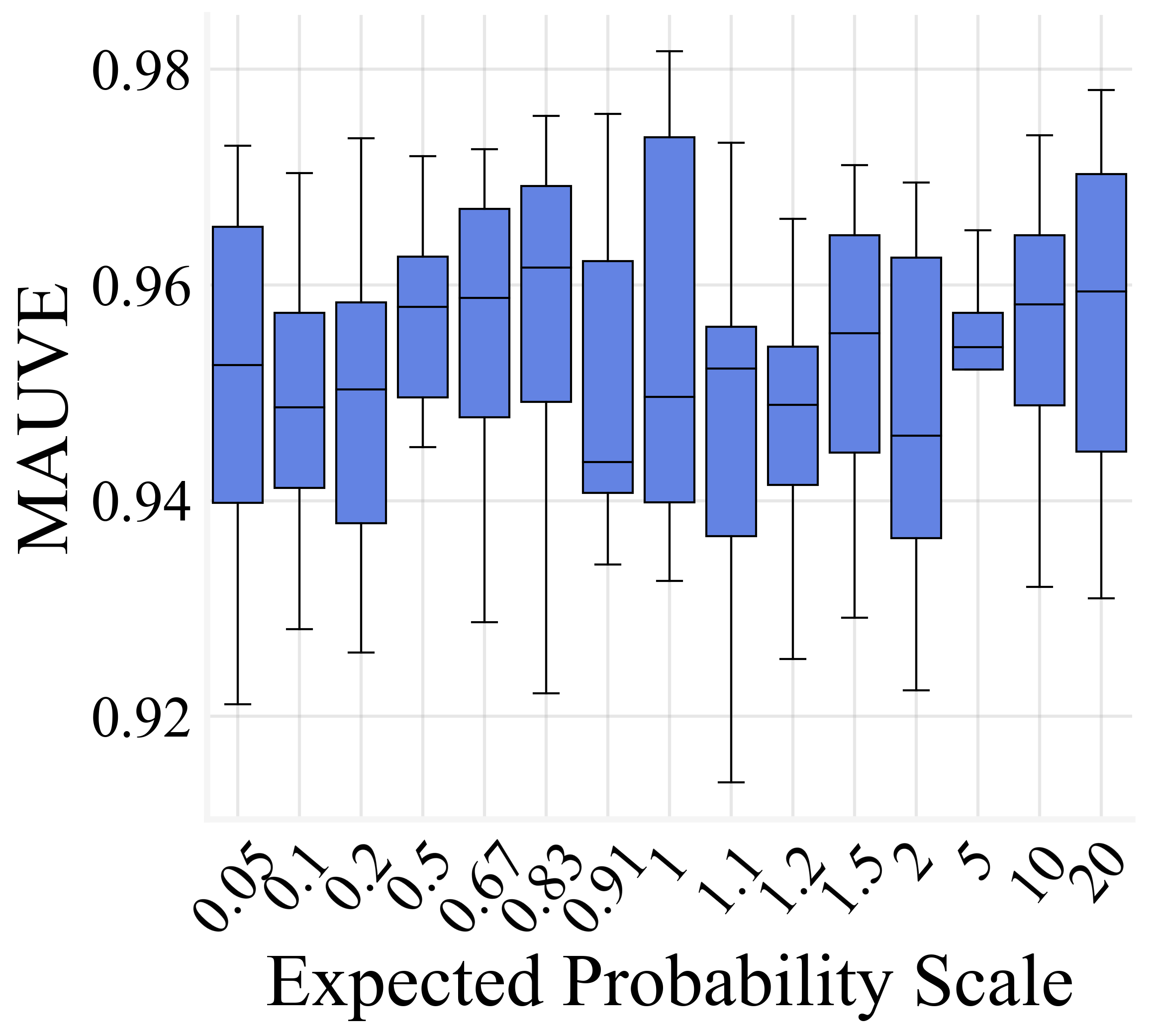}}
\caption{Fine-grained results w.r.t. expected steering scales of GPT2-XL.}
\label{fig:appendix.T1-GPT2XL}
\end{figure*}

\begin{figure*}[t]
\centering
\subfloat[$e_\mathrm{local}$]{
		\includegraphics[width=0.2\linewidth]{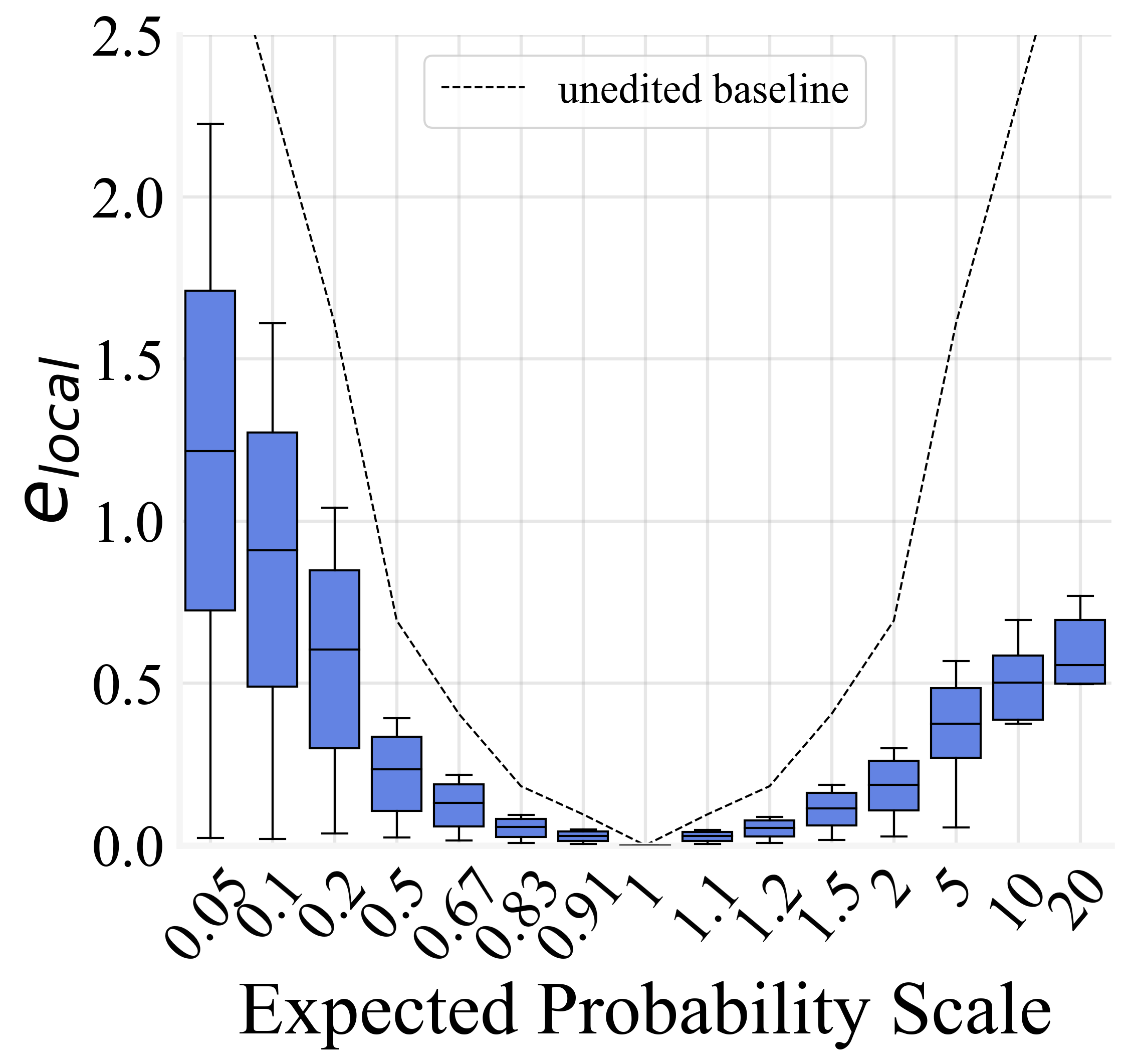}}
\subfloat[$e_\mathrm{id}$]{
		\includegraphics[width=0.2\linewidth]{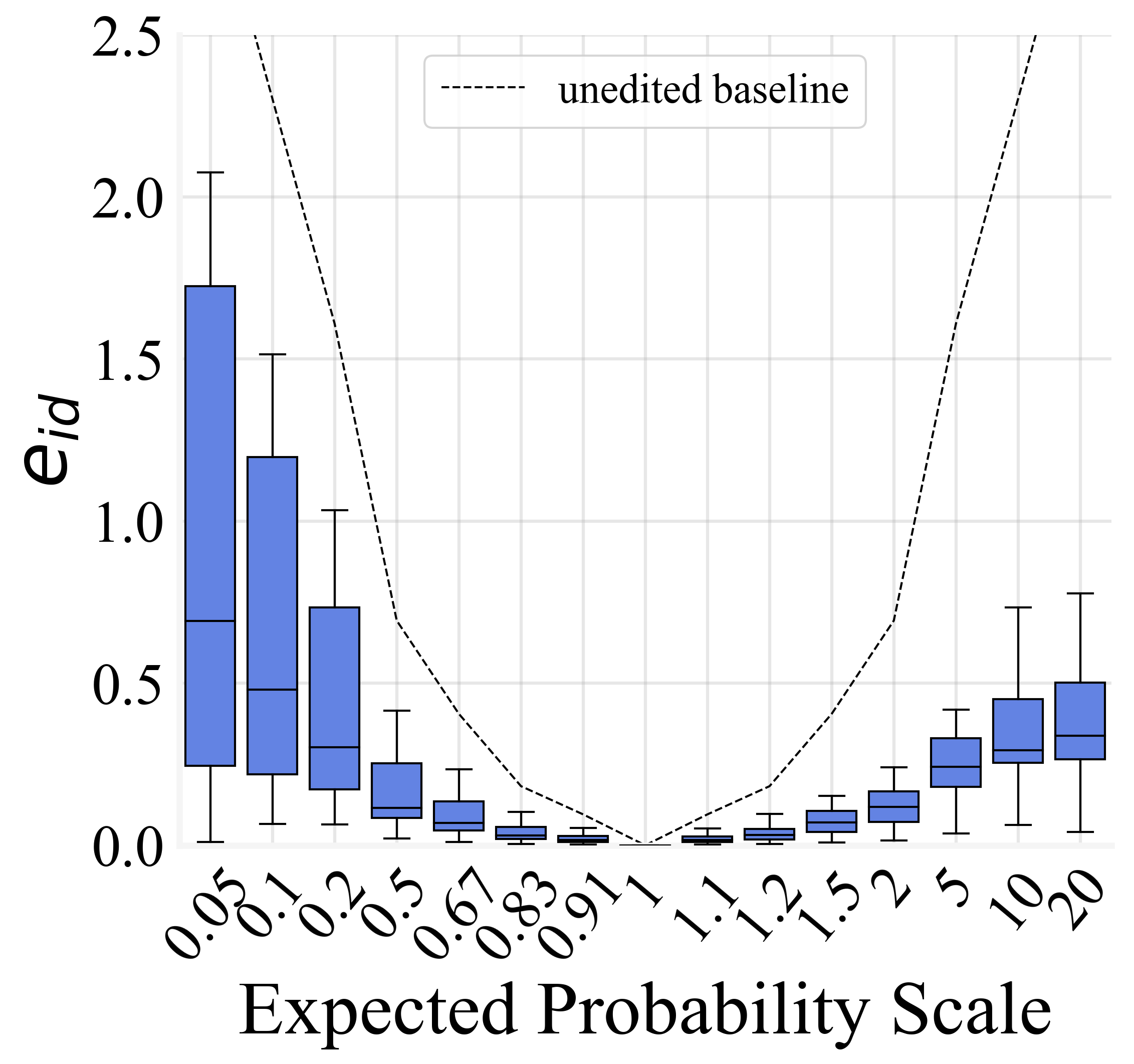}}
\subfloat[$e_\mathrm{ood}$]{
		\includegraphics[width=0.2\linewidth]{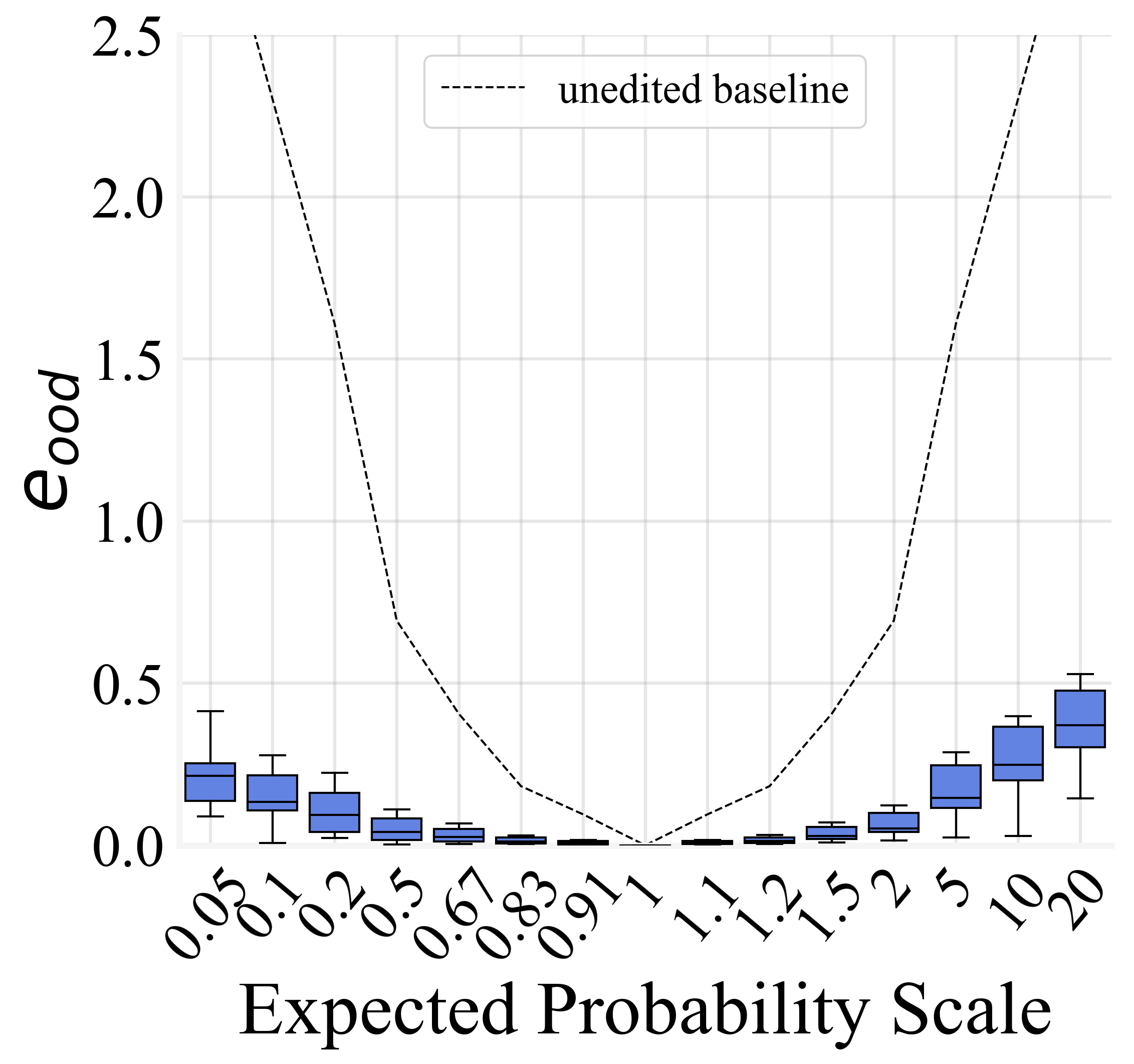}}
\subfloat[$d_\mathrm{KL}^r$]{
		\includegraphics[width=0.2\linewidth]{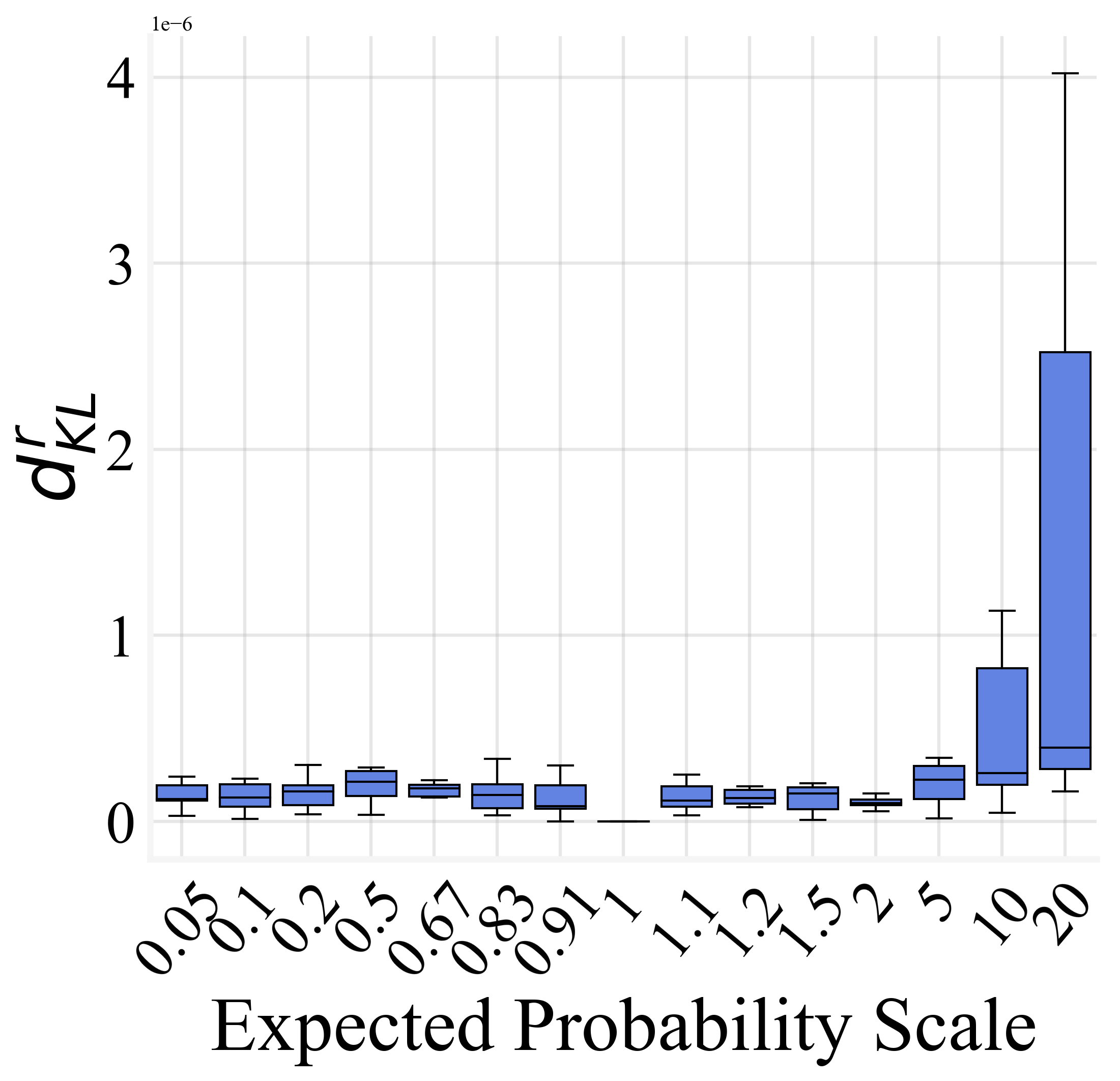}}
\caption{Fine-grained results w.r.t. expected steering scales of GPT-J.}
\label{fig:appendix.T1-GPT2J}
\end{figure*}

\begin{figure*}[t]
\centering
\subfloat[$e_\mathrm{local}$]{
		\includegraphics[width=0.32\linewidth]{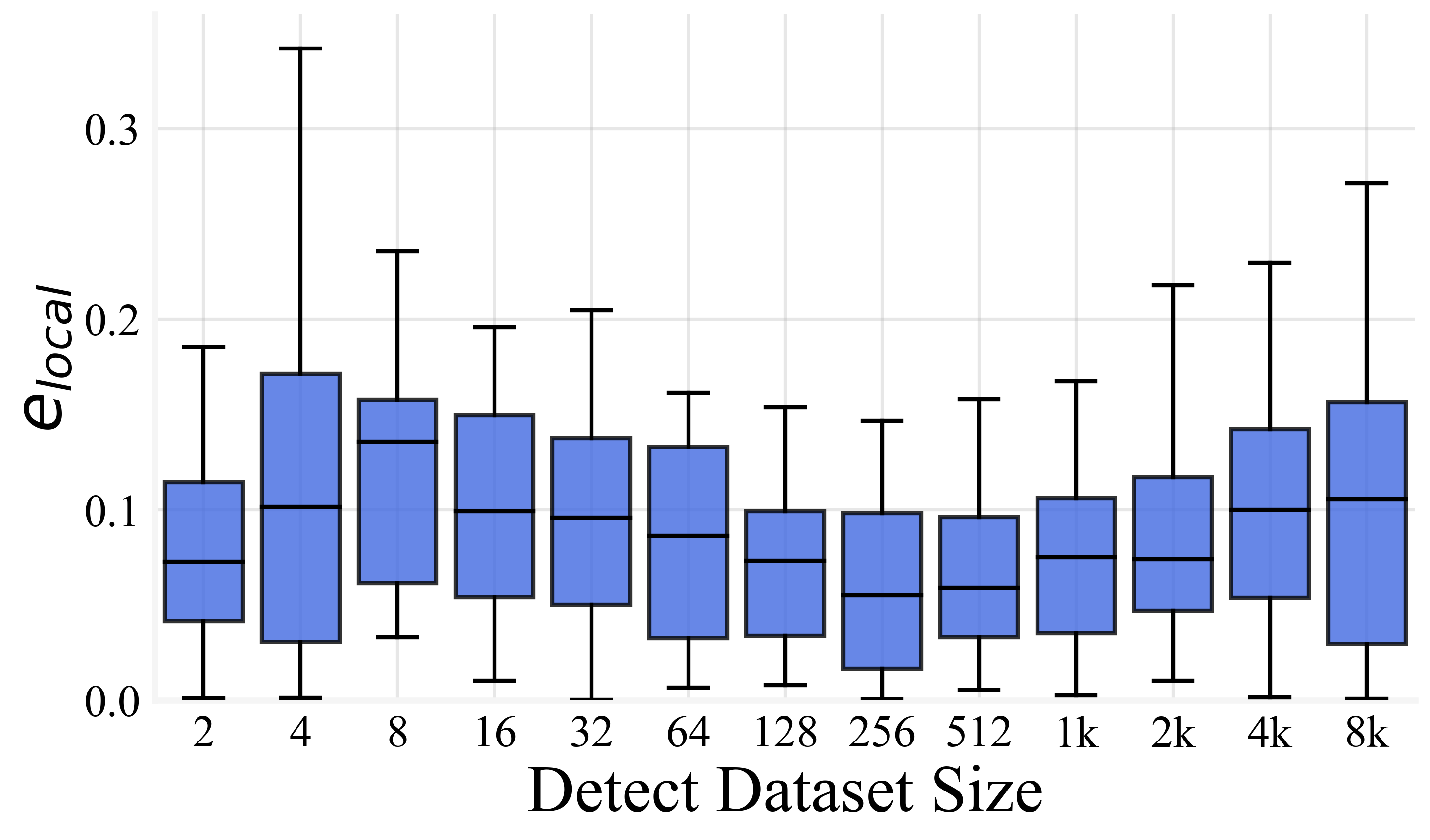}}
\subfloat[$e_\mathrm{id}$]{
		\includegraphics[width=0.32\linewidth]{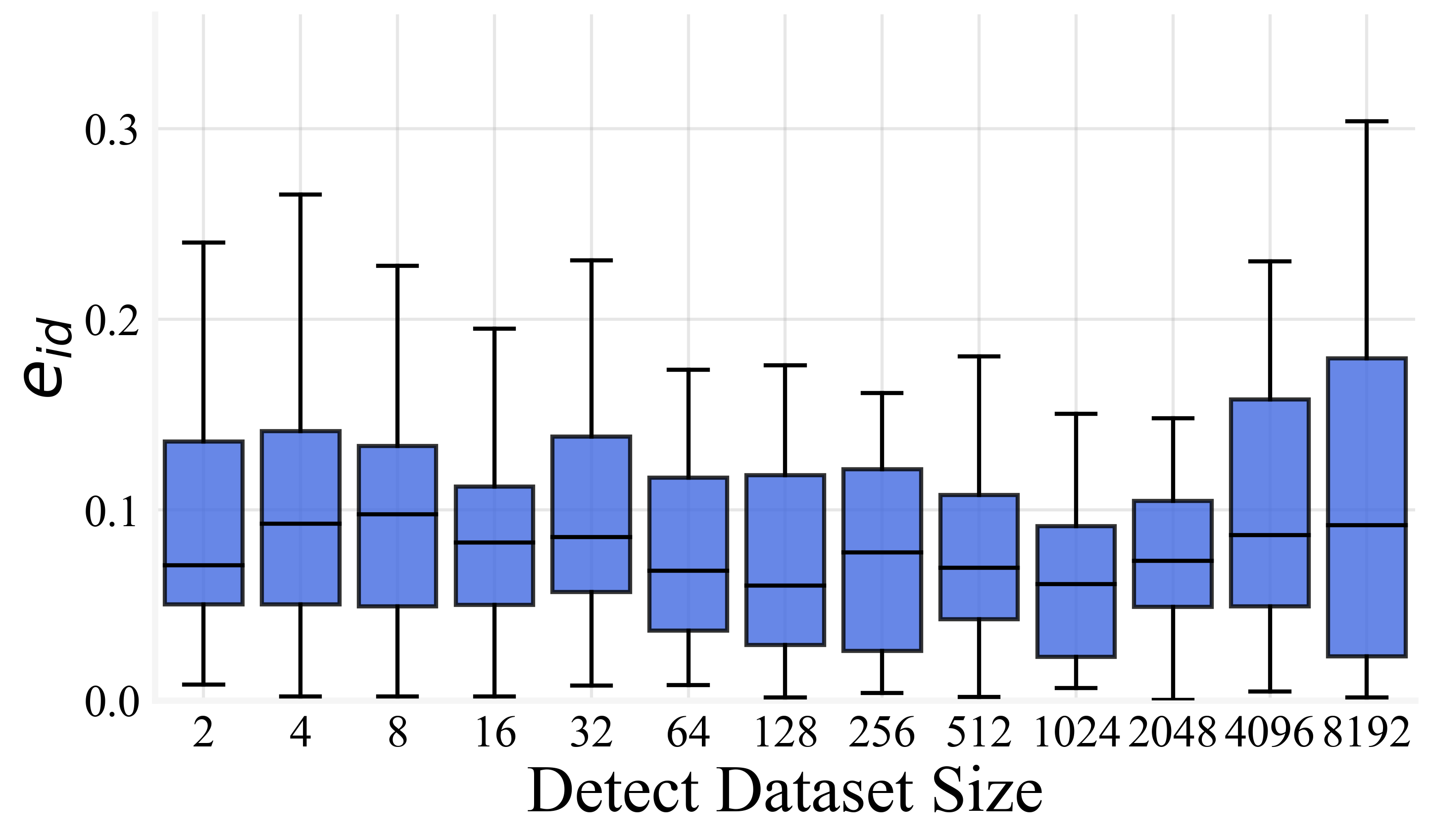}}
\subfloat[$d_\mathrm{KL}^r$]{
		\includegraphics[width=0.32\linewidth]{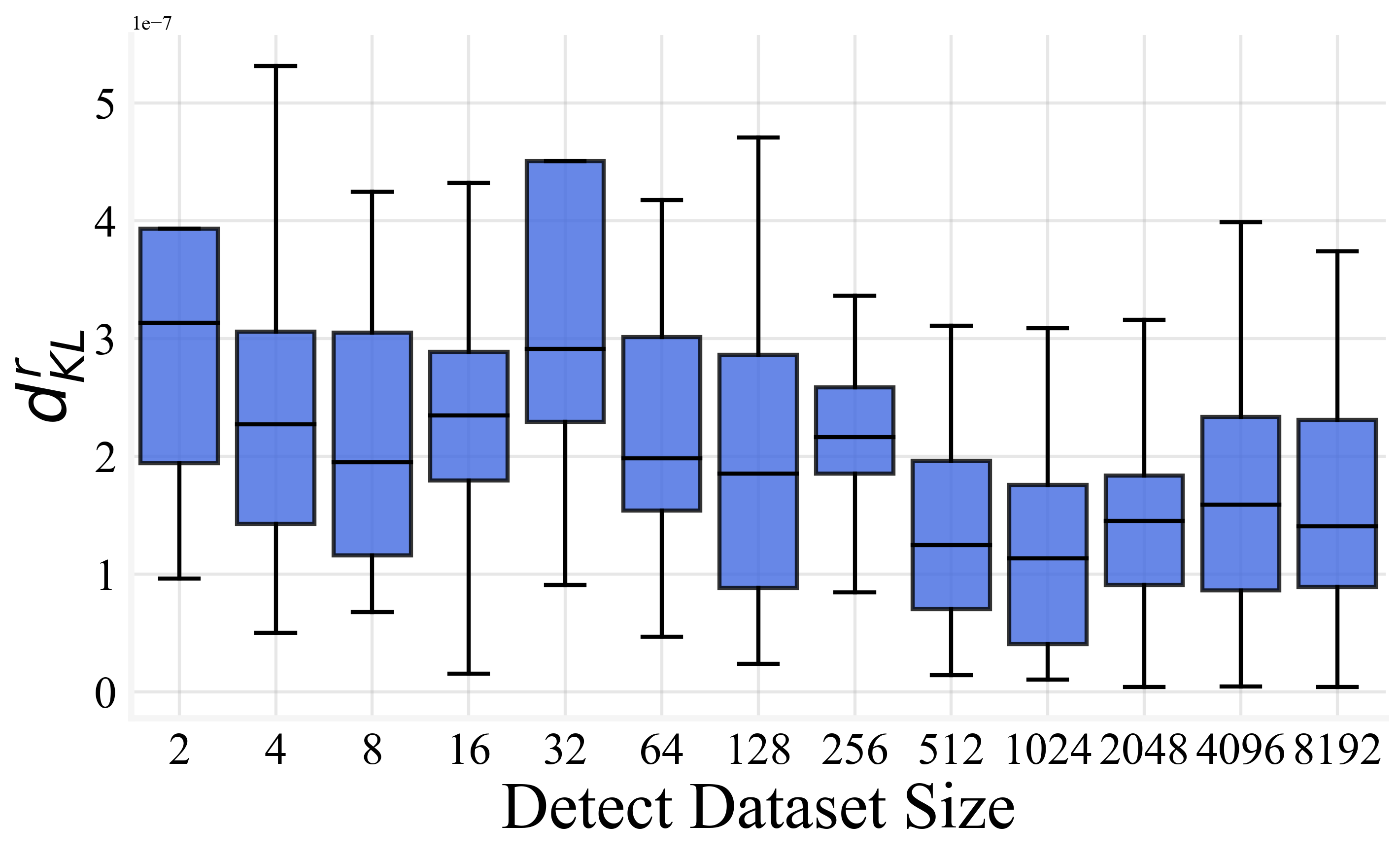}}
\caption{Supplement for Fig.~\ref{fig:length}. The $e_\mathrm{local}$, $e_\mathrm{id}$, $d_\mathrm{KL}^r$ w.r.t. the detect dataset size.}
\label{fig:appendix.S5}
\end{figure*}

\begin{figure*}[t]
\centering
\subfloat{
		\includegraphics[width=0.49\linewidth]{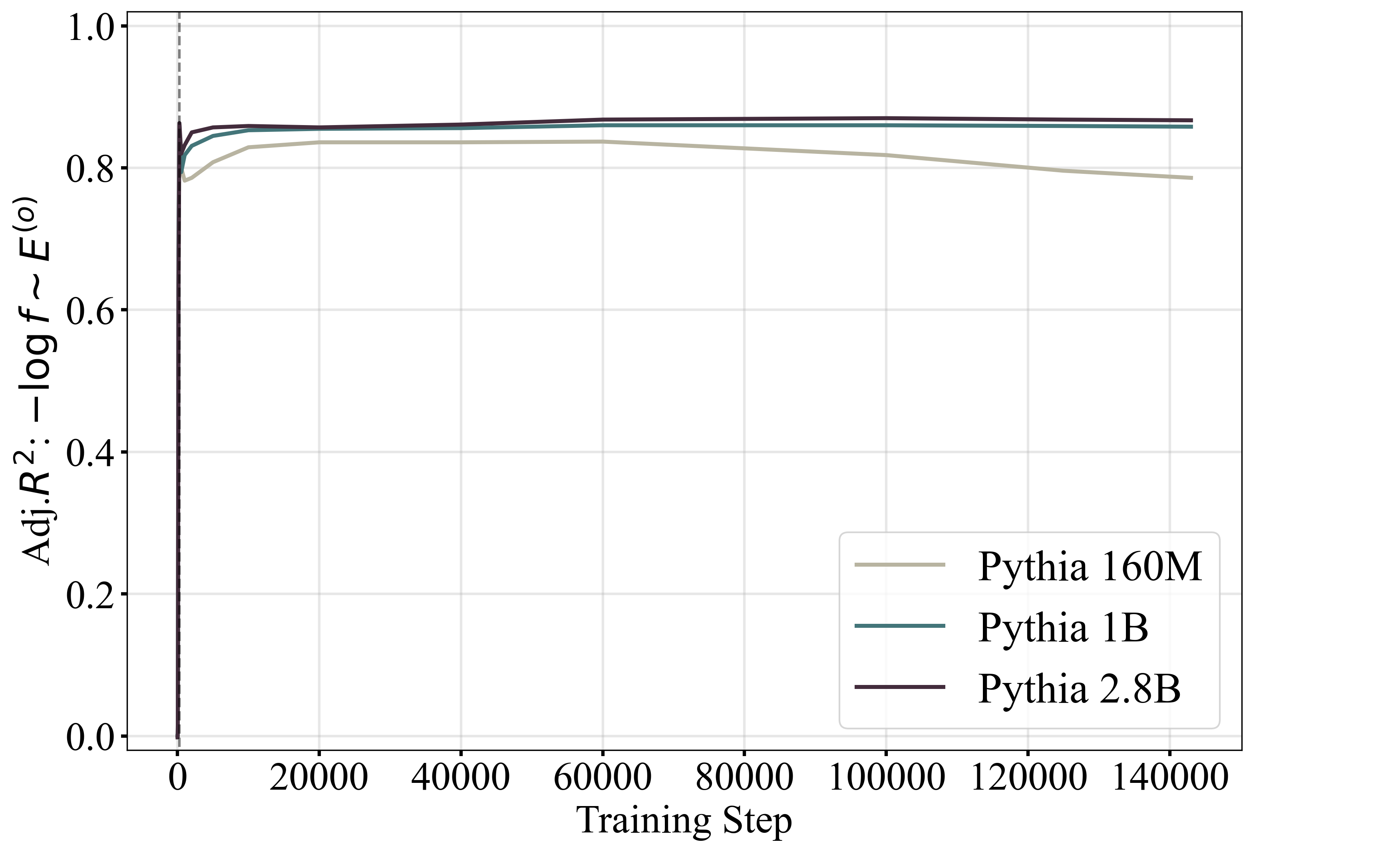}}    
\subfloat{
		\includegraphics[width=0.49\linewidth]{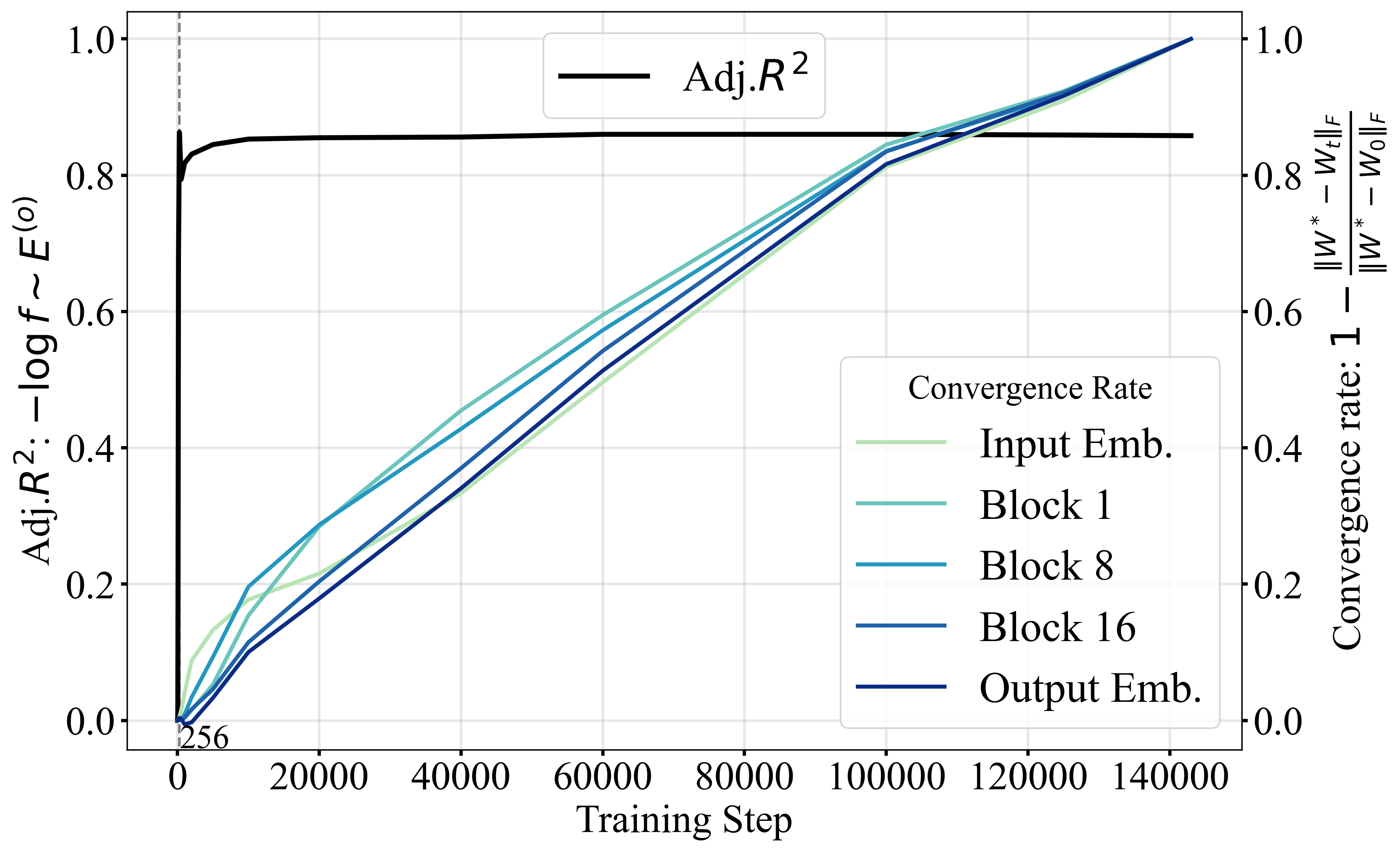}}
\caption{Supplement for Fig.~\ref{fig:TrainD}. Without log-scaling.}
\label{fig:appendix.S7}
\end{figure*}

\subsection{Details of MAUVE Calculation}
\label{sec:appendix.MAUVE}

The MAUVE is a similarity between two datasets. Refer~\citet{pillutla2021mauve} for the calculating details of MAUVE. Here we explain our method to get both datasets for MAUVE to measure the harmfulness of our model steering method. We sample 512 data points from the \textsc{BookCorpus}~\cite{bookcorpus}, and take the first two words of each data point as a prefix to collect a prefix set. Given a generative language model, we input the prefixes and let the model generate naturally into a generated dataset. We do this process on the original model and the steered model and use the two generated datasets to calculate the MAUVE.

\section{Ablation Study on Steering Amount Allocation Parameter $b$}
\label{sec:appendix.Alloc}

We discuss the steering amount allocation parameter $b$ in Algorithm~\ref{alg:algorithm1}. We try different settings of $b$ on GPT2 as shown in Table~\ref{tab:appendix.1} and Fig.~\ref{fig:appendix.X-Y-A}. 

The results show that when the $b$ is not infinities, the steering remains accurate and stable. That is, the algorithm is not sensitive to a positive integer $b$. But we still recommend a large $b$ when the model is large and the sparsity of MLR slope is not fine.

However, when the $b$ is $+\infty$ or $-\infty$, the steering can not be accurate and is easy to get a larger scale than expected. We infer that both of these infinity $b$ increase the norm of the embedding vector too much, leading to an increase in its output probability as well.

\section{Supplementary Experiment Results}
\label{sec:appendix.full}

\noindent\textbf{Supplement for Fig.~\ref{fig:PCA2d-fre}.} Output embedding visualizations w.r.t. output token probability for GPT2-XL, Pythia, GPT-J are shown in Fig.~\ref{fig:appendix.S1}. 

\noindent\textbf{Supplement for Fig.~\ref{fig:fit}.} Output probability fitting results of GPT2-XL and Pythia-2.8B are shown in Fig.~\ref{fig:appendix.S2}. 

\noindent\textbf{Supplement for Fig.~\ref{fig:editxy}.} The expected probability scales and the actually edited scales of GPT2-XL is shown in Fig.~\ref{fig:appendix.S4}.

\noindent\textbf{Supplement for Table~\ref{table:edit_res}.} Fine-grained results w.r.t. expected steering scales of GPT2 (Fig.~\ref{fig:appendix.T1-GPT2}), GPT2-XL (Fig.~\ref{fig:appendix.T1-GPT2XL}), GPT-J (Fig.~\ref{fig:appendix.T1-GPT2J}). 

\noindent\textbf{Supplement for Fig.~\ref{fig:length}.} The $e_\mathrm{local}$, $e_\mathrm{id}$, $d_\mathrm{KL}^r$ w.r.t. the detect dataset size are shown in Fig.~\ref{fig:appendix.S5}. 

\noindent\textbf{Supplement for Fig.~\ref{fig:TrainD}.} The figures without log-scaling are shown in Fig.~\ref{fig:appendix.S7}.

\section{Necessary Statements}
\label{sec:state}

\paragraph{Author Contributions.} Hakaze Cho, also pronounced Yufeng Zhao, handled virtually all the workload in this paper. N.I. is their supervisor, providing important and beneficial guidance, revision, and research support. Y.S., K.T, and M.K. participated in the discussion and provided some helpful suggestions and revisions. 

\paragraph{Repeatability statement.} Models and datasets are all loaded from \verb|huggingface|. All the datasets are shuffled with random seed \verb|42| and cut into required slices. We calculate MAUVE by the package \verb|mauve-text| by default parameters. In experiments, all the logarithms are natural.

\paragraph{License of the artifacts.} The artifacts used in this paper are all open-sourced and are used for their intended usage. \textbf{Models.} GPT2 family are under the \verb|MIT| license, GPT-J and Pythia are under the \verb|apache-2.0| license. \textbf{Datasets.} \textsc{WikiDpr} is under the \verb|cc-by-nc-4.0| license, \textsc{BookCorpus} and \textsc{Pile} is under the \verb|MIT| license. \textbf{Tool.} \textsc{MAUVE} is under the \verb|GPLv3| license.

\end{document}